%% file: main.tex
\documentclass[acmsmall]{acmart}
\AtBeginDocument{%
  \providecommand\BibTeX{{%
    \normalfont B\kern-0.5em{\scshape i\kern-0.25em b}\kern-0.8em\TeX}}}

\setcopyright{acmcopyright}
\copyrightyear{2022}
\acmYear{2022}
\acmDOI{XXXXXXX.XXXXXXX}

\acmJournal{CSUR}
\acmVolume{37}
\acmNumber{4}
\acmArticle{111}
\acmMonth{8}
\usepackage{subcaption}
\usepackage{enumitem}
\usepackage{booktabs}
\usepackage{tabularx}
\usepackage{acro}
\usepackage{makecell}
\usepackage{multirow}
\input{acronyms}
\usepackage{vcell}
\usepackage{tabularray}

\begin{document}

\title{Towards Trustworthy Machine Learning in Production: An Overview of the Robustness in MLOps Approach}

\author{Firas Bayram}
\email{firas.bayram@kau.se}
\orcid{0000-0003-0683-2783}
\affiliation{%
  \institution{Department of Mathematics and Computer Science, Karlstad University}
  \streetaddress{Universitetsgatan 2, 65188}
  \city{Karlstad}
  \country{Sweden}}

\author{Bestoun S. Ahmed}
\affiliation{%
  \institution{Department of Mathematics and Computer Science, Karlstad University}
  \streetaddress{Universitetsgatan 2, 65188}
  \city{Karlstad}
  \country{Sweden}}
\email{bestoun@kau.se}
\orcid{0000-0001-9051-7609}

\renewcommand{\shortauthors}{F.Bayram et al.}

\begin{abstract}
Artificial intelligence (AI), and especially its sub-field of Machine Learning (ML), are impacting the daily lives of everyone with their ubiquitous applications. In recent years, AI researchers and practitioners have introduced principles and guidelines to build systems that make reliable and trustworthy decisions. From a practical perspective, conventional ML systems process historical data to extract the features that are consequently used to train ML models that perform the desired task. However, in practice, a fundamental challenge arises when the system needs to be operationalized and deployed to evolve and operate in real-life environments continuously. To address this challenge, Machine Learning Operations (MLOps) have emerged as a potential recipe for standardizing ML solutions in deployment. Although MLOps demonstrated great success in streamlining ML processes, thoroughly defining the specifications of robust MLOps approaches remains of great interest to researchers and practitioners. In this paper, we provide a comprehensive overview of the trustworthiness property of MLOps systems. Specifically, we highlight technical practices to achieve robust MLOps systems. In addition, we survey the existing research approaches that address the robustness aspects of ML systems in production. We also review the tools and software available to build MLOps systems and summarize their support to handle the robustness aspects. Finally, we present the open challenges and propose possible future directions and opportunities within this emerging field. The aim of this paper is to provide researchers and practitioners working on practical AI applications with a comprehensive view to adopt robust ML solutions in production environments.

\end{abstract}




\maketitle

\section{Introduction}
\label{sec:intro}
In the current landscape of real-world applications, the exponential generation of data coupled with their inherent complexity and dynamism requires the development of efficient analysis mechanisms. With the increasing difficulty in managing data-driven applications for monitoring and prediction, the deployment of artificial intelligence (AI) systems has played a vital role in real-world applications due to fast-paced advances and innovation in the field \cite{sarker2022ai}. These AI systems, equipped with various toolkits, offer a wide range of solutions that can be adopted to help decision makers take rapid action in various applications \cite{al2020generalizing}. Machine learning (ML), the heart of AI, empowers applications to learn and adapt automatically, paving the way for intelligent and autonomous operation. \cite{jordan2015machine}. 

Traditionally, the development pipeline of ML systems involves the processing of historical data, which represents the raw input, to extract the features to train an ML model(s) that would perform the desired task offline to produce the output \cite{du2019techniques}. The pipeline includes additional sub-tasks like feature engineering, validation, testing, and performance evaluation. However, this static approach, while useful for initial research and experimentation, often proved inadequate in real-world scenarios. The fundamental challenge arises when the ML solution needs to be operationalized and deployed to function in a real-world environment \cite{rudin2014machine}. In such scenarios, additional constraints and limitations should be specified to perform the tasks that characterize the ML solution, as the offline solution may not meet real-world requirements. For that, the transformation from offline to online ML system implementation has recently been of great importance to shift towards systems that perform in real-time and continuously circulate in production.

Machine Learning Operations (MLOps) have emerged as a potential recipe for designing ML solutions in online deployment scenarios. MLOps is an extension of DevOps, which is a well-known and mature topic in software engineering, to run ML systems in production \cite{ruf2021demystifying}. MLOps is a set of practices and techniques that aims to automate the delivery of ML models to production \cite{alla2021beginning}. Conceptually, MLOps adopts DevOps principles for building ML systems. In particular, the core principles MLOps uses from DevOps are continuous integration (CI) and continuous delivery (CD). Iteratively and continuously, CI/CD facilitates the automated process of releasing reliable software in short cycles \cite{humble2010continuous}. However, ML systems exhibit different specifications and involve more components than traditional software systems, which require special management and maintenance. MLOps supports the development of ML systems in all phases of the entire ML workflow, from data retrieval to the release and deployment of the solution.

\begin{figure}
    \centering
    \includegraphics[scale=0.25]{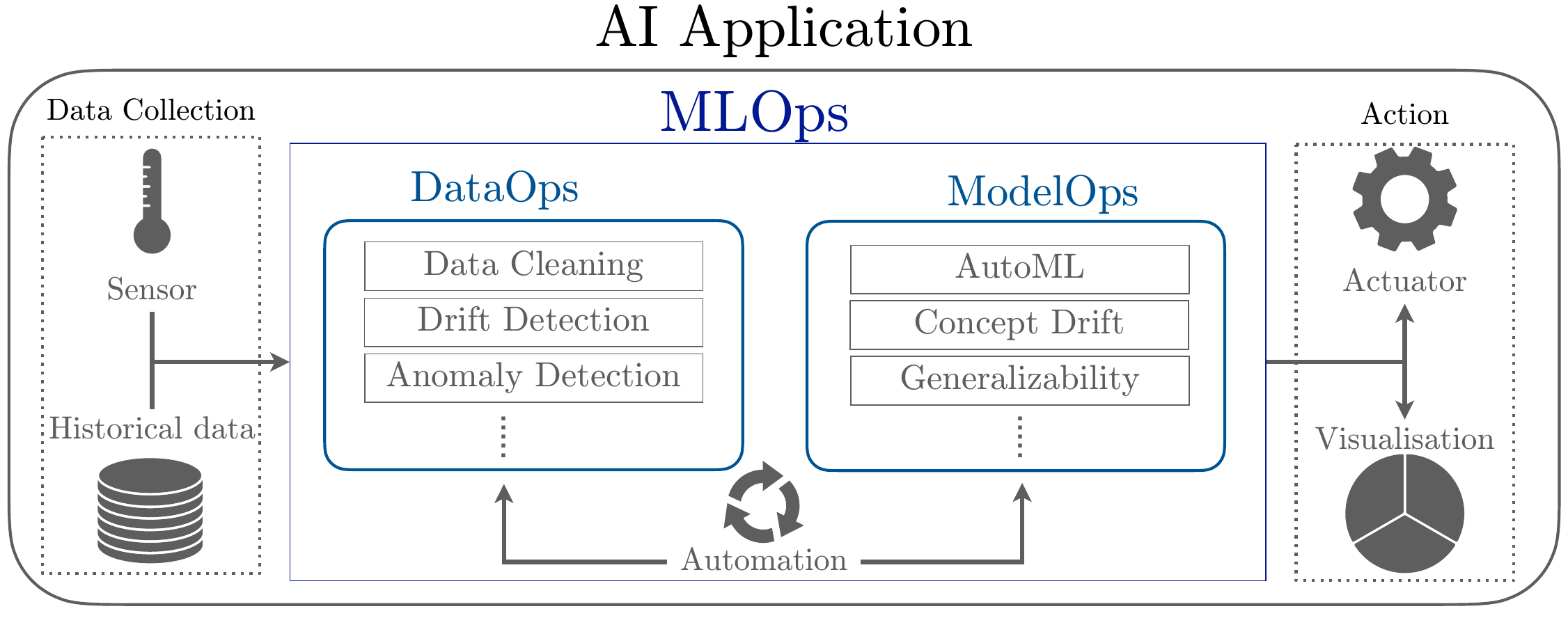}
    \caption{MLOps system inside the AI application}
    \label{fig:AI_MLOps}
\end{figure}

As illustrated in Fig. \ref{fig:AI_MLOps}, MLOps serves as a vital companion throughout the lifecycle of AI products, ensuring the continuous monitoring and delivery of ML systems. However, there exists a great need to establish and clarify the elements and standards necessary for achieving optimal system performance. During the last decades, ML research has been highly dependent on quantitative metrics such as accuracy and loss measures to identify system performance and guide the selection of the ML approach \cite{jobin2019global}. However, at the holistic level of AI application systems, such metrics may not fully capture task performance. For example, a technically accurate solution that compromises ethical standards, such as utilizing private information in datasets or deploying an ML system with weak robustness to changes, poses significant concerns for AI systems. To address these issues, researchers and practitioners in the AI community have begun to consider these aspects and introduce more principles while building AI systems \cite{li2021trustworthy}. The aspects and principles are known as AI trustworthiness \cite{wing2021trustworthy}.

Recently, trustworthy AI has made substantial strides in both academia and industry. Many governing bodies, organizations, and companies have formulated requirements and introduced principles to build trustworthy AI systems. The most popular definition of trustworthy AI was established by the European Union\footnote{\url{https://ec.europa.eu/newsroom/dae/document.cfm?doc_id=60419}}, which states that trustworthy AI systems have three key components: lawful, ethical, and robust \cite{chatila2021trustworthy}. Naturally, any MLOps framework designed within a trustworthy AI product must inherit the same principles and meet the requirements. In fact, from a technical standpoint, the MLOps pipeline is the most significant part of the AI system in production, and the implementation of such a trustworthy strategy must be thoroughly considered when devising an operationalized solution for an AI application.

Since MLOps is more concerned with the performance and quality of the system \cite{renggli2021data} and has less to do with ethical and lawful issues, we will narrow the focus of this paper to the technical robustness aspect of trustworthy AI systems. Robustness, in particular, is of major importance in practical MLOps systems. A robust ML system must be able to retain its effectiveness even when faced with unforeseen or unexpected challenges. The occurrence of system failures or inaccuracies in such settings can potentially cause interruptions or safety risks for users. Therefore, to maintain user trust and confidence in AI solutions, the robustness aspect of MLOps systems must be thoroughly addressed during development and deployment to secure sustained reliability in the long run and ultimately the responsible and effective deployment of ML in the dynamic real world.

This paper formalizes the specifications of robust ML systems in production through an in-depth literature search. In particular, we explore the connection between the robustness property, as a principal component of trustworthy AI systems, and the MLOps approach, as a primary methodology to streamline ML solutions. By formalizing this connection, which has not been exhaustively covered in the literature, we aim to provide researchers and practitioners with a consolidated view of the field.
We summarize the contribution of this paper as follows:
\begin{itemize}
    \item We organize the robustness aspects of ML solutions in production into three components and provide detailed definitions of the related concepts.
    \item We derive a set of specifications for a robust ML system, which is a seminal part of trustworthy AI systems, and reflect it with the workflow of ML solutions in production.
    \item We give a comprehensive overview of the specifications of robust ML systems and the constituent elements of MLOps systems.
    \item We review the literature on existing approaches that address the robustness aspects of ML systems in production.
    \item We iterate the available tools to build MLOps frameworks and discuss their support to deal with the various robustness aspects.
    \item Based on our broad analysis, we propose future directions and opportunities for researchers to further investigate in the area.

\end{itemize}

The remainder of this paper is organized as follows. In Section \ref{sec:related}, we provide a general overview of the field and details of the research methodology followed in this work on surveying trustworthy AI and MLOps. 
% In Section \ref{sec:concepts}, we provide a general overview and definitions related to the specifications of trustworthy AI systems and the elements of MLOps systems.
In Section \ref{sec:spec}, we derive specifications to achieve robust MLOps systems and provide definitions related to the specifications of trustworthy AI systems and the elements of MLOps systems. In Section \ref{sec:acd_surv}, we survey the approaches of academic research that address the robustness aspect of ML systems in production and summarize their pros and cons. We compare the different technologies used to build MLOps systems in Section \ref{sec:tool_surv}. We provide open challenges and potential future directions in Section \ref{sec:future}. Finally, we conclude the survey with Section \ref{sec:conc}.

\section{Motivation and Research Method}
\label{sec:related}
\begin{figure}
\begin{subfigure}[b]{0.49\textwidth} 
    \centering
    \includegraphics[scale=0.41]{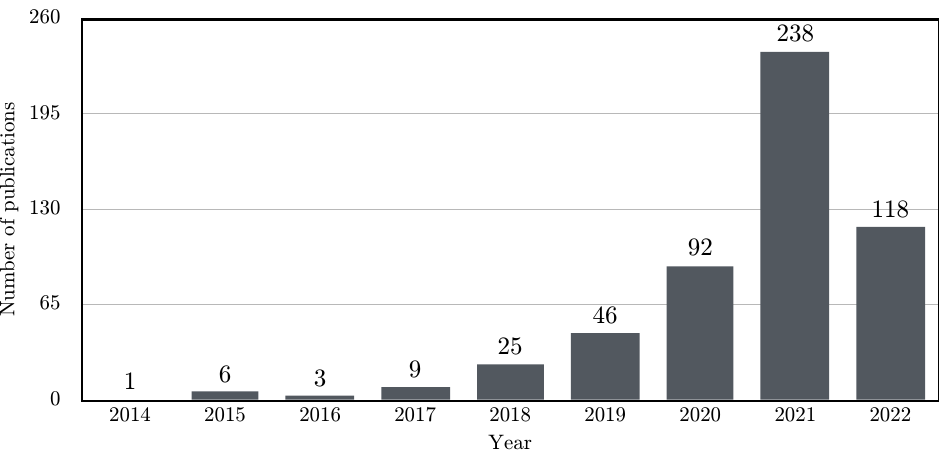}
    \caption{Trustwothy AI.}
    \label{fig:TAI_year}
\end{subfigure}%
\hfill
\begin{subfigure}[b]{0.49\textwidth}
    \centering
    \includegraphics[scale=0.41]{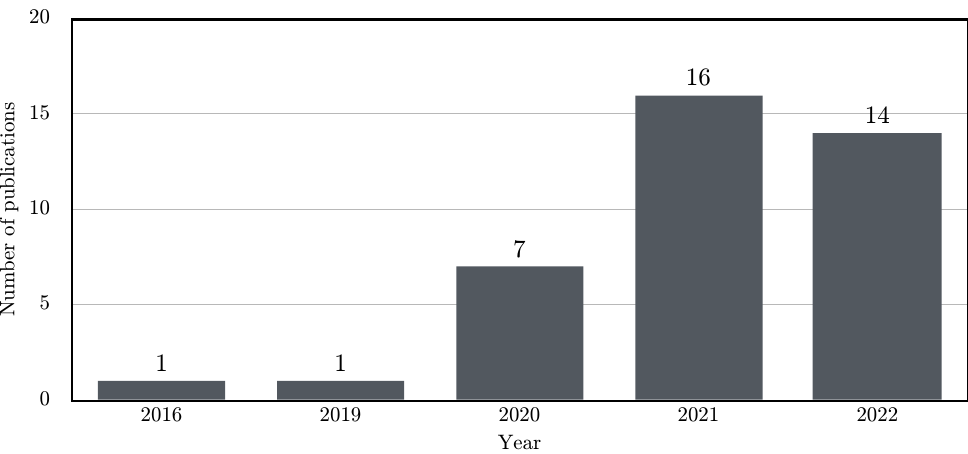}
    \caption{MLOps.}
    \label{fig:MLOPS_year}
\end{subfigure}
\caption{Number of publications retrieved by searching the Web of Science indexing platform for the terms trustworthy AI and MLOps}
\label{fig:pub_year}
\end{figure}

Trustworthy AI and MLOps are relatively new research topics in the maturing field of artificial intelligence (AI) in recent years. When searching for the terms trustworthy AI and MLOps on the Web of Science indexing platform\footnote{\url{https://webofscience.com/}}, the results reveal that the two topics have begun to attract increasing research attention over the past five years. However, trustworthy AI is a more mature topic than MLOps in the literature. As shown in Fig. \ref{fig:pub_year}, trustworthy AI began to emerge in 2018 with 25 publications. The number increased to 238 publications in 2021 and 118 in 2022 (the year of writing this manuscript). Similarly to MLOps, publications have increased in the past two years, from 1 publication in 2019 to 16 in 2021.

\subsection{Trustworthy AI}
Trustworthy AI is a framework to verify that the system is operating according to the expectations of stakeholders \cite{kaur2022trustworthy}. With the rapid growth of AI applications, researchers demonstrated the urge to shift the focus to building trustworthy AI systems \cite{woitsch2021collaborative}. Therefore, trustworthy AI has been introduced in numerous application domains, such as self-driving cars \cite{karmakar2021assessing}, dentistry \cite{ma2022towards}, the architecture, engineering and construction (AEC) industry \cite{emaminejad2022trustworthy}, and education \cite{vincent2020trustworthy}. For example, in the context of self-driving cars, trustworthy AI ensures safe and reliable autonomous vehicle operation, promoting passenger and pedestrian safety. Similarly, the healthcare field also benefits from trustworthy AI systems that help in the precise diagnosis of diseases and the planning of treatments, ultimately improving patient care and treatment outcomes.

Given this, guidelines and specifications have been proposed to frame the limits and set the requirements for a system to be recognized as a trustworthy AI system. These guidelines establish a systematic framework for verifying and confirming the integrity of AI systems in different fields of application. In this section, we summarize the characteristics of trustworthy AI systems and highlight the robustness property and its significance in maintaining the trustworthiness of AI systems.

\subsubsection{Characteristics of Trustworthy AI systems}
\begin{figure}
    \centering
    \includegraphics[scale=0.32]{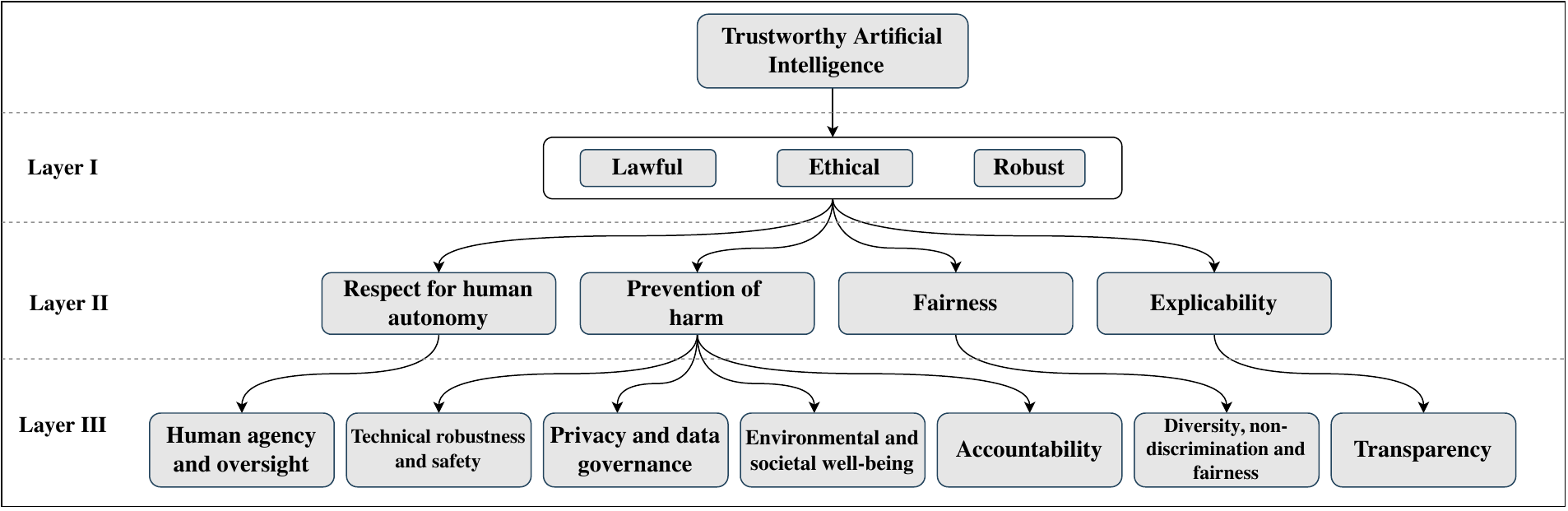}
    \caption{EU's requirement for trustworthy AI inspired by \cite{eu2019ethics, kaur2022trustworthy}.}
    \label{fig:TAI_req}
\end{figure}

The High-Level Expert Group on Artificial Intelligence (AI HLEG) was appointed by the European Commission to develop a guide document that describes the key specifications of trustworthy AI systems \cite{eu2019ethics}. The guidelines aim to provide the principles to which every AI system should adhere. Nevertheless, the guidelines document consists of three layers of abstraction, from the most abstract to the most concrete. At the conceptual level, according to EU guidelines, to make an AI system trustworthy, it should align three main components at the top level throughout its evolution; the components are summarized as follows: \cite{kaur2020requirements}:
\begin{enumerate}
    \item \textbf{Lawful}: it ensures that the AI system must comply with all applicable laws and regulations, such as the General Data Protection Regulation (GDPR).
    \item \textbf{Ethical}: it ensures that the development of AI systems is in accordance with the ethical values of human society.
    \item \textbf{Robust}: it ensures that the AI system is technically and socially reliable so that it does not generate harmful risks that AI systems can cause.
\end{enumerate}
The EU emphasized that the three components should be addressed during the development of AI systems and that the components should interact in deployment to provide a trustworthy service. To follow these guidelines, the document has identified four additional ethical principles for the development, deployment, and use of AI systems, which are: (a) respect for human autonomy, (b) prevention of harm, (c) fairness, and (d) explicability. These four principles are composed of seven fundamental requirements for trustworthy AI systems: (1) human agency and oversight, (2) technical robustness and safety, (3) privacy and data governance, (4) transparency, (5) diversity, non-discrimination, and fairness, (6) environmental and societal well-being, and (7) accountability. The EU requirements for trustworthy AI with different levels of abstraction are summarized in Fig. \ref{fig:TAI_req}.

\subsubsection{Robustness property of Trustworthy AI systems}
\label{sec:robustness}

Technical robustness is one of the imperative requirements for trustworthy AI systems, as discussed previously. The EU document related this requirement to the prevention of harm; in other words, the system should behave in a way that minimizes the associated risks and ensures the physical and mental integrity of humans. However, the technical details of the implementation of robust AI systems are not extensively discussed in the EU guidance document.
Meanwhile, another detailed guide document has recently been published that surveys the topics relevant to building trustworthy AI systems has recently been published by the International Organization for Standardization (ISO) and the International Electrotechnical Commission (IEC) \cite{ISO2020AI}. The technical report stated that the specification of trustworthy AI systems is out of the scope of the document; however, the report discussed related definitions and terminologies, threats and challenges, and mitigation measures for trustworthy AI systems. 

The ISO/IEC report has provided a general definition of robustness as “\textit{the ability of a system to maintain its level of performance under any condition}.” However, this general definition can be more concrete depending on the type of AI system. The robustness property generally ensures that the system can operate continuously according to its design. More specific definitions of robustness have been given in the ISO/IEC technical report depending on the functionality of the AI system, such as interpolation, classification, solving tasks or scoring. 

On the other hand, while accuracy has traditionally been a key metric for the evaluation of AI systems \cite{rozsa2016accuracy}, recent studies in ML \cite{tsipras2018robustness} and deep learning (DL) \cite{yang2020closer} have emphasized the importance of robustness as another critical measure that highlighted a significant trade-off with the accuracy. Specifically, robustness encompasses a broader set of properties than accuracy as it pertains to the system's ability to maintain functionality amidst various challenges and disturbances. In contrast, accuracy is primarily focused on the predictive performance of the models and serves as a means to achieve robust AI systems \cite{hamon2020robustness}. 

Regarding other system performance metrics, robustness can impact latency through additional checks, but optimization techniques such as parallel processing can help mitigate this effect \cite{wu2020collaborate}. However, as robust systems become more complex, interpretability may decrease, although explainable AI can serve to bridge this gap. Furthermore, highly robust systems may pose challenges in terms of maintenance, as they often involve complex error-handling mechanisms and configurations. Nevertheless, modular design principles and automation can help streamline maintenance processes and facilitate the management of robust systems over time \cite{spray2021building}.

Technically, robustness can also be recognized at both the data and algorithm levels \cite{li2021trustworthy}. Since the quality of ML solutions is highly dependent on input data, it is natural that data plays a major role in building robust AI systems. Therefore, data quality issues must be carefully investigated and formulated before implementing an AI solution. An important root of data quality issues is that, in real-life applications, data corruptions and inconsistencies are usually unanticipated. Another major issue is that the data will most likely follow changes during the evolution of the system \cite{thulasidasan2021effective}. Likewise, the robustness property must be considered at the algorithmic level, especially in high dimensions, where computations are relatively expensive \cite{diakonikolas2021robustness}. In most cases, the robustness of the algorithm in ML is coupled with the quality of the data, since it mainly implies the insusceptibility to data disturbances and the ability to generalize training inputs. 
% Sections \ref{sec:dataops} and \ref{sec:modelops} discuss in more detail the data and model issues that affect the robustness of deployed AI systems.

\makeatletter
\newcommand\footnoteref[1]{\protected@xdef\@thefnmark{\ref{#1}}\@footnotemark}
\makeatother

\begin{figure}
    \centering
    \includegraphics[scale=0.46]{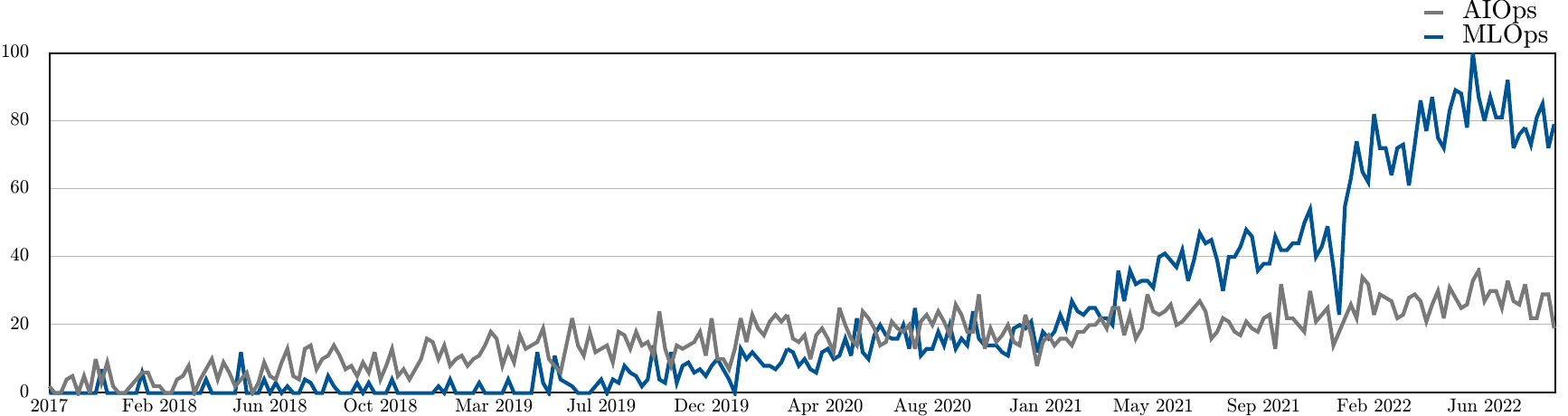}
    \caption{Google search trends \protect\footnotemark for AIOps and MLOps terms since 2017.}
    \label{fig:AIOps_MLOps}
\end{figure}
\footnotetext{\label{note1}https://trends.google.com/}

\color{blue}
\subsection{Machine Learning Operations (MLOps)}
\label{sec:mlops}
The field of MLOps is a relatively nascent topic that studies standardized methodologies and strategies for the deployment of ML solutions in real-world AI applications. In the literature, the term AIOps can also be found to describe ML systems in operations \cite{masood2019aiops}. However, according to Google Trends data\footnoteref{note1} and as shown in Fig. \ref{fig:AIOps_MLOps}, MLOps has started to be used more frequently than AIOps after 2021. MLOps is a heterogeneous discipline based on the application of DevOps principles to ML software systems \cite{john2021towards}.  MLOps provides a comprehensive framework for managing the entire lifecycle of ML models, including data acquisition, data preparation, model training, testing, evaluation, deployment, and continuous monitoring. This approach ensures efficient development and effective maintenance of ML systems in operational environments \cite{hewage2022machine}.

The MLOps lifecycle involves several key stages. It begins with ensuring high-quality, reliable data through the collection, cleaning, and preprocessing of raw data to make it suitable for model training. Next, machine learning models are developed and trained by selecting appropriate algorithms, tuning hyperparameters, and iterating to improve performance \cite{sarker2021machine}. Rigorous testing evaluates the model’s performance on unseen data to ensure generalization and meet accuracy and reliability standards. The trained model is then deployed into a production environment to make predictions on new data without disrupting existing services. Continuous monitoring of the model’s performance in production is essential to detect any degradation, requiring updates and retraining with new data to maintain consistent performance. Managing data versions and model artifacts is crucial for reproducibility and compliance, ensuring all steps in the ML pipeline are well-documented.

In MLOps, testing goes beyond traditional software approaches to specifically address the unique challenges of ML systems. This includes rigorous validation of data quality during collection and preprocessing to optimize model training \cite{zeller2024toward}.Unit testing examines specific components like data pipelines and model algorithms to guarantee consistent performance with different datasets. Integration testing validates the seamless interaction between these components throughout the ML lifecycle, from data ingestion to deployment \cite{medhat2020framework}. Additionally, MLOps emphasizes model explainability and robustness testing to enhance transparency and reliability against varying data conditions \cite{raj2021engineering}. By integrating these specialized testing methodologies, MLOps facilitates the deployment of reliable and scalable ML solutions that meet the complex demands of real-world AI applications.

\color{black}

\section{Specifications of robustness for MLOps systems}
\label{sec:spec}
\color{blue}
Generally speaking, robustness is a broad concept that covers a wide range of aspects and properties \cite{bradley1978robustness}. As stated in the ISO document \cite{ISO2020AI}, robustness is usually defined for a specific problem in a particular context or domain. In the context of MLOps, robustness and resilience are crucial but distinct concepts \cite{bhagoji2018enhancing}. Therefore, researchers from different fields have proposed a set of requirements to provide a robust formulation of their problems \cite{kitano2007towards, lusby2018survey}. Robustness focuses on the ML system's ability to maintain performance under varying and challenging data conditions, ensuring consistent and accurate predictions even with noisy or adversarial data. Resilience, however, pertains to the ML system's capacity to recover from failures or disruptions, ensuring the entire pipeline can continue operating or quickly resume normal functioning after issues such as hardware failures or software bugs. Our main focus is the robustness property, as it is crucial for guaranteeing the reliability and performance of models in real-world scenarios where data variability is common, ensuring the effectiveness of AI solutions in production environments \cite{steidl2023pipeline}. In this section, and based on an exhaustive literature search, we dissect the specifications of robust MLOps frameworks, which is a seminal part of building trustworthy AI systems in production environments.

\begin{figure}
    \centering
    \includegraphics[scale=0.45]{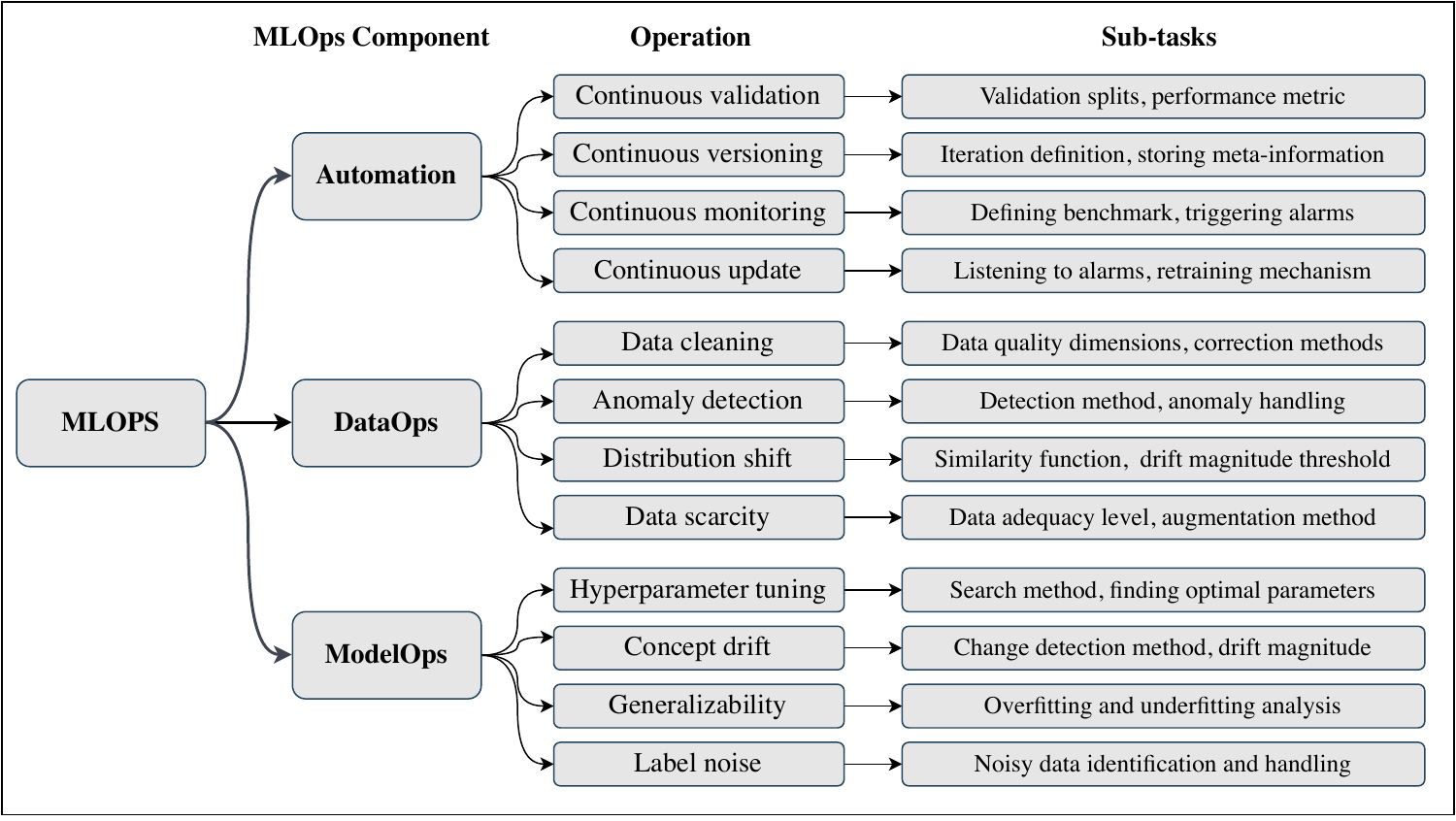}
    \caption{Summary of MLOps components, operations and the corresponding sub-tasks}
    \label{fig:MLOps_tasks}
\end{figure}

\color{black}
To build an overall robust MLOps system, robustness specifications should be addressed in all system processes. Practically speaking, to build a holistic robust ML system in production, we should address the issues and challenges that affect system robustness at both the data levels and the algorithmic levels, in addition to employing a robust automation process that drives the entire system. However, none of the existing frameworks addresses all aspects of building a robust system. Meanwhile, existing frameworks partially handle the robustness of ML systems; examples are: robustness to outliers \cite{ge2019aeromagnetic}, robustness to drift \cite{moulton2018clustering}, or robustness to noise \cite{hosseini2017google}. Therefore, we drill down to the level of granularity of the MLOps system ingredients, i.e., automation, DataOps, and ModelOps. The overall components and operations of the MLOps system, together with the corresponding subtasks, are summarized in Fig. \ref{fig:MLOps_tasks}.

Illustrating robustness with a concrete example of a traffic prediction system utilized in a navigation application. Robustness in this context refers to the system's ability to adapt to unforeseen situations in road conditions, such as accidents or construction, to deliver accurate and timely route recommendations for users. At the data level, challenges may arise from inconsistencies or inaccuracies in input data, such as missing or incorrect traffic flow information. Employing robust data processing techniques such as data cleansing and anomaly detection can ensure the overall validity of the input data. Furthermore, at the model level, robustness involves developing algorithms that are resistant to variations in traffic patterns and environmental factors. Therefore, robust ML models should generalize well across different road networks and traffic conditions, effectively mitigating the impact of data and algorithmic challenges on the accuracy of prediction. In this way, the system reduces the risk of providing inaccurate navigation guidance, promoting user satisfaction and safety during their journeys.

\subsection{Robustness in Automation}
MLOps systems are based on a continuous automated process that validates, monitors, and updates ML artifacts to maintain high performance during the evolution of the system \cite{spjuth2021machine}. To obtain elastic and reliable MLOps frameworks in production, a robust automation system should be implemented to allow rapid updates in the pipeline \cite{karamitsos2020applying}. To achieve that, MLOps leverages the principles of DevOps, CI/CD, which offer practical techniques to manage the service throughout its development stages \cite{virmani2015understanding}. These techniques ensure that elements of the ML system are regularly managed and updated to optimize performance \cite{garg2021continuous}. Since automation involves several steps, ranging from system build-up to system monitoring, the robustness aspect should be addressed at every step.

%%%% Bestoun 

\subsubsection{Robust Continuous Validation}
ML validation is a diagnostic step that precedes the deployment of the designed solution; it aims to verify the performance under conditions different from those used when building the solution \cite{vabalas2019machine}. During ML validation, the robustness of performance could be tested against bias towards training conditions \cite{ren2018learning}. Therefore, it governs the process of identifying potential solutions for real deployment environments by estimating performance under a wide range of problem settings \cite{larracy2021machine}. Furthermore, the implementation of a well-defined validation mechanism not only leads to more reproducible solutions \cite{kaur2022trustworthy} but also helps mitigate the risks associated with ML models in general, such as security concerns \cite{mcgraw2020top}, adversarial attacks \cite{song2019privacy}, and risks related to algorithms and data \cite{tan2022risks}.

The ML validation methodology generally uses several configurations, including dataset splits, performance evaluation metrics, and model hyperparameter settings \cite{wagstaff2012machine}. The hyperparameter settings of the model are considered an aspect of model robustness \cite{du2021understanding} and will be further investigated in a separate section. For the other configurations, there are several techniques and settings that can be found in the literature to develop a robust model evaluation methodology. In particular, how to select the validation datasets and performance metrics to evaluate the performance of the tested solution.

\begin{itemize}
    \item \textbf{Selection of Validation Split.} The factor that typically drives the choice of (a) suitable validation split (s) is the robustness aspect that is being tested during the validation phase. Therefore, to increase the robustness of the ML solution, the system should be thoroughly tested against various potential challenges, such as anomalies, drift, or adversarial attacks \cite{carlini2017towards}. Following that design, it is necessary to prepare curated validation splits that contain potential production data issues to thoroughly verify the robustness of the solution. Mostly, validation splits should include outliers, missing data, or distribution shifts. Then, the performance of the system is recorded across the validation splits. Traditionally, many techniques divide the dataset and achieve more robust systems \cite{ISO2020AI}. Cross-validation is the most common statistical method to estimate the performance of an ML model under specific conditions. In the literature, many techniques to perform cross-validation exist. Generally, K-fold cross-validation is the standard procedure for splitting the dataset. The procedure splits the dataset into $K$ subsets that are used to train and validate the model \cite{anguita2012k}. Some other cross-validation methods are single hold-out random subsampling and leave-one-out cross-validation \cite{berrar2019cross}. However, since building robust solutions requires a systematic way of selecting the splits, random subsampling would not necessarily guarantee the creation of robust splits. 
    % Furthermore, researchers have reinforced the classic cross-validation methods with improved strategies to deal with specific problems. For example, Sugiyama \textit{et al.} \cite{Sugiyama2007ICV} have proposed an importance-weighted cross-validation method and have shown that it provides more robust performance for covariate shift problems. Similarly, a customized version of the cross-validation method has been proposed to work in concept drift situations \cite{le2019automated}.

    \item \textbf{Selection of Performance Metric.} After selecting the validation split, the performance of the ML model in the validation split is statistically quantified. Typically, the learning task determines which performance metric to choose. For example, the most popular metrics in regression problems are empirically calculated based on the distance between the actual and predicted values \cite{botchkarev2018performance}. On the contrary, in classification, most performance metrics are empirically calculated based on the fraction of correctly classified predictions \cite{seliya2009study}. However, in some problems, a single performance metric does not necessarily provide a good indication of the robustness of the solution. For example, in class imbalance problems, the accuracy measure does not provide a reasonable estimate of the quality of the solution, and the F score, which is a combination of precision and recall measures, is a more explanatory indicator of performance \cite{guo2008class}. Another example is Cohen's kappa coefficient \cite{cohen1968weighted} for multilabel classification problems. However, there are other popular performance metrics, such as the area under the receiver operating characteristic curve (AUC-ROC), the confusion matrix, and the coefficient of determination $R^2$ \cite{erickson2021magician}. Therefore, it is essential to carefully select several performance metrics in the evaluation phase to capture robustness characteristics from different perspectives.
\end{itemize}

\subsubsection{Robust Continuous Versioning}
Building ML solutions in production is an iterative process that accompanies the development of the system throughout its life cycle \cite{chowdary2022accelerating}. The ML artifacts will follow changes over time, so it is fundamental to record the historical versions to obtain robust and tractable MLOps systems. Therefore, incorporating a centralized version control mechanism, a principle derived from continuous integration, is the first step after finalizing a releasable solution \cite{treveil2020introducing}. The main benefits of ML versioning are: reproducibility and traceability \cite{matsui2022mlops}. On the one hand, using a unique version number allows us to locate and trace the roots of the fault, and on the other hand, it also allows us to reuse a specific state of the system \cite{xin2018accelerating}. Continuous versioning requires an effective control mechanism for the governance of ML artifacts. A robust versioning mechanism must ensure that the different ML artifacts, such as configuration and hyperparameters, data and algorithm, have consistent version numbers for a specific iteration \cite{hohman2020understanding}.
Moreover, the versioning mechanism should clearly define what is considered an iteration. To determine the variations in the ML artifacts to uniquely identify a distinct iteration in the MLOps life cycle.

%%%%% Bestoun 13 October 
\subsubsection{Robust Continuous Monitoring}
\label{sec:robust-monit}
Lifelong monitoring of MLOps systems is a continuous process to collect feedback on production quality. Monitoring of the post-production model actively checks system performance and diagnoses various elements of the system to alert to abnormal behaviors \cite{breck2017ml}. Since the construction of MLOps systems involves the integration of several components into the system infrastructure, it is essential to establish a comprehensive monitoring strategy to cover these components. In production, model monitoring includes detecting degradation in both model performance and data quality, such as outliers, distributional shift, or concept drift \cite{klaise2020monitoring}. Monitoring requires a benchmark to compare the behavior of the current iteration with a pre-defined reference. For example, an evaluation is performed on the basis of the ground-truth labels for the production data to identify the degradation of the ML system's performance \cite{young2022empirical}. However, the ground-truth might be costly (sometimes impossible) to collect or unavailable immediately \cite{vzliobaite2010change}. Another example is monitoring the input data drift, which will be examined in more detail later, which involves analyzing the distribution of production data. Thus, a reference distribution should be available for comparison. 

\subsubsection{Robust Continuous update}
An update signal is triggered to refresh the learning system to maintain the robustness of the ML model's performance and keep them up to date with the newly arriving data. This step is activated mainly by feedback from model monitoring to extend the existing solution \cite{ISO2020AI}. Typically, the learning algorithm continuously uses the new output of the system to produce new versions of the ML model, known as the feedback loop. To update the model, two key considerations are identified in the literature \cite{schelter2018challenges}; the first is when to update the model. The second is how to execute the update. The first consideration focuses on deciding whether we need to update our learning pipeline. This is closely interlinked with model evaluation results; specifically, a threshold should be set, such as the level of performance degradation or the magnitude of distributional shifts. If these thresholds are exceeded, an update signal should be activated to update the ML artifact with the current context of the environment \cite{usuga2020machine}. A typical example is to re-train the model after following a change in the system, for instance, concept drift occurrence. The second consideration has to do with the update procedure of the model. There are many alternatives to update the model, such as performing re-training, adaptation, or replacement of the model \cite{raschka2019python}.

\subsection{Robustness in DataOps}
\label{sec:robdataops}
ML solutions are mainly based on data manipulation to infer patterns in the system \cite{mahdavinejad2018machine}. Data are pivotal for most of the phases of MLOps pipelines, as they are used to understand the characteristics of our problem and interpret the event episodes of our observations, in addition to practical use to train and validate the ML model. Due to this, robust data operations should be employed in the data artifact to increase the robustness of the overall MLOps system. DataOps is a new term coined collectively by data engineers, data scientists, and data analysts \cite{munappy2020ad}. DataOps refers to end-to-end data processing operations in production and is considered an integral part of the entire MLOps system \cite{atwal2020dataops}. The challenges of the DataOps pipeline can be roughly divided into two groups. The first group deals with traditional data quality issues that primarily affect the accuracy of the predictive learning task. The second group deals with data operationalization, which is primarily concerned with data handling \cite{capizzi2019devops}.

The e-commerce industry serves as an example of how robust DataOps practices are implemented in real-world scenarios to maintain operational efficiency and provide customized shopping experiences to customers. For example, within a large online retailer, ML systems analyze customer behavior and preferences to generate personalized product recommendations and targeted marketing strategies. Here, robust DataOps processes are key for managing extensive volumes of transactional data, customer interactions, and inventory records.
In such a context, the retailer aggregates and processes data from various sources, including website interactions, purchase histories, and demographic profiles. Robust DataOps practices ensure data quality and reliability by employing data cleansing techniques to eliminate inconsistencies and inaccuracies. Furthermore, ongoing monitoring of data quality helps detect anomalies or discrepancies, facilitating prompt corrective actions.

\subsubsection{Robust Data Cleaning}
Data quality is a cornerstone of any data-driven solution since it may increase/decrease the accuracy of the analytic results \cite{olson2003data}. Real-life applications, in which MLOps systems are used primarily, suffer from a high rate of erroneous data recorded from sensors that require cleaning and corrections. Therefore, most of the effort to build an ML solution is devoted to data cleaning to improve data quality \cite{domingos2012few}. Methods for processing data value errors can be roughly divided into two types: discarding erroneous data or cleaning erroneous data \cite{wang2019time}. The latter can also be divided into manual and automatic cleaning. Manual cleaning provides more accurate results, but is more expensive to implement \cite{wang2019time}. Data cleaning generally involves performing several activities to foster high levels of data reliability. These activities should be designed coherently to properly handle data corruption and imperfections. The main dimensions of data quality can be summarized as follows \cite{chu2016data}:

\begin{itemize}
    \item \textbf{Handling Missing and Duplicate Values.} 
    Missing values and duplicates are very common in real-world applications and handling them is the most basic data cleaning task \cite{ehrlinger2018treating}. In real-world scenarios, missing values can appear for many reasons, such as sensor failures in industrial problems or incomplete laboratory test results in clinical problems \cite{chao2019recent}. From a robustness perspective, missing data is problematic for several ML models and should be resolved. As an example, the classification and regression tree (CART) models handle the issue by using surrogate splits to mitigate the effect of missing values \cite{sugimoto2013comparison}. Traditionally, missing values are handled by discarding the tuple that includes incomplete data or replacing missing values with a statistically inferred value, known as imputation \cite{luo20183d}. Another issue of data quality is data duplication, or redundancy, which is a term that refers to the situation where some data records are repeated more than once in the dataset \cite{zhao2021impact}. Similarly to missing values, the duplication of data samples has a negative implication in learning, as it can skew the input distribution \cite{chowdhury2003data}, resulting in duplicate bias \cite{allamanis2019adverse}. To detect duplicate values, Euclidean distance algorithms and Dynamic Time Warping (DTW) are the most widely used techniques, with the latter providing more convenient results in the case of big data \cite{zhou2017machine}. Data fusion is the most widely used approach to correct data duplication \cite{meng2020survey}.

    \color{blue}
    \item \textbf{Anomaly Detection.} In the ML community, there is a large body of literature dedicated to studying the detection of extreme values in the dataset, known as anomaly detection. Anomalies are classified into three types: point, contextual, or collective anomalies \cite{chandola2009anomaly}. Extensive prior works in anomaly detection have explored various methodologies and applications across different fields, providing a foundational understanding that is directly applicable to MLOps. In addition to the negative effect that anomalies have on the performance of predictive systems \cite{ibidunmoye2015performance}, anomalies can indicate malicious activities or adversarial attacks \cite{ilyas2019data}, which can affect the robustness of the overall system. The integration of these established anomaly detection techniques is crucial in MLOps pipelines, where they serve to identify and mitigate abnormal behavior, thereby maintaining system integrity and performance. Thus, anomaly detection is an essential step in any MLOps pipeline to detect abnormal behavior in the system and alleviate associated problems.
    \color{black}
    
    \item \textbf{Domain-based Constraints.} Another step in the data cleaning phase is to deal with data samples that violate the restrictions on attribute values. In classical database management systems, the domain-based constraint is a type of integrity constraint that specifies the semantic rules of the data values that should follow \cite{qian1986knowledge}. In almost every application, there are specific rules for the types and values of the different features of the dataset. Simple examples of domain-based constraints are that ages cannot be negative or names cannot be numerical. Usually, domain experts specify the possible ranges and types of attribute values to accurately design a handling mechanism for inconsistent observations in the system \cite{park2021facilitating}.
\end{itemize}

\subsubsection{Robustness to Distribution Shift}
\label{sec:rob_dist_shi}
In production environments, the distribution of input data $P(X)$ often demonstrates distinct differences from that of training data due to the dynamic nature of post-deployment data \cite{zhou2022open}. For lifelong learning systems, the distribution change is an unresolved problem for predictive ML performance, as it tends to decline after the change \cite{zenke2017continual}. Data distribution is a key foundation for many ML algorithms \cite{WITTEN2017Probabilistic}. In deployment, MLOps systems should exhibit high robustness to distributional shifts; this requires systematic detection and adaptation mechanisms to cope with changes \cite{Dries2009AD}. Theoretically, distributional shifts may not be associated with a change in the target concept, known as virtual drift \cite{hinder2020towards}. However, in practice, the model should always be reviewed when there is a change in the data distribution \cite{Tsymbal2008Local}. The data used to train ML algorithms is assumed to be sampled from a distribution defined by an underlying generating function \cite{Hoens2011Imbalance}. As discussed in Section \ref{sec:robust-monit}, continuous monitoring strategies are deployed for a constant diagnosis of ML artifacts. Regarding distributional shifts, monitoring is generally performed by determining whether the data distribution of the present interval is significantly different from the past distribution using statistical tests \cite{Kawahara201sequential}. However, some parameters must be defined to establish a robust monitoring strategy, such as the window size to estimate the probability distribution and the change significance threshold, which are selected primarily manually. For adaptation, there exist different techniques that increase the robustness of ML models to distributional shifts, such as importance weighting and uncertainty estimation \cite{zhou2021bayesian}.

\subsubsection{Robustness to Data Scarcity}
Some real-world applications generate a few data points, especially in the medical sector and many industrial production processes \cite{zhou2022time}. Other related issues caused by data scarcity are applications with underreported data that cause the problem to be sparse \cite{dvorzak2016sparse}, or the problem of class imbalance, which is characterized by data-scarce regions of the underrepresented classes \cite{lyu2021synthesizing}. Therefore, the lack of historical data would hinder the development of a learning solution with good generalization performance. Inadequate training samples used to train the model in the training phase would lead to unexpected production behavior or bias towards the overrepresented population \cite{babbar2019data}. Therefore, data enrichment techniques, such as data augmentation, have been utilized to enhance data-scarce applications by generating more data samples based on the limited available data, or transfer learning by using pre-learned ML models and employing them in a similar task \cite{shafahi2019adversarially}. 
\color{blue}
Additionally, Generative Adversarial Networks (GANs) offer a promising approach to address data scarcity \cite{bansal2022systematic}. GANs are a type of deep learning model that can create entirely synthetic data closely resembling the real data distribution, allowing for significant data expansion. This is especially beneficial for scarce datasets, as demonstrated in applications like diagnosing COVID-19 from lung scans with limited real data \cite{ali2023leveraging}. Research has also explored the effectiveness of GANs in the energy sector for power consumption prediction and in defect detection \cite{xu2024scarcity}.
\color{black} 

\subsubsection{Robustness and Data Processing Resources.}
In information systems, the resources are not infinite. Any solution developed must be designed within the limits of stringent capacity constraints. In relation to deployment systems, data manifest the bottleneck in the MLOps pipeline \cite{ratner2017snorkel}. Undoubtedly, performing different data operations requires huge computing and storage resources that should typically be performed in real-time, which causes great difficulty for production systems \cite{diallo2012real}. As a result, the different stages of the DataOps pipelines should respect the trade-off between the complexity of the operations and the applicability of the solution online. However, not all solutions are suitable for deployment. More robust MLOps systems would require a large number of resources, which may not always be available \cite{alla2021mlops}. Therefore, to determine the pertinence and simplicity of the solution, the MLOps pipeline should be tested in an environment identical to that of deployment.

% There are additional aspects of data operations that can affect the overall performance of the system. For example, compliance with ethical rules and laws and regulations such as privacy and safety can influence the performance of ML models \cite{tom2020protecting}. In particular, adding more information about a customer may lead to better prediction, but at the same time, it may also violate the data disclosure agreement \cite{salim2020privacy}. Since these issues are more related to lawful and ethical principles, we chose to omit them because we focus on the implementation practices of ML systems in this research.

\subsection{Robustness in ModelOps}
\label{sec:robmodelops}
ML models in the wild must continually interact with the surrounding environment to acquire more knowledge and consequently improve their performance. Typically, the environment and its conditions are dynamic and, as the systems evolve, the MLOps systems should perform a set of operations to keep the ML models up-to-date \cite{dogan2021machine}. ModelOps is a general term used to describe operations in any type of model and not specifically for ML models \cite{hummer2019modelops}. Researchers used the term ModelOps interchangeably with MLOps when the field emerged in 2018 and 2019 \cite{treveil2020introducing}. Subsequently, the MLOps term was mainly used to describe the entire framework that handles ML systems in production. Therefore, to alleviate the confusion, we will refer to the set of operations that are performed on the learning task of the ML model as ModelOps, to follow the naming convention of the MLOps discipline. This clear detachment of data and model operations would facilitate the delineation of the different elements of the MLOps system. To build robust learning algorithms in the long run, the ModelOps pipeline should be designed to be well-performing and address potential challenges and issues.

In our online retail scenario, ModelOps ensures that ML models deployed in production can effectively adapt to seasonal variations and other unforeseen events typical in the retail industry. Alongside addressing seasonal shifts in customer behavior, these models must also handle sudden changes in market trends, promotions, and customer preferences. Robust ModelOps practices involve continuously monitoring model performance and regularly retraining the model to accommodate these dynamic factors. Additionally, robust ModelOps pipelines include strategies for model interpretation, helping stakeholders understand and trust the decisions made by the ML system. Furthermore, ensuring the generalizability of ML models across different customer segments and market conditions is imperative for their success in real-world retail environments. In this section, we thoroughly explore the specifications for building robust ModelOps pipelines; see Section \ref{sec:mlops}. We discuss related topics that impact the robustness of learning algorithms as found in the literature.

\subsubsection{Robust Hyperparameter Optimization}
Most learning algorithms require identifying optimal values for a set of hyperparameters that minimize the generalization error of a given function \cite{bergstra2012random}. Thornton \textit{et al.} \cite{thornton2013auto} tested 39 popular ML classifiers on 21 datasets with different hyperparameter settings. The results showed that, on average, the accuracy of the model changes by 46\% with the algorithm and the hyperparameter values. However, finding the best combination of hyperparameter values is not straightforward since it involves pre-defining several attributes, such as hyperparameter search methods and value ranges \cite{hertel2020sherpa}. The challenge of the hyperparameter tuning methods of the ML model in production is the search space, which is very large for most learning algorithms \cite{luo2016review}, especially complex algorithms such as neural networks \cite{feurer2019hyperparameter}. Therefore, researchers have called for robust and automatic hyperparameter optimization methods. This robust and automated mechanism must be addressed when designing the ModelOps pipeline. Robust hyperparameter optimization strategies would select the combination that produces consistent performance when used to evaluate both validation and test splits \cite{racca2021robust}.

\subsubsection{Robustness to Concept Drift}
Another considerable challenge for every lifelong learning system, and MLOps in particular, is the change in the probability distribution that represents the relationship between input data and target variables $P(y|X)$, known as concept drift \cite{Song2021SEGA}. It is worth noting that concept drift and distribution shift, are types of the general dataset shift phenomenon \cite{Candela2009DatasetS}. Concept drift is a relevant topic in data stream mining research, wherein the incoming data arrive sequentially with a unique timestamp \cite{agrahari2021concept}. Sequential order allows one to segment the learned concepts into periods of time to check for changes. Various reasons can cause concept drift in the system, such as machinery degradation, variations in customer behavior and demands, or adversarial attacks. Similarly to the robustness of the distributional shift presented in Section \ref{sec:rob_dist_shi}, the concept drift is first diagnosed and detected to activate an adaptation strategy \cite{Lu2016LearningUnder}. Typically, concept drift is detected by a degradation in the predictive performance of the model or a significant deviation in the distribution of input data, or hybrid monitoring for both \cite{Bayram2022Degradation}. However, the model is usually re-trained with the most recent data to provide the adaptation mechanism and more robust performance. Another approach is to utilize ensemble approaches to combine multiple results for the final output \cite{Elwell2011NSE}. ModelOps should find the balance between the complexity of the concept drift handling solution and the time required to execute the adaptation, as concept drift could have severe consequences for the overall system if not addressed quickly.

\subsubsection{Model Generalizability.}
One of the main properties that improve the robustness of an ML solution is the ability to generalize the pattern learned from training data to the entire domain of unseen distributions \cite{wang2022generalizing}. In ML research, the term generalizability is tied to underfitting and overfitting, which are generally used to analyze generalization performance. An optimal balance between underfitting and overfitting is required to achieve strong generalizability when building the ML solution \cite{ying2019overview}. In practice, ML models should generalize well using the available dataset without waiting to obtain large labeled datasets \cite{zhou2021domain}. Therefore, data augmentation procedures have been leveraged to improve the generalization of ML models \cite{zhou2021domain}. A principal approach to validate/improve generalization is to expand the training dataset by adding adversarial examples that can potentially be present in the production data. Other operations that can be found in the literature to handle the generalization of a learning model are cross-validation, regularization, and transfer learning \cite{li2021trustworthy}. However, the different operations should be integrated into ModelOps pipelines based on actively evaluating the generalizability of the ML model.

\subsubsection{Robustness to Label Noise.} 
ML models make decisions based on historical data used in the training phase. Therefore, any corruption in historical data would lead to poor deployment decisions. One type of corruption that may appear in the data is label noise, which affects the interpretation of the data \cite{garcia2015effect}. There are different sources of label noise, including incomplete information, measurement errors, or incorrect labels in the data acquisition phase \cite{koziarski2020combined}. The removal of noisy labels is not always feasible, as it is difficult to determine whether the labels are indeed noisy \cite{garcia2015effect}. For example, distinguishing between noisy and correct data records in duplicate records assigned to different target variables in the same dataset is usually challenging or costly \cite{xiao2015learning}. Therefore, the implementation of noise-tolerant strategies in learning algorithms is widely adopted to build solutions that are robust to noise. Researchers have developed model-based methods in noisy environments to obtain noise-tolerant versions of conventional ML learning algorithms. Probabilistic methods are also effective in dealing with label noise, such as Bayesian or clustering methods \cite{frenay2013classification}. These modifications to the learning algorithms should be addressed while building ModelOps pipelines.

% quately address data- and model-related issues for achieving robustness {Tripathi, S., Muhr, D., Brunner, M., Jodlbauer, H., Dehmer, M., & Emmert-Streib, F. (2021). Ensuring the robustness and reliability of data-driven knowledge discovery models in production and manufacturing. Frontiers in artificial intelligence, 4, 22.}
% it is out of scope for ISO document
% quantifies trust
% 1. https://ieeexplore.ieee.org/stamp/stamp.jsp?arnumber=8610145
% 2. https://www.sciencedirect.com/science/article/pii/S0950584919302484
% 3. https://www.mdpi.com/2073-8994/14/3/628
% 4. https://www.sciencedirect.com/science/article/pii/S0020025511005354
% 5. https://ieeexplore.ieee.org/stamp/stamp.jsp?arnumber=9133387
% 6. https://citeseerx.ist.psu.edu/viewdoc/download?doi=10.1.1.847.8172&rep=rep1&type=pdf

A final remark to note is that operations in the different stages of automation management, DataOps, and ModelOps, usually overlap with each other rather than being separate components of linear pipelines. This is why we can view automation processing as the engine of MLOps systems, which mainly distinguishes it from the classical offline frameworks for ML projects. Essentially, MLOps automation management controls and connects different operations in the life cycle of the ML system by activating a specific stage based on the results of another stage. Therefore, it can be run as a standalone service. Using the automated mechanism, the costs of human effort are minimal, and the workflow is more reliable and carried out faster.

\section{Literature Review on Robust ML Approaches}

\label{sec:acd_surv}
To compile the latest approaches devoted to developing ML systems that address the robustness property in deployment, we conduct an extensive literature search for existing solutions that handle one or more robustness aspects in production that have been proposed in the research community. We begin by reviewing existing surveys that are similar in scope to our work. We then
% Existing solutions that handle one or more robustness aspects in production are broadly reviewed in this section through an extensive literature search. 
The methods are organized into three levels: automation management, data, and algorithmic. Specifically, we survey the methods that are concerned with the robustness aspects that we presented in Section \ref{sec:spec}. The retrieved approaches are organized according to the robustness aspect addressed in Table \ref{tab:robust_refs}. 

\begin{table}
\centering
\caption{Summary of robust ML approaches in production}
\label{tab:robust_refs}
\begin{tabular}{l|l|l} 
\toprule
\textbf{MLOps Component}              & {\textbf{Operation}} & \textbf{References}  \\ 
\hline
\multirow{4}{*}{\textbf{Automation }} & Continuous validation                   &       \cite{li2020cms, shahoud2019facilitating, raj2021edge}              \\
                                      & Continuous versioning                  &        \cite{zarate2022k2e, van2017versioning, vartak2015supporting}              \\
                                      & Continuous monitoring                   &         \cite{crankshaw2014missing, silva2020benchmarking, raj2021edge}             \\
                                      & Continuous update                  &            \cite{wu2020priu, wu2020deltagrad, semola2022continual, huang2021modelci}         \\ 
\hline
\multirow{4}{*}{\textbf{DataOps}}     & Robust data cleaning                           &           \cite{karlavs2020nearest, schelter2018automating, breck2019data, yu2022dataops, shrivastava2020dqlearn}            \\
                                      & Robustness to distribution shift                       &           \cite{greco2021drift, soin2022chexstray, bayram2022drift, Chen2021Mandoline, chen2021did, breck2017ml}             \\
                                      & Robustness to data scarcity                     &           \cite{re2019overton, ratner2017snorkel, ESR2022}    \\
                                      & Resource scheduling                       &               \cite{min2021sensix++, antonini2022tiny} \\ 
\hline
\multirow{4}{*}{\textbf{ModelOps}}    & Robust hyperparameter tuning                   &       \cite{kurian2021boat, akiba2019optuna, meister2020maggy}              \\
                                      & Robustness to concept drift                           &               \cite{vieira2021driftage, yang2021cade, mallick2022matchmaker}           \\
                                      & Generalizability                        &               \cite{ho2020extensions, zhang2020generalizing, hansen2021generalization}    \\
                                      & Robustness to label noise                             &               \cite{gudovskiy2021autodo, mathew2021defraudnet, renggli2020ease}        \\
\bottomrule
\end{tabular}
\end{table}

\subsection{Related Surveys}

Several works have been published to characterize the trustworthiness of AI systems and their different properties. Kaur \textit{et al.} \cite{kaur2020requirements} reviewed trustworthy AI approaches based on EU guidelines. The authors dissected some principles and suggestions for building trustworthy AI systems. However, the article mainly discusses the mechanisms of interactions between humans and machines to improve the performance of AI systems. In another article \cite{chatila2021trustworthy}, Chatila \textit{et al.} also discussed the requirements to design trustworthy AI systems. The article discussed the alignment of the deployment and use of such systems with human rights and values. A more detailed article by Kaur \textit{et al.} \cite{kaur2022trustworthy} presents a comprehensive overview of trustworthy AI systems. The article discussed existing strategies to standardize trustworthy AI systems. Furthermore, the article discussed approaches to mitigate AI risks by involving users and society to enrich AI systems. However, the article did not discuss in technical detail how to implement trustworthy AI systems.

A recent article published by Wing \cite{wing2021trustworthy} supports the trustworthiness of AI systems by benefiting from other communities. The article discusses the advantages of bringing together researchers in different disciplines: trustworthy computing, formal methods, and AI. The author showed that connecting research communities across industry, government, and academia is crucial to establishing trustworthy AI. Similarly, Kumar \textit{et al.} have discussed the requirements of trustworthy AI systems in pervasive computing and big data phenomena \cite{kumar2020trustworthy}. However, the authors reviewed the ethical issues of algorithms and data and provided a real-world application in smart cities.

From a computational perspective, a comprehensive survey has been presented to guide practitioners in achieving trustworthy AI \cite{liu2021trustworthy}. The survey article provided a broad taxonomy of the field and recent advances. In particular, the article summarized the application of trustworthy AI systems in real-world applications. Furthermore, the article has reviewed the interactions among the different aspects of trustworthy AI. In a relevant work, Li \textit{et al.} \cite{li2021trustworthy} have followed a systematic approach to introduce the aspects of trustworthy AI from theoretical and operational perspectives. The authors have also presented guidelines for the different stakeholders in the final product to construct trustworthy AI solutions. However, the article did not merely review existing approaches to build robust AI systems in production.

For MLOps, a few papers have reviewed the current state-of-the-art. The central part of the papers focused on current trends and challenges in building MLOps systems. For example, Kreuzberger \textit{et al.} \cite{kreuzberger2022machine} have highlighted the open challenges in the field and presented an overview of the MLOps architecture. In a review paper, Zhao also discussed the current trends and challenges of MLOps systems \cite{zhao2021machine}. In particular, the paper has focused on applying DevOps principles to ML projects and the additional components needed to satisfy the requirements of MLOps systems. The interplay between AutoML and MLOps has been investigated by Symeonidis \textit{et al.} \cite{symeonidis2022mlops}. In addition to reviewing the current challenges, the authors have proposed an approach based on the incorporation of explainability and sustainability to increase the robustness of MLOps systems. Similarly, Tamburri has described the current challenges in the implementation of sustainability and interpretability in MLOps frameworks \cite{tamburri2020sustainable}. Meanwhile, Testi \textit{et al.} \cite{testi2022mlops} presented a recent taxonomy of MLOps in an overview paper. The paper has defined and standardized the methodologies for building MLOps frameworks. In addition, the paper has also proposed an operationalized approach to building ML systems. In terms of the stages and activities involved in the adoption of MLOps, a systematic review of the literature has described the details of the development of ML systems in operation \cite{john2021towards}. Furthermore, the authors have presented a maturity model that should be followed to evolve MLOps practices. The importance of MLOps in data science has been investigated by M{\"a}kinen \textit{et al.} \cite{makinen2021needs}. The authors have collected responses from 331 professionals from 63 countries working in the ML domain. The question was to investigate whether professionals tend to build data-centric or model-centric solutions, or both. The study found that the system should integrate both solutions in the construction of the MLOps pipeline.

\subsection{Robust Approaches in Automation}
Unlike conventional ML schemes in which algorithms work offline, in production, ML solutions are required to leverage the available knowledge to solve new problems continuously and automatically throughout their lifespan. Automation management involves several operations, including continuous validation, continuous versioning, continuous monitoring, and continuous update. This section surveys current methods that address these operations in the literature. The overall approaches are summarized in terms of their pros and cons in Table \ref{tab:aut_approaches}.

\begin{table}
\centering
\caption{Summary of robust approaches in automation: Pros and cons}
\label{tab:aut_approaches}
\resizebox{\linewidth}{!}{%
\begin{tabular}{@{}llll@{}}
\toprule
\textbf{Operation} & \textbf{Reference} & \textbf{Pros} & \textbf{Cons} \\ \midrule

\multirow{4}{*}{\begin{tabular}[c]{@{}l@{}}\textbf{Continuous} \\ \textbf{validation}\end{tabular}}  & \cite{li2020cms} & No manual intervention needed & Update timing is not dynamic \\
& \cite{shahoud2019facilitating} & Configurable user interface & Not fully automated \\
& \begin{tabular}[l]{@{}l@{}}\cite{raj2021edge} \\  ~ \end{tabular} & \begin{tabular}[l]{@{}l@{}}Fully automated and scalable, \\  language and library agnostic\end{tabular}& \begin{tabular}[l]{@{}l@{}}Security and privacy issues \\are not addressed \end{tabular}\\ \midrule

\multirow{3}{*}{\begin{tabular}[c]{@{}l@{}}\textbf{Continuous} \\ \textbf{versioning}\end{tabular}}  & \cite{zarate2022k2e} & Versioning models and data & Cannot integrate external datasets \\
& \begin{tabular}[l]{@{}l@{}}\cite{van2017versioning}   \\ ~ \end{tabular}& \begin{tabular}[l]{@{}l@{}}Versioning multiple pipelines \\simultaneously \end{tabular}& \begin{tabular}[l]{@{}l@{}} Manual setting of the code \\ version \end{tabular}\\
& \cite{vartak2015supporting} & Fast iterative ML modeling & Still in the initial implementation\\ \midrule

\multirow{3}{*}{\begin{tabular}[c]{@{}l@{}}\textbf{Continuous} \\ \textbf{monitoring}\end{tabular}} & \begin{tabular}[l]{@{}l@{}} \cite{crankshaw2014missing} \\ ~ \end{tabular}& \begin{tabular}[l]{@{}l@{}} Scalable and user-friendly \\access with low latency \end{tabular}& \begin{tabular}[l]{@{}l@{}}Supports limited learning \\ tasks \end{tabular}\\
& \cite{silva2020benchmarking} & Front and back ends monitoring & Limited profiling capabilities \\
& \cite{raj2021edge} & Real-time processing & Not tested in the long run \\ \midrule
\multirow{6}{*}{\begin{tabular}[c]{@{}l@{}}\textbf{Continuous} \\ \textbf{update}\end{tabular}} & \cite{wu2020priu} & No accuracy loss & Only tested on limited ML models\\
& \cite{wu2020deltagrad} & Very rapid retraining & Limited to specific types of models \\
& \begin{tabular}[l]{@{}l@{}} \cite{semola2022continual} \\ ~ \end{tabular} & \begin{tabular}[l]{@{}l@{}} User-friendly for on-demand \\usage in edge and cloud systems \end{tabular} & \begin{tabular}[l]{@{}l@{}}Limited investigation of real-world\\ use cases \end{tabular} \\

& \begin{tabular}[l]{@{}l@{}}\cite{huang2021modelci} \\ ~ \end{tabular}& \begin{tabular}[l]{@{}l@{}} Collaborative management through \\a web interface \end{tabular} & \begin{tabular}[l]{@{}l@{}} Inference job underutilizes the  \\ allocated GPUs \end{tabular} \\ \bottomrule
\end{tabular}%
}
\end{table}

\subsubsection{Continuous Validation Approaches.} Li and Gui proposed a container-based Continuous Machine Learning and Serving (CMS) platform \cite{li2020cms}. CMS platform incorporates offline and online stages for model validation to assess the robustness of the ML algorithm before using it in production. While it separates validation stages effectively, it lacks dynamic update timing and relies heavily on a pre-defined Mean Absolute Error (MAE) threshold. In offline validation, 20\% of the dataset is used to test the model, the quality of the prediction is evaluated by calculating the Mean Absolute Error (MAE). Subsequently, a new model is introduced to the system if its MAE is less than the current model's error. Once a model passes the offline validation, the online validation is performed to evaluate the model in live data. A pre-defined MAE threshold is then used to evaluate the model's performance during an arbitrary time period. In another approach, Shahoud \textit{et al.} have proposed a framework for managing ML tasks in big data environments \cite{shahoud2019facilitating}. The ML application is based on a microservice-based architecture that supports a scalable solution. Model validation is performed using 20\% of the total dataset. To evaluate the performance of the ML model, the mean of the performance metric is calculated for each experiment that is repeated three times. To verify the efficiency of the solution in production, the authors focused on the execution time and overhead metrics of the framework of the different forecasting algorithms rather than prediction accuracy. For artificial intelligence (AIoT) applications, Raj \textit{et al.} \cite{raj2021edge} have proposed a new robust automated framework for edge machine learning operations (Edge MLOps). The results of the root mean square error (RMSE) were validated by 10-fold cross-validation splits to select the best ML model of four candidate models. However, integrating additional performance metrics and considering cross-validation techniques could enhance its robustness in model selection.

\subsubsection{Continuous Versioning Approaches} Knowledge to Environment (K2E) platform was presented to govern data and model catalogues \cite{zarate2022k2e}. K2E architecture includes a version control manager that stores the meta-information of the various versions of data and models in the system. The meta-information incorporates the version numbers and pointers to each version's location in the internal datalake. The K2E platform offers efficient version control but lacks integration with external datasets, which limits its versatility. Weide \textit{et al.} \cite{van2017versioning} have presented another versioning scheme. The scheme uses semantic versioning to distinguish between the types of change that occurred in the ML pipeline, which can offer a very useful interpretation of the evolution of the system. In addition, the scheme allows multiple versions to run simultaneously without conflict, but managing multiple simultaneous versions may pose additional complexity. For iterative model building, the SHERLOCK system was built that enables fast iterative modeling \cite{vartak2015supporting}. The system integrates a central repository, ModelDB, that stores all the meta-information about the model in tables. Meta-information includes model specification, model performance, and results of the different models. The meta-information is used later on to compare the models or reuse particular results.

\subsubsection{Continuous Monitoring Approaches} Velox system has been developed for a scalable model serving pipelines in a low-latency environment \cite{crankshaw2014missing}. Velox automatically monitors the quality of the model evaluated by selected performance metrics. Additionally, Velox keeps track of the errors associated with each ML model. Monitoring strategies allow the system to continuously verify the generalizability of the model and send re-training signals. However, Velox lacks comprehensive profiling capabilities. In another work, Silva \textit{et al.} presented a monitoring and benchmarking environment that can monitor both the prediction quality and the system resources \cite{silva2020benchmarking}. In addition to prediction performance, the approach is designed to monitor five metrics of resource consumption: completion time, CPU and GPU usage, memory usage, disk input/output (IO) operations, and network traffic. Edge MLOps \cite{raj2021edge} also monitors significant RMSE deviations from the performance of the ML model. A fixed threshold value is empirically defined to be used as a benchmark value to compare the current RMSE metric of the deployed ML model. However, relying solely on this fixed threshold may limit its effectiveness in performance management.

\subsubsection{Continuous Update Approaches} A Provenance-Based Approach for Incrementally Updating Regression Models (PrIU), and its optimized version PrIU-opt, have been proposed to incrementally update the model parameters while preserving prediction performance, but are limited in their testing on specific models \cite{wu2020priu}. The approaches use gradient descent and its variants to trace the provenance of input data and, consequently, update the model parameters. Similarly, DeltaGrad was proposed for the fast re-training of ML models based on (stochastic) gradient descent \cite{wu2020deltagrad}. DeltaGrad algorithm is inspired by ideas from the Quasi-Newton methods \cite{byrd1994representations} and can be used to update ML models in production. In addition, in the case of changes in the data, the algorithm can delete the old data and learn from the newly acquired data. Continual-Learning-as-a-Service (CLaaS) paradigm was designed to address the continual learning of ML models in production \cite{semola2022continual}. Instead of re-training the ML model from scratch, CLaaS integrates several continual learning approaches that update the models upon signals that indicate drastic changes in the distribution of the data and the performance of the ML model. In a related approach, the ModelCI-e (continuous integration and evolution) plugin was designed to easily integrate with existing serving systems \cite{huang2021modelci}. In the online phase, the system automatically checks for significant drifts in the data distribution and invokes an appropriate continual learning method if the drift magnitude exceeds a certain threshold. Nevertheless, the current approach does not fully utilize the allocated GPUs for inference tasks. Optimizing resource allocation and implementing dynamic load balancing strategies could improve resource utilization efficiency.

\subsection{Robust Approaches in DataOps}
As discussed in Section \ref{sec:robdataops}, DataOps involves performing several operations, including data cleaning, ensuring the robustness of the entire ML system to data corruptions such as distributional shifts and imbalance datasets, in addition to robustness to data scarcity. In this section, we review existing solutions that address the robustness aspects of DataOps in ML solutions in production. The advantages and disadvantages of the different approaches are presented in Table \ref{tab:dataops_approaches}

\begin{table}
\centering
\caption{Summary of robust approaches in DataOps: Pros and cons}
\label{tab:dataops_approaches}
\resizebox{\linewidth}{!}{
\begin{tabular}{@{}llll@{}}
\toprule
{\textbf{Operation}} & \textbf{Reference} & \textbf{Pros}  & \textbf{Cons} \\
\hline
\multirow{6}{*}{\begin{tabular}[c]{@{}l@{}}\textbf{Data} \\ \textbf{cleaning}\end{tabular}} & \cite{karlavs2020nearest} & Handling missing data efficiently & Requires some manual efforts\\
& \cite{schelter2018automating} & Scalable to large datasets & Not validated in streaming scenarios\\
& \begin{tabular}[l]{@{}l@{}}\cite{breck2019data} \\ ~ \end{tabular}& \begin{tabular}[l]{@{}l@{}} Supports single-batch and inter-\\batch validation \end{tabular} & \begin{tabular}[l]{@{}l@{}} System's implementation influenced \\ by Google's infrastructure   \end{tabular} \\
& \begin{tabular}[l]{@{}l@{}}\cite{yu2022dataops} \\ ~ \end{tabular}& \begin{tabular}[l]{@{}l@{}} User-friendly platform without\\ the need for coding \end{tabular} & \begin{tabular}[l]{@{}l@{}} Lower proficiency in programming\\ may impact the task efficiency   \end{tabular}\\
& \cite{shrivastava2020dqlearn} & User-configurable flexibility & Potential information loss \\
\hline
\multirow{8}{*}{\begin{tabular}[c]{@{}l@{}}\textbf{Distribution} \\ \textbf{shift}\end{tabular}} & \cite{greco2021drift} & Real-time processing & Prone to false positives\\
& \begin{tabular}[l]{@{}l@{}}\cite{soin2022chexstray} \\ ~ \end{tabular} & \begin{tabular}[l]{@{}l@{}} VAE-based detection of changes \\ without ground-truth labels \end{tabular} & \begin{tabular}[l]{@{}l@{}} Evaluated using use-cases not fully\\ representing real-word scenarios \end{tabular}\\
& \cite{bayram2022drift} & Fully automated and scalable & Supports limited ML models\\
& \begin{tabular}[l]{@{}l@{}}\cite{Chen2021Mandoline} \\ ~ \end{tabular}& \begin{tabular}[l]{@{}l@{}}Can handle complex and high-\\ dimensional features  \end{tabular}& \begin{tabular}[l]{@{}l@{}}Only focused on image and textual\\   data\end{tabular}\\
& \cite{chen2021did} & Low computation cost  & Limited to classification tasks\\
& \cite{breck2017ml} & Rapid, low-latency inference & Only applicable to certain tasks\\
\hline
\multirow{5}{*}{\begin{tabular}[c]{@{}l@{}}\textbf{Data} \\ \textbf{scarcity}\end{tabular}} & \begin{tabular}[l]{@{}l@{}}\cite{re2019overton} \\ ~ \end{tabular}&  \begin{tabular}[l]{@{}l@{}}Support multitask learning and a \\code-free approach \end{tabular}& \begin{tabular}[l]{@{}l@{}}Does not have support for model \\versioning \end{tabular}\\
& \begin{tabular}[l]{@{}l@{}}\cite{ratner2017snorkel}  \\ ~ \end{tabular} & \begin{tabular}[l]{@{}l@{}} Reduces the cost and difficulty of \\ training ML models \end{tabular} & \begin{tabular}[l]{@{}l@{}} Requires large amount of unlabeled \\data\end{tabular} \\
& \begin{tabular}[l]{@{}l@{}}\cite{ESR2022} \\ ~ \end{tabular}& \begin{tabular}[l]{@{}l@{}}Uses domain knowledge to  \\generate synthetic data \end{tabular}&\begin{tabular}[l]{@{}l@{}} Was evaluated using simple DL \\  architectures\end{tabular}\\
\hline
\multirow{3}{*}{\begin{tabular}[c]{@{}l@{}}\textbf{Resource} \\ \textbf{scheduling}\end{tabular}} & \begin{tabular}[l]{@{}l@{}}\cite{min2021sensix++}\\ ~ \end{tabular} & \begin{tabular}[l]{@{}l@{}}Efficiently serves multiple models  \\ on edge devices architectures\end{tabular}& \begin{tabular}[l]{@{}l@{}}Limited concurrent access to Coral\\ TPUs from different processors \end{tabular}\\
& \begin{tabular}[l]{@{}l@{}}\cite{antonini2022tiny}\\ ~ \end{tabular} & \begin{tabular}[l]{@{}l@{}}Operates on resource-constrained \\ IoT sensors \end{tabular}&  \begin{tabular}[l]{@{}l@{}}Long deployment time for new ML \\ models \end{tabular}\\
\bottomrule
\end{tabular}}
\end{table}

\subsubsection{Data Cleaning Approaches}
CPClean approach was proposed to handle datasets with missing values to train ML models \cite{karlavs2020nearest}. The approach uses sequential information maximization to minimize conditional entropy to find the best cleaning strategy. For missing value imputation, CPClean was tested using both simple imputation methods and ML-based imputation methods. However, CPClean requires manual efforts, which could be a bottleneck in large-scale applications. In a different approach, Schelter \textit{et al.} have developed an automated data quality verification platform \cite{schelter2018automating}. The platform integrates an incremental validation methodology of data quality for evolving datasets. In addition, the platform also supports anomaly detection using statistics from historical data quality metrics, , but real-time capabilities should be enhanced. Similarly, Breck \textit{et al.} designed an automated data validation system in the context of ML \cite{breck2019data}. Single-batch validation is performed to analyze deviations from the expected states of data samples within a batch to detect instability in data characteristics. Recently, the DataOps-4G platform was proposed for both specialists and generalists, but it may face limitations in handling advanced data cleaning tasks due to limited programming knowledge. DataOps-4G enables the discovery of data quality issues without the need to write code \cite{yu2022dataops}. The platform handled five data quality issues including missing, non-unique, duplicate, imprecise, and inconsistent values. The platform can be integrated into ML production environments by bundling functionalities and codes into DataOps pipelines. Another toolkit was built to handle data quality issues, namely DQLearn \cite{shrivastava2020dqlearn}.

\subsubsection{Robust Approaches to Distribution Shift}
In a recent study, DRIFT LENS, a real-time unsupervised drift detection approach has been designed to identify and cope with changes in the system using MLOps pipelines \cite{greco2021drift}. DRIFT LENS drift processing includes offline and online phases, but may be prone to false positives, which could impact decision-making accuracy. In the offline phase, baseline data is extracted to represent the learned concept during training. Baseline data are used to compare the distribution of production data in the online phase to detect changes. Changes are detected on the basis of a comparison with a drift-magnitude threshold value. For medical imaging AI applications, CheXstray has been proposed to achieve unsupervised real-time drift monitoring \cite{soin2022chexstray}. For shift detection, the approach employs statistical tests to compare the distribution of a reference dataset with the detection window. In the detection phase, for numerical features, the Kolmogorov-Smirnov (K-S) test is used to measure the distribution shift. In contrast, the chi-square goodness of fit test is used for categorical variables. For scalable industrial processes, an automated and adaptive approach has been proposed to handle distribution changes \cite{bayram2022drift}. The approach uses a covariate shift adaptation technique, namely importance weighting \cite{Sugiyama2012IntroductiontoCS}, which adjusts the data distribution of the production data samples to match the data distribution of the training datasets. Additionally, the approach has also been used for the rapid integration of new machinery into industrial systems. Yet, the approach may be limited to specific ML models and requires further testing across diverse domains to assess its robustness. Similarly, the MANDOLINE framework has been proposed to evaluate ML models under distribution shift \cite{Chen2021Mandoline}. MANDOLINE uses user-specified slicing functions to capture the distribution shift. The slicing functions are also used to find importance weighting estimates using the Kullback-Leibler importance estimation procedure (KLIEP) \cite{Sugiyama2007KLIEP}. In a different approach, the MASA sampling algorithm has been developed to detect the distribution shift \cite{chen2021did}. MASA partitions the dataset into several clusters to estimate changes in ML models using the confusion matrix that is used to evaluate ML performance. This process is repeated until it reaches the stopping rule or the maximum number of iterations. However, it could miss subtle distribution shifts and is limited to classification tasks, potentially restricting its utility in broader contexts. Another system proposed by Breck \textit{et al.} is capable of tracking and identifying distribution shifts between different batches of data \cite{breck2017ml}. 

\subsubsection{Robust Approaches to Data Scarcity}
Overton system has been developed to help engineers automate the ML life cycle by providing tools to maintain the AI application \cite{re2019overton}. To improve existing data, Overton provides a feature to augment the data by creating synthetic data. The system also supports cold-start use cases where there are no training data available to train the ML model. However, Overton lacks model versioning capabilities. In a different study, the Snorkel system was developed to create labeled datasets using labeling functions \cite{ratner2017snorkel}. The system can quickly create training data sets by combining sources of weak supervision. Each data point will be annotated with a probabilistic label that will eventually be used by discriminative ML models. For industrial applications with data scarcity, a data augmentation method has been proposed to generate a new set of augmented data points \cite{ESR2022}. The method first uses domain knowledge to decompose the data and then augments the dataset with random and fresh samples, but requires further validation using complex models to ensure its effectiveness across diverse applications.

\subsubsection{Resource Scheduling Approaches} SensiX++ system has been developed to incorporate MLOps capabilities with edge devices \cite{min2021sensix++}. The system features several components to manage and orchestrate ML solutions in production. The components provide system-wide orchestration and allow the automation and optimization of the data operations to serve multiple models simultaneously. An adaptive scheduler is also incorporated to manage the system resources and inform the input data requirement of the ML models. However, it may require improvements in concurrent resource access. Another new framework used to orchestrate the MLOps system at the edge of IoT systems is Tiny-MLOps \cite{antonini2022tiny}. The framework includes several features to deliver adaptation and evolving capabilities to devices with limited resources. Tiny-MLOps is able to orchestrate the entire life cycle of the MLOps system that executes a specific ML-based task, but faces challenges related to long deployment times for new ML models.

\subsection{Robust Approaches in ModelOps}
ML models in production systems continuously evolve and acquire new knowledge. As explained in Section \ref{sec:robmodelops}, various issues should be considered in ML to retain robustness, including hyperparameter optimization, concept drift, generalizability, and label noise. In this section, we examine the literature and survey the approaches that handle the robustness aspects of ModelOps. The pros and cons of the various approaches are outlined in Table \ref{tab:modelops_approaches} to provide a comprehensive summary.

\subsubsection{Hyperparameter Optimization Approaches} A Bayesian Optimization AutoML Time-series (BOAT) framework was integrated with the MLOps system to automatically select the best hyperparameter configurations \cite{kurian2021boat}, but its applicability is limited to univariate time-series data. The framework employs a Monte Carlo Tree Search-based (MCTS) Bayesian optimizer that can efficiently find the optimal hyperparameter configuration in conditional and high-dimensional spaces. Another popular framework for hyperparameter optimization is Optuna \cite{akiba2019optuna}. Optuna has the define-by-run feature that allows users to construct the hyperparameter search space dynamically, yet it lacks support for external optimizers. The framework uses a combination of efficient sampling and pruning algorithms to optimize the optimal hyperparameter search. In a different study, the MAGGY framework was introduced for asynchronous parallel hyperparameter search \cite{meister2020maggy}. The framework incorporates several hyperparameter optimization techniques, Bayesian optimization, and Gaussian processes, in addition to user-defined optimizers. MAGGY achieves asynchrony using an early stopping mechanism. However, the framework was evaluated on a single use case, raising concerns about its generalizability.
    
\begin{table}
\centering
\caption{Summary of robust approaches in ModelOps: Pros and cons}
\label{tab:modelops_approaches}
\resizebox{\linewidth}{!}{
\begin{tabular}{@{}llll@{}}
\toprule
{\textbf{Operation}} & \textbf{Reference} & \textbf{Pros}  & \textbf{Cons} \\
\hline
\multirow{5}{*}{\begin{tabular}[c]{@{}l@{}}\textbf{Hyperparameter} \\ \textbf{optimization}\end{tabular}} &  \begin{tabular}[l]{@{}l@{}}\cite{kurian2021boat} \\ ~ \end{tabular} & \begin{tabular}[l]{@{}l@{}}Customizable Python API with  \\centralized user interface   \end{tabular}& \begin{tabular}[l]{@{}l@{}}Supports only univariate time-\\series data \end{tabular}\\
& \begin{tabular}[l]{@{}l@{}}\cite{akiba2019optuna} \\ ~ \end{tabular}& \begin{tabular}[l]{@{}l@{}}Dynamic construction of search \\space for parameters \end{tabular} & \begin{tabular}[l]{@{}l@{}}No interface to support deploying  \\external optimizers \end{tabular} \\
& \begin{tabular}[l]{@{}l@{}}\cite{meister2020maggy} \\ ~ \end{tabular}& \begin{tabular}[l]{@{}l@{}}Use of parallel compute resources\\  and global optimization \end{tabular}& \begin{tabular}[l]{@{}l@{}}Was only evaluated on a single \\   use-case \end{tabular}\\
\hline
\multirow{5}{*}{\begin{tabular}[c]{@{}l@{}}\textbf{Concept} \\ \textbf{drift}\end{tabular}} & \cite{vieira2021driftage} & Flexibility and interpretability & Limited to Python\\
& \begin{tabular}[l]{@{}l@{}}\cite{yang2021cade} \\ ~\end{tabular}& \begin{tabular}[l]{@{}l@{}}Explainable and low computational \\ complexity \end{tabular}& \begin{tabular}[l]{@{}l@{}}Mainly focused on detecting new \\ concepts \end{tabular}\\
& \begin{tabular}[l]{@{}l@{}}\cite{mallick2022matchmaker} \\ ~\end{tabular}& \begin{tabular}[l]{@{}l@{}}Scalable, adaptable, and flexible for \\large-scale systems\end{tabular}& \begin{tabular}[l]{@{}l@{}}Cannot handle unseen drifts  \\leading to accuracy drops\end{tabular}\\
\hline
\multirow{3}{*}{\textbf{Generalization}} & \cite{ho2020extensions} & Handles high-dimensional data & Limited to specific tasks\\
& \cite{zhang2020generalizing} & Efficient in unseen domains & Handles supervised tasks only\\
& \cite{hansen2021generalization} & Reduces observational overfitting & Limited to specific tasks\\
\hline
\multirow{3}{*}{\begin{tabular}[c]{@{}l@{}}\textbf{Label} \\ \textbf{noise}\end{tabular}} & \cite{gudovskiy2021autodo} & Robust to train dataset distortions& High number of hyperparameters\\
& \cite{mathew2021defraudnet} & Support real-time systems & Cost and infrastructural limitations\\
& \cite{renggli2020ease} & User-friendly interface & Limited to classification\\
\bottomrule
\end{tabular}}
\end{table}

\subsubsection{Robust Approaches to Concept Drift} A novel multi-agent system framework, Driftage, was proposed to handle concept drift in deployment \cite{vieira2021driftage}. The framework was built based on Monitor-Analyse-Plan-Execute over a shared Knowledge (MAPE-K) architecture for self-adaptation \cite{iglesia2015mape}. For concept drift handling, Driftage follows two flows, Monitor–Analyser and Planner–Executor, for drift detection and analysis. Furthermore, the Analyser and Planner components can be executed using various drift handling techniques. For security applications, the Contrastive Autoencoder for Drifting Detection and Explanation (CADE) method was proposed to detect and understand concept drift occurrence \cite{yang2021cade}. For drift detection, the method defines the distance function from the training dataset based on contrastive learning. For drift explanation, the method uses distance changes to find the features that drive the significant deviation between the drifting data and its nearest class. For industrial processes, an Online Autoregression with Deep Long-Short-Term Memory (OAR-DLSTM) method was developed to handle hybrid recurring concept drifts. The method adopts an Online Autoregression (OAR) algorithm to alleviate the challenge of transfer and overlap between concepts. For making predictions, the method defines a sub-model based on the types of recurring concepts. For large-scale systems, Matchmaker was introduced to deal with covariate shift and concept drift problems \cite{mallick2022matchmaker}. The method ranks batches taken from the training dataset based on spatial proximity to the test sample and the accuracy of the trained ML model. After that, the rankings are combined and the model with the best performance is selected for deployment.

\subsubsection{Generalization Approaches}
External data sources have been used by Ho \textit{et al.} to improve the generalizability of ML models \cite{ho2020extensions}. The authors have shown that external validation outperforms internal validation in improving the generalizability of ML models. In addition, considerations for acquiring a proper external validation dataset and deploying the procedures were presented in the article. Finally, the external validation procedure was extended using convergent and divergent validations. For medical imaging, the BigAug method was proposed to generalize 3D medical image segmentation models to unseen domains \cite{zhang2020generalizing}. The method used data augmentation techniques in the source domain to train neural networks on a large augmented dataset that accommodates several domain-shift situations. However, its effectiveness can be constrained to tasks within the medical imaging domain. Similarly, in reinforcement learning, the SOft Data Augmentation (SODA) method was proposed to improve the generalizability of models using data augmentation and domain randomization \cite{hansen2021generalization}. The method incorporates a novel DMControl generalization benchmark that can achieve high performance in test-time generalization to unseen environments.

\subsubsection{Robust Approaches to Label Noise} An automated dataset optimization (AutoDO) model has been proposed to handle datasets with label noise \cite{gudovskiy2021autodo}. AutoDO employs a soft-label sub-model to estimate hyperparameters for each label in the training dataset based on a noise-free validation dataset. These soft labels are then generated using a softmax function. While this method effectively deals with label noise, it may require tuning a large number of hyperparameters, which can be labor-intensive and time-consuming. Furthermore, the effectiveness of AutoDO may be influenced by the quality and representativeness of the noise-free validation dataset. On the other hand, DeFraudNet, as outlined in \cite{mathew2021defraudnet}, utilizes auto-encoder reconstruction error to reduce label noise. By taking both an unlabeled training dataset and a golden label generation dataset as input, DeFraudNet outputs a denoised, weak-labeled training dataset. This approach integrates label generation with model deployment, offering a comprehensive solution to address label noise. However, it may face challenges related to cost and infrastructure limitations, particularly in scenarios where extensive computational resources are required for training and deploying the model. In a more recent study, the Snoopy system was presented to facilitate the execution of automatic feasibility studies before the deployment of ML solutions \cite{renggli2020ease}. The study has focused on label noise situations, which are sources for the non-zero irreducible error, but its applicability is limited to classification tasks. The core component of Snoopy is the Bayes error rate (BER), which estimates the error of a given task.

\color{black}
\section{Review of MLOps Tools and Technologies}
\label{sec:tool_surv}
Currently, several tools and platforms are available to operationalize ML solutions in production. We will explore the most popular MLOps tools on the basis of their official documentation. Specifically, we review the support of each tool for the various robustness aspects that were surveyed in the previous sections. This can help practitioners in selecting the most useful tool(s) for their applications and in deploying a robust solution. From a technological perspective, the most popular public tools that manage the ML life cycle in deployments are \cite{testi2022mlops}: Amazon SageMaker by Amazon Web Service (AWS) \cite{liberty2020elastic}, Microsoft Azure  \cite{soh2020machine}, Vertex AI (formerly known as AI Platform) \cite{bisong2019overview}, MLFlow  \cite{zaharia2018accelerating} and TensorFlow Extended (TFX) \cite{baylor2017tfx}. The platforms can be categorized into cloud-based platforms: AWS, Azure, and Google Cloud, and non-cloud-based platforms: MLFlow and TFX. Cloud-based platforms are known to be more elastic and easier to use with their interactive interface. However, they are not as customizable as the non-cloud platforms. We provide a brief description of each of these MLOps platforms:

\begin{enumerate}
    \item \textbf{Amazon SageMaker:} Amazon's interface to productize ML solutions in AWS. The platform offers a free trial to start using its services. The platform provides a wide range of functionalities to manage ML solutions in the cloud through interactive interfaces. Moreover, SageMaker features many built-in algorithms and utilities, in addition to the ability to customize them by users. SageMaker also includes the ability to estimate training costs before running the jobs. To store ML artifacts, SageMaker uses Amazon Simple Storage Service (Amazon S3) buckets.
    
    \item \textbf{Microsoft Azure:} Microsoft's cloud-based platform that supports MLOps pipelines through a set of services. The platform includes an Azure Machine Learning Studio environment to operationalize ML projects. Azure also includes a free trial option to get acquainted with its services. One of the main services provided by Microsoft Azure is Azure DevOps, which manages and automates ML project pipelines. Additionally, Azure SDKs include a diverse range of libraries that support many programming languages to build ML projects. In addition to the built-in packages, Azure also supports ML solutions customization.
    
    \item \textbf{Vertex AI:} Google's cloud-based framework to orchestrate the MLOps pipeline in Google Cloud. The AI platform provides several services for building, evaluating and deploying ML solutions. To start with the services, free credit is offered that can be used to test the platform. Many functionalities and components are integrated into the platform. Dataflow is the component that is concerned with managing the data life cycle and supports batch and stream data processing. Another key component is Cloud Build, which enables smooth automation of ML application operations with a serverless CI/CD platform.
    
    \item \textbf{MLFlow:} An open source platform built based on \textit{open interface} philosophy. MLFlow allows the management of the entire ML life cycle through four primary components: MLflow Tracking, MLflow Projects, MLflow Models, and MLflow Models. MLFlow can deploy ML projects and libraries in any programming language. It can also be integrated as a web service with cloud platforms to deploy ML pipelines. MLFlow offers a useful feature in practice that allows users to track performance using interactive plots using a custom-defined metric.
    
    \item \textbf{TFX:} An open source general-purpose platform for production-scale deployments of ML pipelines. TFX was developed on the basis of the TensorFlow \footnote{\url{https://www.tensorflow.org/}} software library for ML and AI. TFX supports the most popular programming languages. To implement an ML pipeline, TFX allows users to configure a sequence of components that can handle data processing and model deployment. For quality monitoring, TFX uses validation checks for every stage of the ML pipeline. These validations can evaluate the data quality and performance of the ML models.
\end{enumerate}

To gain insight into the support for the robustness aspect of MLOpsm tools, we compared the built-in features offered by each tool. The comparison results are summarized in Table \ref{tab:tools_summary}. We can see that all tools support automation management of the ML life cycle. These continuous operations are the cornerstone of any ML project deployed in real-world applications, and they distinguish them from offline ML projects. For DataOps, all platforms provide data management tools, such as data analysis or visualization, to facilitate data quality tracking. On the contrary, other operations are still missing from the tools, such as data augmentation for DataOps. For the rest of the operations, most of the tools support distribution shift detection by employing statistical tests to monitor the data distribution. For ModelOps, all tools incorporate hyperparameter tuning methods to tweak the ML models. Detecting concept drift is only included in AWS SageMake with the CloudWatch metrics repository. CloudWatch monitors the deviation in model performance by monitoring the defined performance metric and activates the model re-training procedure automatically. RobustDG toolkit \cite{mahajan2021connection} was developed by Microsoft to enhance the generalizability of the ML model to unseen domains. The toolkit can be easily extended to write problem-specific domain generalization algorithms. Label noise is handled by the data labeling functionality offered by AWS and Vertext AI. Lastly, an important remark is that we have surveyed the built-in features that are integrated by the tools. However, the majority of platforms exhibit flexibility by allowing developers to write their own solutions and deploy them in the framework. 

\begin{table}
\centering
\caption{Summary of the built-in features of MLOps tools}
% \refstepcounter{table}
\label{tab:tools_summary}
\begin{tabular}{l|l|ccccc} 

\toprule
                                      & \multicolumn{1}{c|}{\multirow{2}{*}{\textbf{Operation}}} & \multicolumn{5}{c}{\textbf{MLOps tool }}      \\ 
\cline{3-7}
                                      & \multicolumn{1}{c|}{}                                    & \textbf{AWS} & \textbf{Azure} & \textbf{Vertex AI} & \textbf{MLFlow} & \textbf{TFX}  \\ 
\hline
\multirow{4}{*}{\textbf{Automation }} & \textbf{Continuous~validation}                                    &      \checkmark     &   \checkmark     &      \checkmark     &    \checkmark    &    \checkmark  \\
                                      & \textbf{Continuous~versioning}                                   &     \checkmark      &    \checkmark    &     \checkmark      &   \checkmark     &    \checkmark  \\
                                      & \textbf{Continuous monitoring}                                    &      \checkmark     &     \checkmark   &    \checkmark        &     \checkmark    &   \checkmark    \\
                                      & \textbf{Continuous update}                                   &       \checkmark    &     \checkmark   &     \checkmark      &    \checkmark    &     \checkmark \\ 
\hline
\multirow{4}{*}{\textbf{DataOps}}     & \textbf{Data cleaning}                                            &      \checkmark     &   \checkmark     &   \checkmark         &  \checkmark       &     \checkmark \\
                                      & \textbf{Anomaly detection}                                        &       \checkmark    &    \checkmark    &     \checkmark      &      -  &   \checkmark \\
                                      & \textbf{Distribution shift}                                       &      \checkmark     &    \checkmark    &    \checkmark       &    -    &  \checkmark    \\
                                      & \textbf{Data augmentation}                                        &       -    &    -   &      -     &  -      &    -  \\ 
\hline
\multirow{4}{*}{\textbf{ModelOps}}    & \textbf{Hyperparameter tuning}                                    &      \checkmark     &    \checkmark    &      \checkmark     &     \checkmark    &    \checkmark  \\
                                      & \textbf{Concept drift}                                            &      \checkmark     &   -    &      -     &   -     &   -   \\
                                      & \textbf{Generalizability}                                         &      -     &   \checkmark    &     -       &   -     &   -   \\
                                      & \textbf{Label noise}                                              &       \checkmark    &   -    &    \checkmark      &   -     &   -   \\
\bottomrule
\end{tabular}
\end{table}

\color{blue}
\section{Insights and Open Challenges}
\label{sec:future}
In this paper, we highlight the lessons learned from the surveyed methods, open challenges and potential future directions to build robust MLOps systems. Although the existing literature provides information on key practices for improving the technical robustness of MLOps approaches, significant gaps still exist, necessitating further research and exploration. Through an extensive literature search, we identified the key insights and several challenges in achieving robustness in MLOps systems and outline opportunities for enhancing the maturity of the field. Our aim is to provide a roadmap to address these challenges and accelerate the widespread adoption of robust machine learning systems in AI applications.

\subsection{Lessons Learned from Robust MLOps Methods}

Several lessons and insights can be inferred from the literature based on the surveyed methods.

\begin{itemize}
    \item \textbf{Monitoring and Performance Optimization:} Implementing advanced monitoring practices in MLOps is crucial for maintaining the reliability and performance of ML systems in production environments. Beyond merely tracking prediction accuracy, real-time monitoring provides deep insights into model behavior and system health, enabling proactive identification of potential issues before they escalate into significant problems.
    
    \item \textbf{Scalable Infrastructure and Resource Management:} The scalability of infrastructure is essential for accommodating varying workloads in MLOps. Leveraging cloud-native solutions such as AWS, Azure, or GCP provides scalable infrastructure and efficient resource management. This approach supports dynamic workloads and large-scale deployments, enhancing operational flexibility and cost optimization.
    
    \item \textbf{Customization:} Customization plays a pivotal role in maximizing the effectiveness of MLOps tools and practices. While off-the-shelf MLOps tools offer a broad range of functionalities, customizing these tools to fit specific project requirements can significantly improve outcomes. Integrating tailored solutions and leveraging developer expertise enables organizations to address unique challenges effectively.
\end{itemize}

\subsection{Challenges in Implementing Robustness}
Implementing robustness in MLOps systems presents multifaceted challenges that require strategic approaches and careful consideration across various dimensions. These challenges span from quantifying overall system robustness to managing critical aspects such as data quality, model interpretability, concept drift, and continuous adaptation.

\begin{itemize}
    \item \textbf{Quantifying System Robustness:} A key challenge in MLOps is quantifying system robustness effectively. This involves developing rigorous methodologies to assess resilience across various dimensions simultaneously, such as handling data perturbations, ensuring model performance in diverse conditions, and maintaining system reliability in dynamic operational environments.
    
    \item \textbf{Integration and Orchestration:} 
    Deploying and managing ML models at scale within production environments presents technical challenges related to integration and versioning integrity. Ensuring seamless integration with existing infrastructure, efficient resource management, and maintaining version control across deployments are critical concerns. Adopting containerization technologies, orchestration frameworks such as Kubernetes\footnote{https://kubernetes.io/}, and advanced DevOps practices can streamline deployment processes and enhance scalability, reliability, and maintainability of MLOps systems.

    \item \textbf{Cost Management:} Optimizing operational costs associated with running ML workloads, such as infrastructure and resource utilization, is critical. Achieving a balance between performance and cost efficiency necessitates advanced resource allocation and management strategies.

    \item \textbf{Human-in-the-Loop:}  Effectively integrating human feedback and intervention into automated MLOps pipelines remains a significant challenge. Incorporating dynamic domain expertise and user feedback to enhance model performance and adaptability is an active area of research \cite{mosqueira2023human}.

    \item \textbf{Heterogeneous Data Sources and Generalization:} Distributed systems, such as federated learning, involve training models across multiple devices or institutions with varying datasets that may differ significantly in distribution, quality, and size \cite{zhang2021survey}. Ensuring that the federated model can generalize well across these heterogeneous data sources without overfitting to any particular client's data remains a substantial challenge.
    
\end{itemize}

\color{black}
\subsection{Future Opportunities}
Based on the open challenges identified, we suggest the following directions to explore in the future.

\subsubsection{Trade-offs in Robustness Aspects of MLOps}
The underlying effects of the various robustness aspects are not investigated for ML systems in production. Due to its high impact on the system design, the trade-off between the robustness aspects of MLOps subtasks must be closely examined. This trade-off determines the balance between the various aspects since addressing a particular aspect can improve/deteriorate the system robustness to other aspects. Many studies have reported side effects between robustness aspects \cite{lu2017dynamic, keskar2017improving}. Improving one aspect of robustness may come at the expense of another, highlighting the need for careful consideration and strategic trade-offs. For example, optimizing model accuracy through complex ensemble techniques might sacrifice interpretability, making it challenging to understand the model's decisions. Therefore, it would be valuable to explore how this interplay can be managed in real-world systems. This would help mitigate the potentially reflecting effects of aspects and emphasize specific issues that appear more frequently in certain problems.

\subsubsection{Quantification of Overall Robustness}
The current evaluation methods primarily focus on quantitative performance metrics such as accuracy and precision. However, robustness encompasses broader aspects such as resilience to adversarial attacks, model interpretability, and generalization across diverse datasets. For example, robustness evaluations may involve stress testing models with adversarial examples, analyzing model behavior under different environmental conditions, and assessing the impact of data distribution shifts on model performance. In general, the issue arises in relation to the interpretation of the results. For example, in the literature, concept drift is evaluated mainly using performance degradation \cite{Bayram2022Degradation}. Nevertheless, other factors could cause performance to degrade, and it does not necessarily indicate a concept drift occurrence. Therefore, developing comprehensive and aspect-specific methodologies for evaluating robustness can provide deeper insights into system performance and identify potential vulnerabilities. Additionally, the use of explainable AI techniques can improve our understanding of system behavior and observations, facilitating the construction of quantifiable solutions \cite{belle2021principles}.

\subsubsection{Holistic Robust MLOps Approach}
Building a holistic, robust ML system for production requires comprehensive frameworks that address all aspects of robustness simultaneously. While existing approaches in the literature partially handle robustness, there is a lack of clarity on how these approaches collectively address the multitude of factors influencing robustness. To bridge this gap, it is imperative to adopt a unified approach that considers the interplay between various factors that influence robustness. The plethora of MLOps tools and platforms available today presents a unique opportunity to construct a holistic, robust MLOps approach. Using the customizable features and integrative capabilities of these tools, a holistic robust MLOps system can be achieved ultimately in real-world production environments. 

\subsubsection{Flexible MLOps Architecture}
The dynamicity of real-world applications requires adaptation in the architectural design of solution systems \cite{oreizy1999architecture}. Traditional monolithic architectures are often rigid and difficult to modify, hindering agility and innovation in MLOps systems. An interesting line of research is to develop modular and extensible architectures that enable flexible integration of new data sources, algorithms, and computational resources. One promising approach is the adoption of microservices-based architectures, which facilitates the development of scalable and decoupled MLOps systems, allowing practitioners to iterate and experiment with different components independently. Additionally, the implementation of containerization and orchestration technologies can enhance scalability and resource utilization, enabling the deployment and management of ML workloads more efficiently. Therefore, future research should focus on designing extensible architectures that promote the development of elastic solutions. These solutions should possess the ability to adapt and evolve in response to changing business requirements and technological advancements.

\color{black}

\section{Conclusion}
\label{sec:conc}
In this survey, we start with the EU's guidelines for the requirements of AI systems to be trustworthy. One of the key principles suggested in the guidelines is that the system should be technically robust. However, the robustness property may have different definitions depending on the domain and the application. However the common point of all definitions is that a robust system can withstand unforeseen conditions during its evolution. AI solutions and its most widespread sub-field of ML are employed in a vast spectrum of applications. In real-world scenarios, the deployment of ML solutions in production requires a set of operations to be performed. MLOps approach has emerged to standardize ML workflows in deployment settings. Therefore, the robustness property should be addressed in MLOps towards building trustworthy ML in production. Through a thorough literature search, we draw out the current challenges and devise the specifications and practices for robust-driven ML solutions in production. By embracing these practices to deliver MLOps pipelines, the primary attention is devoted to building robust ML systems in the long run rather than focusing on short-term gains. In the literature, a key observation is that the robustness property has been used loosely to refer to an approach that partially addresses robustness aspects. Therefore, to build an overall robust ML framework in production, robustness aspects should be fully covered in the constituent ingredients of MLOps, specifically, in the three interdependent components of MLOps: automation management, which is the pillar of MLOps, DataOps, and ModelOps. For each component, we provide related definitions and concepts. Additionally, we organize existing academic research that handles the robustness aspects of production ML according to MLOps components. We also review the most prevalent MLOps tools and their default support of robustness aspects to enable researchers and practitioners to select the suitable tool for their problem.

\begin{acks}

This work has been funded by the Knowledge Foundation of Sweden (KKS) through the Synergy Project AIDA - A Holistic AI-driven Networking and Processing Framework for Industrial IoT (Rek:20200067).
\end{acks}

\bibliographystyle{ACM-Reference-Format}
\bibliography{sample-base}

% %%
% %% If your work has an appendix, this is the place to put it.
% \appendix









\end{document}

%% file: acronyms.tex
\DeclareAcronym{rnn}{
	short = RNN,
	long = {Recurrent Neural Network}
}

\DeclareAcronym{dl}{
	short = DL,
	long = {deep learning}
}

\DeclareAcronym{ml}{
	short = ML,
	long = {machine learning}
}

\DeclareAcronym{ai}{
	short = AI,
	long = {artificial intelligence}
}

\DeclareAcronym{pdf}{
	short = pdf,
	long = {probability density function}
}

\DeclareAcronym{erm}{
	short = ERM,
	long = {empirical risk minimization}
}

\DeclareAcronym{kmm}{
	short = KMM,
	long = {Kernel Mean Matching}
}

\DeclareAcronym{kliep}{
	short = KLIEP,
	long = {Kullback-Leibler Importance Estimation Procedure}
}

\DeclareAcronym{lsif}{
	short = LSIF,
	long = {least squares importance fitting}
}

\DeclareAcronym{atc}{
	short = ATC,
	long = {Average Thresholded Confidence}
}

\DeclareAcronym{doc}{
	short = DoC,
	long = {difference of confidences}
}

\DeclareAcronym{autoeval}{
	short = AutoEval,
	long = {Automatic model Evaluation}
}

\DeclareAcronym{nlp}{
	short = NLP,
	long = {natural language processing}
}

\DeclareAcronym{squad}{
	short = SQuAD,
	long = {Stanford Question Answering Dataset}
}

\DeclareAcronym{svm}{
	short = SVM,
	long = { Support Vector Machines}
}

\DeclareAcronym{gnb}{
	short = GNB,
	long = {Gaussian Naïve Bayes}
}

\DeclareAcronym{rf}{
	short = RF,
	long = {Random Forests}
}

\DeclareAcronym{lr}{
	short = LR,
	long = {Logistic Regression}
}

\DeclareAcronym{knn}{
	short = KNN,
	long = {K-Nearest-Neighbours}
}

\DeclareAcronym{mcc}{
	short = MCC,
	long = {Matthews coefficient of correlation}
}

\DeclareAcronym{arima}{
	short = ARIMA,
	long = {AutoRegressive Integrated Moving Average}
}

\DeclareAcronym{ems}{
	short = EMS,
	long = {energy management system}
}

\DeclareAcronym{lstm}{
	short = LSTM,
	long = {Long Short-Term Memory}
}

\DeclareAcronym{jsd}{
	short = JSD,
	long = {Jensen-Shannon divergence}
}

\DeclareAcronym{kld}{
	short = KLD,
	long = {Kullback–Leibler Divergence}
}

\DeclareAcronym{kde}{
	short = KDE,
	long = {Kernel Density Estimation}
}

\DeclareAcronym{amise}{
	short = AMISE,
	long = {asymptotic mean integrated square error}
}

\DeclareAcronym{hpo}{
	short = HPO,
	long = {Hyperparameter optimization}
}

\DeclareAcronym{bo}{
	short = BO,
	long = {Bayesian optimization}
}

\DeclareAcronym{adwin}{
	short = ADWIN,
	long = {ADaptive WINdowing}
}

\DeclareAcronym{dnn}{
	short = DNN,
	long = {Deep Neural Network}
}

\DeclareAcronym{mlops}{
	short = MLOps,
	long = {Machine Learning Operations}
}

\DeclareAcronym{ci}{
	short = CI,
	long = {continuous integration}
}

\DeclareAcronym{cd}{
	short = CD,
	long = {continuous delivery}
}

\DeclareAcronym{aec}{
	short = AEC,
	long = {architecture, engineering and construction}
}

\DeclareAcronym{aihleg}{
	short = AI HLEG,
	long = {High-Level Expert Group on Artificial Intelligence }
}

\DeclareAcronym{gdpr}{
	short = GDPR,
	long = {General Data Protection Regulation}
}

\DeclareAcronym{iso}{
	short = ISO,
	long = {International Organization for Standardization}
}

\DeclareAcronym{iec}{
	short = IEC,
	long = {International Electrotechnical Commission}
}

\DeclareAcronym{qa}{
	short = QA,
	long = {quality assurance}
}

\DeclareAcronym{sqa}{
	short = SQA,
	long = {Software Quality Assurance}
}

\DeclareAcronym{dq}{
    short = DQ,
    long = {Data Quality}
}

\DeclareAcronym{cart}{
    short = CART,
    long = {classification and regression tree}
}

%% file: main.bbl

\begin{thebibliography}{202}

%%% the Web address.

\ifx \showCODEN    \undefined \def \showCODEN     #1{\unskip}     \fi
\ifx \showDOI      \undefined \def \showDOI       #1{#1}\fi
\ifx \showISBNx    \undefined \def \showISBNx     #1{\unskip}     \fi
\ifx \showISBNxiii \undefined \def \showISBNxiii  #1{\unskip}     \fi
\ifx \showISSN     \undefined \def \showISSN      #1{\unskip}     \fi
\ifx \showLCCN     \undefined \def \showLCCN      #1{\unskip}     \fi
\ifx \shownote     \undefined \def \shownote      #1{#1}          \fi
\ifx \showarticletitle \undefined \def \showarticletitle #1{#1}   \fi
\ifx \showURL      \undefined \def \showURL       {\relax}        \fi
% invisible to TeX
\providecommand\bibfield[2]{#2}
\providecommand\bibinfo[2]{#2}
\providecommand\natexlab[1]{#1}
\providecommand\showeprint[2][]{arXiv:#2}

\bibitem[Agrahari and Singh(2021)]%
        {agrahari2021concept}
\bibfield{author}{\bibinfo{person}{Supriya Agrahari} {and} \bibinfo{person}{Anil~Kumar Singh}.} \bibinfo{year}{2021}\natexlab{}.
\newblock \showarticletitle{Concept drift detection in data stream mining: A literature review}.
\newblock \bibinfo{journal}{\emph{Journal of King Saud University-Computer and Information Sciences}} (\bibinfo{year}{2021}).
\newblock


\bibitem[Akiba et~al\mbox{.}(2019)]%
        {akiba2019optuna}
\bibfield{author}{\bibinfo{person}{Takuya Akiba}, \bibinfo{person}{Shotaro Sano}, \bibinfo{person}{Toshihiko Yanase}, \bibinfo{person}{Takeru Ohta}, {and} \bibinfo{person}{Masanori Koyama}.} \bibinfo{year}{2019}\natexlab{}.
\newblock \showarticletitle{Optuna: A next-generation hyperparameter optimization framework}. In \bibinfo{booktitle}{\emph{Proceedings of the 25th ACM SIGKDD international conference on knowledge discovery \& data mining}}. \bibinfo{pages}{2623--2631}.
\newblock


\bibitem[Al~Ridhawi et~al\mbox{.}(2020)]%
        {al2020generalizing}
\bibfield{author}{\bibinfo{person}{Ismaeel Al~Ridhawi}, \bibinfo{person}{Safa Otoum}, \bibinfo{person}{Moayad Aloqaily}, {and} \bibinfo{person}{Azzedine Boukerche}.} \bibinfo{year}{2020}\natexlab{}.
\newblock \showarticletitle{Generalizing AI: challenges and opportunities for plug and play AI solutions}.
\newblock \bibinfo{journal}{\emph{IEEE Network}} \bibinfo{volume}{35}, \bibinfo{number}{1} (\bibinfo{year}{2020}), \bibinfo{pages}{372--379}.
\newblock


\bibitem[Ali et~al\mbox{.}(2023)]%
        {ali2023leveraging}
\bibfield{author}{\bibinfo{person}{Hazrat Ali}, \bibinfo{person}{Christer Gr{\"o}nlund}, {and} \bibinfo{person}{Zubair Shah}.} \bibinfo{year}{2023}\natexlab{}.
\newblock \showarticletitle{Leveraging GANs for data scarcity of COVID-19: Beyond the hype}. In \bibinfo{booktitle}{\emph{Proceedings of the IEEE/CVF Conference on Computer Vision and Pattern Recognition}}. \bibinfo{pages}{659--667}.
\newblock


\bibitem[Alla and Adari(2021a)]%
        {alla2021beginning}
\bibfield{author}{\bibinfo{person}{Sridhar Alla} {and} \bibinfo{person}{Suman~Kalyan Adari}.} \bibinfo{year}{2021}\natexlab{a}.
\newblock \bibinfo{booktitle}{\emph{Beginning MLOps with MLFlow}}.
\newblock \bibinfo{publisher}{Springer}.
\newblock


\bibitem[Alla and Adari(2021b)]%
        {alla2021mlops}
\bibfield{author}{\bibinfo{person}{Sridhar Alla} {and} \bibinfo{person}{Suman~Kalyan Adari}.} \bibinfo{year}{2021}\natexlab{b}.
\newblock \showarticletitle{What is mlops?}
\newblock In \bibinfo{booktitle}{\emph{Beginning MLOps with MLFlow}}. \bibinfo{publisher}{Springer}, \bibinfo{pages}{79--124}.
\newblock


\bibitem[Allamanis(2019)]%
        {allamanis2019adverse}
\bibfield{author}{\bibinfo{person}{Miltiadis Allamanis}.} \bibinfo{year}{2019}\natexlab{}.
\newblock \showarticletitle{The adverse effects of code duplication in machine learning models of code}. In \bibinfo{booktitle}{\emph{Proceedings of the 2019 ACM SIGPLAN International Symposium on New Ideas, New Paradigms, and Reflections on Programming and Software}}. \bibinfo{pages}{143--153}.
\newblock


\bibitem[Anguita et~al\mbox{.}(2012)]%
        {anguita2012k}
\bibfield{author}{\bibinfo{person}{Davide Anguita}, \bibinfo{person}{Luca Ghelardoni}, \bibinfo{person}{Alessandro Ghio}, \bibinfo{person}{Luca Oneto}, {and} \bibinfo{person}{Sandro Ridella}.} \bibinfo{year}{2012}\natexlab{}.
\newblock \showarticletitle{The ‘K’in K-fold cross validation}. In \bibinfo{booktitle}{\emph{20th European Symposium on Artificial Neural Networks, Computational Intelligence and Machine Learning (ESANN)}}. i6doc. com publ, \bibinfo{pages}{441--446}.
\newblock


\bibitem[Antonini et~al\mbox{.}(2022)]%
        {antonini2022tiny}
\bibfield{author}{\bibinfo{person}{Mattia Antonini}, \bibinfo{person}{Miguel Pincheira}, \bibinfo{person}{Massimo Vecchio}, {and} \bibinfo{person}{Fabio Antonelli}.} \bibinfo{year}{2022}\natexlab{}.
\newblock \showarticletitle{Tiny-MLOps: a framework for orchestrating ML applications at the far edge of IoT systems}. In \bibinfo{booktitle}{\emph{2022 IEEE International Conference on Evolving and Adaptive Intelligent Systems (EAIS)}}. IEEE, \bibinfo{pages}{1--8}.
\newblock


\bibitem[Atwal(2020)]%
        {atwal2020dataops}
\bibfield{author}{\bibinfo{person}{Harvinder Atwal}.} \bibinfo{year}{2020}\natexlab{}.
\newblock \showarticletitle{Dataops technology}.
\newblock In \bibinfo{booktitle}{\emph{Practical DataOps}}. \bibinfo{publisher}{Springer}, \bibinfo{pages}{215--247}.
\newblock


\bibitem[Babbar and Sch{\"o}lkopf(2019)]%
        {babbar2019data}
\bibfield{author}{\bibinfo{person}{Rohit Babbar} {and} \bibinfo{person}{Bernhard Sch{\"o}lkopf}.} \bibinfo{year}{2019}\natexlab{}.
\newblock \showarticletitle{Data scarcity, robustness and extreme multi-label classification}.
\newblock \bibinfo{journal}{\emph{Machine Learning}} \bibinfo{volume}{108}, \bibinfo{number}{8} (\bibinfo{year}{2019}), \bibinfo{pages}{1329--1351}.
\newblock


\bibitem[Bansal et~al\mbox{.}(2022)]%
        {bansal2022systematic}
\bibfield{author}{\bibinfo{person}{Ms~Aayushi Bansal}, \bibinfo{person}{Dr~Rewa Sharma}, {and} \bibinfo{person}{Dr~Mamta Kathuria}.} \bibinfo{year}{2022}\natexlab{}.
\newblock \showarticletitle{A systematic review on data scarcity problem in deep learning: solution and applications}.
\newblock \bibinfo{journal}{\emph{ACM Computing Surveys (CSUR)}} \bibinfo{volume}{54}, \bibinfo{number}{10s} (\bibinfo{year}{2022}), \bibinfo{pages}{1--29}.
\newblock


\bibitem[Baylor et~al\mbox{.}(2017)]%
        {baylor2017tfx}
\bibfield{author}{\bibinfo{person}{Denis Baylor}, \bibinfo{person}{Eric Breck}, \bibinfo{person}{Heng-Tze Cheng}, \bibinfo{person}{Noah Fiedel}, \bibinfo{person}{Chuan~Yu Foo}, \bibinfo{person}{Zakaria Haque}, \bibinfo{person}{Salem Haykal}, \bibinfo{person}{Mustafa Ispir}, \bibinfo{person}{Vihan Jain}, \bibinfo{person}{Levent Koc}, {et~al\mbox{.}}} \bibinfo{year}{2017}\natexlab{}.
\newblock \showarticletitle{TFX: A tensorflow-based production-scale machine learning platform}. In \bibinfo{booktitle}{\emph{Proceedings of the 23rd ACM SIGKDD International Conference on Knowledge Discovery and Data Mining}}. \bibinfo{pages}{1387--1395}.
\newblock


\bibitem[Bayram et~al\mbox{.}(2022b)]%
        {bayram2022drift}
\bibfield{author}{\bibinfo{person}{Firas Bayram}, \bibinfo{person}{Bestoun~S Ahmed}, \bibinfo{person}{Erik Hallin}, {and} \bibinfo{person}{Anton Engman}.} \bibinfo{year}{2022}\natexlab{b}.
\newblock \showarticletitle{A Drift Handling Approach for Self-Adaptive ML Software in Scalable Industrial Processes}. In \bibinfo{booktitle}{\emph{2022 37th IEEE/ACM International Conference on Automated Software Engineering (ASE)}}. IEEE.
\newblock


\bibitem[Bayram et~al\mbox{.}(2022a)]%
        {Bayram2022Degradation}
\bibfield{author}{\bibinfo{person}{Firas Bayram}, \bibinfo{person}{Bestoun~S. Ahmed}, {and} \bibinfo{person}{Andreas Kassler}.} \bibinfo{year}{2022}\natexlab{a}.
\newblock \showarticletitle{From concept drift to model degradation: An overview on performance-aware drift detectors}.
\newblock \bibinfo{journal}{\emph{Knowledge-Based Systems}}  \bibinfo{volume}{245} (\bibinfo{year}{2022}), \bibinfo{pages}{108632}.
\newblock
\showISSN{0950-7051}


\bibitem[Belle and Papantonis(2021)]%
        {belle2021principles}
\bibfield{author}{\bibinfo{person}{Vaishak Belle} {and} \bibinfo{person}{Ioannis Papantonis}.} \bibinfo{year}{2021}\natexlab{}.
\newblock \showarticletitle{Principles and practice of explainable machine learning}.
\newblock \bibinfo{journal}{\emph{Frontiers in big Data}} (\bibinfo{year}{2021}), \bibinfo{pages}{39}.
\newblock


\bibitem[Bergstra and Bengio(2012)]%
        {bergstra2012random}
\bibfield{author}{\bibinfo{person}{James Bergstra} {and} \bibinfo{person}{Yoshua Bengio}.} \bibinfo{year}{2012}\natexlab{}.
\newblock \showarticletitle{Random search for hyper-parameter optimization.}
\newblock \bibinfo{journal}{\emph{Journal of machine learning research}} \bibinfo{volume}{13}, \bibinfo{number}{2} (\bibinfo{year}{2012}).
\newblock


\bibitem[Berrar(2019)]%
        {berrar2019cross}
\bibfield{author}{\bibinfo{person}{Daniel Berrar}.} \bibinfo{year}{2019}\natexlab{}.
\newblock \bibinfo{title}{Cross-Validation.}
\newblock
\newblock


\bibitem[Bhagoji et~al\mbox{.}(2018)]%
        {bhagoji2018enhancing}
\bibfield{author}{\bibinfo{person}{Arjun~Nitin Bhagoji}, \bibinfo{person}{Daniel Cullina}, \bibinfo{person}{Chawin Sitawarin}, {and} \bibinfo{person}{Prateek Mittal}.} \bibinfo{year}{2018}\natexlab{}.
\newblock \showarticletitle{Enhancing robustness of machine learning systems via data transformations}. In \bibinfo{booktitle}{\emph{2018 52nd Annual Conference on Information Sciences and Systems (CISS)}}. IEEE, \bibinfo{pages}{1--5}.
\newblock


\bibitem[Bisong(2019)]%
        {bisong2019overview}
\bibfield{author}{\bibinfo{person}{Ekaba Bisong}.} \bibinfo{year}{2019}\natexlab{}.
\newblock \showarticletitle{An overview of google cloud platform services}.
\newblock \bibinfo{journal}{\emph{Building Machine Learning and Deep Learning Models on Google Cloud Platform}} (\bibinfo{year}{2019}), \bibinfo{pages}{7--10}.
\newblock


\bibitem[Botchkarev(2018)]%
        {botchkarev2018performance}
\bibfield{author}{\bibinfo{person}{Alexei Botchkarev}.} \bibinfo{year}{2018}\natexlab{}.
\newblock \showarticletitle{Performance metrics (error measures) in machine learning regression, forecasting and prognostics: Properties and typology}.
\newblock \bibinfo{journal}{\emph{arXiv preprint arXiv:1809.03006}} (\bibinfo{year}{2018}).
\newblock


\bibitem[Bradley(1978)]%
        {bradley1978robustness}
\bibfield{author}{\bibinfo{person}{James~V Bradley}.} \bibinfo{year}{1978}\natexlab{}.
\newblock \showarticletitle{Robustness?}
\newblock \bibinfo{journal}{\emph{Brit. J. Math. Statist. Psych.}} \bibinfo{volume}{31}, \bibinfo{number}{2} (\bibinfo{year}{1978}), \bibinfo{pages}{144--152}.
\newblock


\bibitem[Breck et~al\mbox{.}(2017)]%
        {breck2017ml}
\bibfield{author}{\bibinfo{person}{Eric Breck}, \bibinfo{person}{Shanqing Cai}, \bibinfo{person}{Eric Nielsen}, \bibinfo{person}{Michael Salib}, {and} \bibinfo{person}{D Sculley}.} \bibinfo{year}{2017}\natexlab{}.
\newblock \showarticletitle{The ML test score: A rubric for ML production readiness and technical debt reduction}. In \bibinfo{booktitle}{\emph{2017 IEEE International Conference on Big Data (Big Data)}}. IEEE, \bibinfo{pages}{1123--1132}.
\newblock


\bibitem[Breck et~al\mbox{.}(2019)]%
        {breck2019data}
\bibfield{author}{\bibinfo{person}{Eric Breck}, \bibinfo{person}{Neoklis Polyzotis}, \bibinfo{person}{Sudip Roy}, \bibinfo{person}{Steven Whang}, {and} \bibinfo{person}{Martin Zinkevich}.} \bibinfo{year}{2019}\natexlab{}.
\newblock \showarticletitle{Data Validation for Machine Learning}. In \bibinfo{booktitle}{\emph{MLSys}}.
\newblock


\bibitem[Byrd et~al\mbox{.}(1994)]%
        {byrd1994representations}
\bibfield{author}{\bibinfo{person}{Richard~H Byrd}, \bibinfo{person}{Jorge Nocedal}, {and} \bibinfo{person}{Robert~B Schnabel}.} \bibinfo{year}{1994}\natexlab{}.
\newblock \showarticletitle{Representations of quasi-Newton matrices and their use in limited memory methods}.
\newblock \bibinfo{journal}{\emph{Mathematical Programming}} \bibinfo{volume}{63}, \bibinfo{number}{1} (\bibinfo{year}{1994}), \bibinfo{pages}{129--156}.
\newblock


\bibitem[Capizzi et~al\mbox{.}(2019)]%
        {capizzi2019devops}
\bibfield{author}{\bibinfo{person}{Antonio Capizzi}, \bibinfo{person}{Salvatore Distefano}, {and} \bibinfo{person}{Manuel Mazzara}.} \bibinfo{year}{2019}\natexlab{}.
\newblock \showarticletitle{From devops to devdataops: Data management in devops processes}. In \bibinfo{booktitle}{\emph{International Workshop on Software Engineering Aspects of Continuous Development and New Paradigms of Software Production and Deployment}}. Springer, \bibinfo{pages}{52--62}.
\newblock


\bibitem[Carlini and Wagner(2017)]%
        {carlini2017towards}
\bibfield{author}{\bibinfo{person}{Nicholas Carlini} {and} \bibinfo{person}{David Wagner}.} \bibinfo{year}{2017}\natexlab{}.
\newblock \showarticletitle{Towards evaluating the robustness of neural networks}. In \bibinfo{booktitle}{\emph{2017 ieee symposium on security and privacy (sp)}}. Ieee, \bibinfo{pages}{39--57}.
\newblock


\bibitem[Chandola et~al\mbox{.}(2009)]%
        {chandola2009anomaly}
\bibfield{author}{\bibinfo{person}{Varun Chandola}, \bibinfo{person}{Arindam Banerjee}, {and} \bibinfo{person}{Vipin Kumar}.} \bibinfo{year}{2009}\natexlab{}.
\newblock \showarticletitle{Anomaly detection: A survey}.
\newblock \bibinfo{journal}{\emph{ACM computing surveys (CSUR)}} \bibinfo{volume}{41}, \bibinfo{number}{3} (\bibinfo{year}{2009}), \bibinfo{pages}{1--58}.
\newblock


\bibitem[Chao et~al\mbox{.}(2019)]%
        {chao2019recent}
\bibfield{author}{\bibinfo{person}{Guoqing Chao}, \bibinfo{person}{Yuan Luo}, {and} \bibinfo{person}{Weiping Ding}.} \bibinfo{year}{2019}\natexlab{}.
\newblock \showarticletitle{Recent advances in supervised dimension reduction: A survey}.
\newblock \bibinfo{journal}{\emph{Machine learning and knowledge extraction}} \bibinfo{volume}{1}, \bibinfo{number}{1} (\bibinfo{year}{2019}), \bibinfo{pages}{341--358}.
\newblock


\bibitem[Chatila et~al\mbox{.}(2021)]%
        {chatila2021trustworthy}
\bibfield{author}{\bibinfo{person}{Raja Chatila}, \bibinfo{person}{Virginia Dignum}, \bibinfo{person}{Michael Fisher}, \bibinfo{person}{Fosca Giannotti}, \bibinfo{person}{Katharina Morik}, \bibinfo{person}{Stuart Russell}, {and} \bibinfo{person}{Karen Yeung}.} \bibinfo{year}{2021}\natexlab{}.
\newblock \showarticletitle{Trustworthy ai}.
\newblock In \bibinfo{booktitle}{\emph{Reflections on Artificial Intelligence for Humanity}}. \bibinfo{publisher}{Springer}, \bibinfo{pages}{13--39}.
\newblock


\bibitem[Chatterjee et~al\mbox{.}(2022)]%
        {ESR2022}
\bibfield{author}{\bibinfo{person}{Ayan Chatterjee}, \bibinfo{person}{Bestoun~S Ahmed}, \bibinfo{person}{Erik Hallin}, {and} \bibinfo{person}{Anton Engman}.} \bibinfo{year}{2022}\natexlab{}.
\newblock \showarticletitle{Testing of machine learning models with limited samples: an industrial vacuum pumping application}. In \bibinfo{booktitle}{\emph{Proceedings of the 30th ACM Joint European Software Engineering Conference and Symposium on the Foundations of Software Engineering}}. \bibinfo{pages}{1280--1290}.
\newblock


\bibitem[Chen et~al\mbox{.}(2021a)]%
        {chen2021did}
\bibfield{author}{\bibinfo{person}{Lingjiao Chen}, \bibinfo{person}{Tracy Cai}, \bibinfo{person}{Matei Zaharia}, {and} \bibinfo{person}{James Zou}.} \bibinfo{year}{2021}\natexlab{a}.
\newblock \showarticletitle{Did the model change? Efficiently assessing machine learning API shifts}.
\newblock \bibinfo{journal}{\emph{arXiv preprint arXiv:2107.14203}} (\bibinfo{year}{2021}).
\newblock


\bibitem[Chen et~al\mbox{.}(2021b)]%
        {Chen2021Mandoline}
\bibfield{author}{\bibinfo{person}{Mayee Chen}, \bibinfo{person}{Karan Goel}, \bibinfo{person}{Nimit~S Sohoni}, \bibinfo{person}{Fait Poms}, \bibinfo{person}{Kayvon Fatahalian}, {and} \bibinfo{person}{Christopher R{\'e}}.} \bibinfo{year}{2021}\natexlab{b}.
\newblock \showarticletitle{Mandoline: Model Evaluation under Distribution Shift}. In \bibinfo{booktitle}{\emph{International Conference on Machine Learning}}. PMLR, \bibinfo{pages}{1617--1629}.
\newblock


\bibitem[Chowdary et~al\mbox{.}(2022)]%
        {chowdary2022accelerating}
\bibfield{author}{\bibinfo{person}{Mandepudi~Nobel Chowdary}, \bibinfo{person}{Bussa Sankeerth}, \bibinfo{person}{Chennupati~Kumar Chowdary}, {and} \bibinfo{person}{Manu Gupta}.} \bibinfo{year}{2022}\natexlab{}.
\newblock \showarticletitle{Accelerating the Machine Learning Model Deployment using MLOps}. In \bibinfo{booktitle}{\emph{Journal of Physics: Conference Series}}, Vol.~\bibinfo{volume}{2327}. IOP Publishing, \bibinfo{pages}{012027}.
\newblock


\bibitem[Chowdhury and Alspector(2003)]%
        {chowdhury2003data}
\bibfield{author}{\bibinfo{person}{Abdur Chowdhury} {and} \bibinfo{person}{Joshua Alspector}.} \bibinfo{year}{2003}\natexlab{}.
\newblock \showarticletitle{Data duplication: an imbalance problem?}. In \bibinfo{booktitle}{\emph{ICML’2003 Workshop on Learning from Imbalanced Data Sets (II), Washington, DC}}. Citeseer.
\newblock


\bibitem[Chu et~al\mbox{.}(2016)]%
        {chu2016data}
\bibfield{author}{\bibinfo{person}{Xu Chu}, \bibinfo{person}{Ihab~F Ilyas}, \bibinfo{person}{Sanjay Krishnan}, {and} \bibinfo{person}{Jiannan Wang}.} \bibinfo{year}{2016}\natexlab{}.
\newblock \showarticletitle{Data cleaning: Overview and emerging challenges}. In \bibinfo{booktitle}{\emph{Proceedings of the 2016 international conference on management of data}}. \bibinfo{pages}{2201--2206}.
\newblock


\bibitem[Cohen(1968)]%
        {cohen1968weighted}
\bibfield{author}{\bibinfo{person}{Jacob Cohen}.} \bibinfo{year}{1968}\natexlab{}.
\newblock \showarticletitle{Weighted kappa: nominal scale agreement provision for scaled disagreement or partial credit.}
\newblock \bibinfo{journal}{\emph{Psychological bulletin}} \bibinfo{volume}{70}, \bibinfo{number}{4} (\bibinfo{year}{1968}), \bibinfo{pages}{213}.
\newblock


\bibitem[Commission et~al\mbox{.}(2019)]%
        {eu2019ethics}
\bibfield{author}{\bibinfo{person}{European Commission}, \bibinfo{person}{Content Directorate-General~for Communications~Networks}, {and} \bibinfo{person}{Technology}.} \bibinfo{year}{2019}\natexlab{}.
\newblock \bibinfo{booktitle}{\emph{Ethics guidelines for trustworthy AI}}.
\newblock \bibinfo{publisher}{Publications Office}.
\newblock


\bibitem[Crankshaw et~al\mbox{.}(2014)]%
        {crankshaw2014missing}
\bibfield{author}{\bibinfo{person}{Daniel Crankshaw}, \bibinfo{person}{Peter Bailis}, \bibinfo{person}{Joseph~E Gonzalez}, \bibinfo{person}{Haoyuan Li}, \bibinfo{person}{Zhao Zhang}, \bibinfo{person}{Michael~J Franklin}, \bibinfo{person}{Ali Ghodsi}, {and} \bibinfo{person}{Michael~I Jordan}.} \bibinfo{year}{2014}\natexlab{}.
\newblock \showarticletitle{The missing piece in complex analytics: Low latency, scalable model management and serving with velox}.
\newblock \bibinfo{journal}{\emph{arXiv preprint arXiv:1409.3809}} (\bibinfo{year}{2014}).
\newblock


\bibitem[Diakonikolas et~al\mbox{.}(2021)]%
        {diakonikolas2021robustness}
\bibfield{author}{\bibinfo{person}{Ilias Diakonikolas}, \bibinfo{person}{Gautam Kamath}, \bibinfo{person}{Daniel~M Kane}, \bibinfo{person}{Jerry Li}, \bibinfo{person}{Ankur Moitra}, {and} \bibinfo{person}{Alistair Stewart}.} \bibinfo{year}{2021}\natexlab{}.
\newblock \showarticletitle{Robustness meets algorithms}.
\newblock \bibinfo{journal}{\emph{Commun. ACM}} \bibinfo{volume}{64}, \bibinfo{number}{5} (\bibinfo{year}{2021}), \bibinfo{pages}{107--115}.
\newblock


\bibitem[Diallo et~al\mbox{.}(2012)]%
        {diallo2012real}
\bibfield{author}{\bibinfo{person}{Ousmane Diallo}, \bibinfo{person}{Joel~JPC Rodrigues}, {and} \bibinfo{person}{Mbaye Sene}.} \bibinfo{year}{2012}\natexlab{}.
\newblock \showarticletitle{Real-time data management on wireless sensor networks: A survey}.
\newblock \bibinfo{journal}{\emph{Journal of Network and Computer Applications}} \bibinfo{volume}{35}, \bibinfo{number}{3} (\bibinfo{year}{2012}), \bibinfo{pages}{1013--1021}.
\newblock


\bibitem[Dogan and Birant(2021)]%
        {dogan2021machine}
\bibfield{author}{\bibinfo{person}{Alican Dogan} {and} \bibinfo{person}{Derya Birant}.} \bibinfo{year}{2021}\natexlab{}.
\newblock \showarticletitle{Machine learning and data mining in manufacturing}.
\newblock \bibinfo{journal}{\emph{Expert Systems with Applications}}  \bibinfo{volume}{166} (\bibinfo{year}{2021}), \bibinfo{pages}{114060}.
\newblock


\bibitem[Domingos(2012)]%
        {domingos2012few}
\bibfield{author}{\bibinfo{person}{Pedro Domingos}.} \bibinfo{year}{2012}\natexlab{}.
\newblock \showarticletitle{A few useful things to know about machine learning}.
\newblock \bibinfo{journal}{\emph{Commun. ACM}} \bibinfo{volume}{55}, \bibinfo{number}{10} (\bibinfo{year}{2012}), \bibinfo{pages}{78--87}.
\newblock


\bibitem[Dries and R{\"{u}}ckert(2009)]%
        {Dries2009AD}
\bibfield{author}{\bibinfo{person}{Anton Dries} {and} \bibinfo{person}{Ulrich R{\"{u}}ckert}.} \bibinfo{year}{2009}\natexlab{}.
\newblock \showarticletitle{Adaptive concept drift detection}.
\newblock \bibinfo{journal}{\emph{Statistical Analysis and Data Mining}} \bibinfo{volume}{2}, \bibinfo{number}{5-6} (\bibinfo{year}{2009}), \bibinfo{pages}{311--327}.
\newblock


\bibitem[Du et~al\mbox{.}(2019)]%
        {du2019techniques}
\bibfield{author}{\bibinfo{person}{Mengnan Du}, \bibinfo{person}{Ninghao Liu}, {and} \bibinfo{person}{Xia Hu}.} \bibinfo{year}{2019}\natexlab{}.
\newblock \showarticletitle{Techniques for interpretable machine learning}.
\newblock \bibinfo{journal}{\emph{Commun. ACM}} \bibinfo{volume}{63}, \bibinfo{number}{1} (\bibinfo{year}{2019}), \bibinfo{pages}{68--77}.
\newblock


\bibitem[Du et~al\mbox{.}(2021)]%
        {du2021understanding}
\bibfield{author}{\bibinfo{person}{Xianping Du}, \bibinfo{person}{Hongyi Xu}, {and} \bibinfo{person}{Feng Zhu}.} \bibinfo{year}{2021}\natexlab{}.
\newblock \showarticletitle{Understanding the effect of hyperparameter optimization on machine learning models for structure design problems}.
\newblock \bibinfo{journal}{\emph{Computer-Aided Design}}  \bibinfo{volume}{135} (\bibinfo{year}{2021}), \bibinfo{pages}{103013}.
\newblock


\bibitem[Dvorzak and Wagner(2016)]%
        {dvorzak2016sparse}
\bibfield{author}{\bibinfo{person}{Michaela Dvorzak} {and} \bibinfo{person}{Helga Wagner}.} \bibinfo{year}{2016}\natexlab{}.
\newblock \showarticletitle{Sparse Bayesian modelling of underreported count data}.
\newblock \bibinfo{journal}{\emph{Statistical Modelling}} \bibinfo{volume}{16}, \bibinfo{number}{1} (\bibinfo{year}{2016}), \bibinfo{pages}{24--46}.
\newblock


\bibitem[Ehrlinger et~al\mbox{.}(2018)]%
        {ehrlinger2018treating}
\bibfield{author}{\bibinfo{person}{Lisa Ehrlinger}, \bibinfo{person}{Thomas Grubinger}, \bibinfo{person}{Bence Varga}, \bibinfo{person}{Mario Pichler}, \bibinfo{person}{Thomas Natschl{\"a}ger}, {and} \bibinfo{person}{J{\"u}rgen Zeindl}.} \bibinfo{year}{2018}\natexlab{}.
\newblock \showarticletitle{Treating missing data in industrial data analytics}. In \bibinfo{booktitle}{\emph{2018 Thirteenth International Conference on Digital Information Management (ICDIM)}}. IEEE, \bibinfo{pages}{148--155}.
\newblock


\bibitem[Elwell and Polikar(2011)]%
        {Elwell2011NSE}
\bibfield{author}{\bibinfo{person}{Ryan Elwell} {and} \bibinfo{person}{Robi Polikar}.} \bibinfo{year}{2011}\natexlab{}.
\newblock \showarticletitle{Incremental learning of concept drift in nonstationary environments}.
\newblock \bibinfo{journal}{\emph{IEEE Transactions on Neural Networks}} \bibinfo{volume}{22}, \bibinfo{number}{10} (\bibinfo{year}{2011}), \bibinfo{pages}{1517--1531}.
\newblock


\bibitem[Emaminejad and Akhavian(2022)]%
        {emaminejad2022trustworthy}
\bibfield{author}{\bibinfo{person}{Newsha Emaminejad} {and} \bibinfo{person}{Reza Akhavian}.} \bibinfo{year}{2022}\natexlab{}.
\newblock \showarticletitle{Trustworthy AI and robotics: Implications for the AEC industry}.
\newblock \bibinfo{journal}{\emph{Automation in Construction}}  \bibinfo{volume}{139} (\bibinfo{year}{2022}), \bibinfo{pages}{104298}.
\newblock


\bibitem[Erickson and Kitamura(2021)]%
        {erickson2021magician}
\bibfield{author}{\bibinfo{person}{Bradley~J Erickson} {and} \bibinfo{person}{Felipe Kitamura}.} \bibinfo{year}{2021}\natexlab{}.
\newblock \showarticletitle{Magician’s corner: 9. Performance metrics for machine learning models}.
\newblock \bibinfo{journal}{\emph{Radiology: Artificial Intelligence}} \bibinfo{volume}{3}, \bibinfo{number}{3} (\bibinfo{year}{2021}).
\newblock


\bibitem[Feurer and Hutter(2019)]%
        {feurer2019hyperparameter}
\bibfield{author}{\bibinfo{person}{Matthias Feurer} {and} \bibinfo{person}{Frank Hutter}.} \bibinfo{year}{2019}\natexlab{}.
\newblock \showarticletitle{Hyperparameter optimization}.
\newblock In \bibinfo{booktitle}{\emph{Automated machine learning}}. \bibinfo{publisher}{Springer, Cham}, \bibinfo{pages}{3--33}.
\newblock


\bibitem[Fr{\'e}nay and Verleysen(2013)]%
        {frenay2013classification}
\bibfield{author}{\bibinfo{person}{Beno{\^\i}t Fr{\'e}nay} {and} \bibinfo{person}{Michel Verleysen}.} \bibinfo{year}{2013}\natexlab{}.
\newblock \showarticletitle{Classification in the presence of label noise: a survey}.
\newblock \bibinfo{journal}{\emph{IEEE transactions on neural networks and learning systems}} \bibinfo{volume}{25}, \bibinfo{number}{5} (\bibinfo{year}{2013}), \bibinfo{pages}{845--869}.
\newblock


\bibitem[Garcia et~al\mbox{.}(2015)]%
        {garcia2015effect}
\bibfield{author}{\bibinfo{person}{Lu{\'\i}s~PF Garcia}, \bibinfo{person}{Andr{\'e}~CPLF de Carvalho}, {and} \bibinfo{person}{Ana~C Lorena}.} \bibinfo{year}{2015}\natexlab{}.
\newblock \showarticletitle{Effect of label noise in the complexity of classification problems}.
\newblock \bibinfo{journal}{\emph{Neurocomputing}}  \bibinfo{volume}{160} (\bibinfo{year}{2015}), \bibinfo{pages}{108--119}.
\newblock


\bibitem[Garg et~al\mbox{.}(2021)]%
        {garg2021continuous}
\bibfield{author}{\bibinfo{person}{Satvik Garg}, \bibinfo{person}{Pradyumn Pundir}, \bibinfo{person}{Geetanjali Rathee}, \bibinfo{person}{PK Gupta}, \bibinfo{person}{Somya Garg}, {and} \bibinfo{person}{Saransh Ahlawat}.} \bibinfo{year}{2021}\natexlab{}.
\newblock \showarticletitle{On continuous integration/continuous delivery for automated deployment of machine learning models using mlops}. In \bibinfo{booktitle}{\emph{2021 IEEE fourth international conference on artificial intelligence and knowledge engineering (AIKE)}}. IEEE, \bibinfo{pages}{25--28}.
\newblock


\bibitem[Ge et~al\mbox{.}(2019)]%
        {ge2019aeromagnetic}
\bibfield{author}{\bibinfo{person}{Jian Ge}, \bibinfo{person}{Han Li}, \bibinfo{person}{Hongpeng Wang}, \bibinfo{person}{Haobin Dong}, \bibinfo{person}{Huan Liu}, \bibinfo{person}{Wenjie Wang}, \bibinfo{person}{Zhiwen Yuan}, \bibinfo{person}{Jun Zhu}, {and} \bibinfo{person}{Haiyang Zhang}.} \bibinfo{year}{2019}\natexlab{}.
\newblock \showarticletitle{Aeromagnetic compensation algorithm robust to outliers of magnetic sensor based on Huber loss method}.
\newblock \bibinfo{journal}{\emph{IEEE Sensors Journal}} \bibinfo{volume}{19}, \bibinfo{number}{14} (\bibinfo{year}{2019}), \bibinfo{pages}{5499--5505}.
\newblock


\bibitem[Greco and Cerquitelli(2021)]%
        {greco2021drift}
\bibfield{author}{\bibinfo{person}{Salvatore Greco} {and} \bibinfo{person}{Tania Cerquitelli}.} \bibinfo{year}{2021}\natexlab{}.
\newblock \showarticletitle{Drift Lens: Real-time unsupervised Concept Drift detection by evaluating per-label embedding distributions}. In \bibinfo{booktitle}{\emph{2021 International Conference on Data Mining Workshops (ICDMW)}}. IEEE, \bibinfo{pages}{341--349}.
\newblock


\bibitem[Gudovskiy et~al\mbox{.}(2021)]%
        {gudovskiy2021autodo}
\bibfield{author}{\bibinfo{person}{Denis Gudovskiy}, \bibinfo{person}{Luca Rigazio}, \bibinfo{person}{Shun Ishizaka}, \bibinfo{person}{Kazuki Kozuka}, {and} \bibinfo{person}{Sotaro Tsukizawa}.} \bibinfo{year}{2021}\natexlab{}.
\newblock \showarticletitle{Autodo: Robust autoaugment for biased data with label noise via scalable probabilistic implicit differentiation}. In \bibinfo{booktitle}{\emph{Proceedings of the IEEE/CVF Conference on Computer Vision and Pattern Recognition}}. \bibinfo{pages}{16601--16610}.
\newblock


\bibitem[Guo et~al\mbox{.}(2008)]%
        {guo2008class}
\bibfield{author}{\bibinfo{person}{Xinjian Guo}, \bibinfo{person}{Yilong Yin}, \bibinfo{person}{Cailing Dong}, \bibinfo{person}{Gongping Yang}, {and} \bibinfo{person}{Guangtong Zhou}.} \bibinfo{year}{2008}\natexlab{}.
\newblock \showarticletitle{On the class imbalance problem}. In \bibinfo{booktitle}{\emph{2008 Fourth international conference on natural computation}}, Vol.~\bibinfo{volume}{4}. IEEE, \bibinfo{pages}{192--201}.
\newblock


\bibitem[Hamon et~al\mbox{.}(2020)]%
        {hamon2020robustness}
\bibfield{author}{\bibinfo{person}{Ronan Hamon}, \bibinfo{person}{Henrik Junklewitz}, {and} \bibinfo{person}{Ignacio Sanchez}.} \bibinfo{year}{2020}\natexlab{}.
\newblock \showarticletitle{Robustness and explainability of artificial intelligence}.
\newblock \bibinfo{journal}{\emph{Publications Office of the European Union}} (\bibinfo{year}{2020}).
\newblock


\bibitem[Hansen and Wang(2021)]%
        {hansen2021generalization}
\bibfield{author}{\bibinfo{person}{Nicklas Hansen} {and} \bibinfo{person}{Xiaolong Wang}.} \bibinfo{year}{2021}\natexlab{}.
\newblock \showarticletitle{Generalization in reinforcement learning by soft data augmentation}. In \bibinfo{booktitle}{\emph{2021 IEEE International Conference on Robotics and Automation (ICRA)}}. IEEE, \bibinfo{pages}{13611--13617}.
\newblock


\bibitem[Hertel et~al\mbox{.}(2020)]%
        {hertel2020sherpa}
\bibfield{author}{\bibinfo{person}{Lars Hertel}, \bibinfo{person}{Julian Collado}, \bibinfo{person}{Peter Sadowski}, \bibinfo{person}{Jordan Ott}, {and} \bibinfo{person}{Pierre Baldi}.} \bibinfo{year}{2020}\natexlab{}.
\newblock \showarticletitle{Sherpa: Robust hyperparameter optimization for machine learning}.
\newblock \bibinfo{journal}{\emph{SoftwareX}}  \bibinfo{volume}{12} (\bibinfo{year}{2020}), \bibinfo{pages}{100591}.
\newblock


\bibitem[Hewage and Meedeniya(2022)]%
        {hewage2022machine}
\bibfield{author}{\bibinfo{person}{Nipuni Hewage} {and} \bibinfo{person}{Dulani Meedeniya}.} \bibinfo{year}{2022}\natexlab{}.
\newblock \showarticletitle{Machine learning operations: A survey on MLOps tool support}.
\newblock \bibinfo{journal}{\emph{arXiv preprint arXiv:2202.10169}} (\bibinfo{year}{2022}).
\newblock


\bibitem[Hinder et~al\mbox{.}(2020)]%
        {hinder2020towards}
\bibfield{author}{\bibinfo{person}{Fabian Hinder}, \bibinfo{person}{Andr{\'e} Artelt}, {and} \bibinfo{person}{Barbara Hammer}.} \bibinfo{year}{2020}\natexlab{}.
\newblock \showarticletitle{Towards non-parametric drift detection via dynamic adapting window independence drift detection (DAWIDD)}. In \bibinfo{booktitle}{\emph{International Conference on Machine Learning}}. PMLR, \bibinfo{pages}{4249--4259}.
\newblock


\bibitem[Ho et~al\mbox{.}(2020)]%
        {ho2020extensions}
\bibfield{author}{\bibinfo{person}{Sung~Yang Ho}, \bibinfo{person}{Kimberly Phua}, \bibinfo{person}{Limsoon Wong}, {and} \bibinfo{person}{Wilson Wen~Bin Goh}.} \bibinfo{year}{2020}\natexlab{}.
\newblock \showarticletitle{Extensions of the external validation for checking learned model interpretability and generalizability}.
\newblock \bibinfo{journal}{\emph{Patterns}} \bibinfo{volume}{1}, \bibinfo{number}{8} (\bibinfo{year}{2020}), \bibinfo{pages}{100129}.
\newblock


\bibitem[Hoens et~al\mbox{.}(2011)]%
        {Hoens2011Imbalance}
\bibfield{author}{\bibinfo{person}{T.~R. Hoens}, \bibinfo{person}{R. Polikar}, {and} \bibinfo{person}{N. Chawla}.} \bibinfo{year}{2011}\natexlab{}.
\newblock \showarticletitle{Learning from streaming data with concept drift and imbalance: an overview}.
\newblock \bibinfo{journal}{\emph{Progress in Artificial Intelligence}}  \bibinfo{volume}{1} (\bibinfo{year}{2011}), \bibinfo{pages}{89--101}.
\newblock


\bibitem[Hohman et~al\mbox{.}(2020)]%
        {hohman2020understanding}
\bibfield{author}{\bibinfo{person}{Fred Hohman}, \bibinfo{person}{Kanit Wongsuphasawat}, \bibinfo{person}{Mary~Beth Kery}, {and} \bibinfo{person}{Kayur Patel}.} \bibinfo{year}{2020}\natexlab{}.
\newblock \showarticletitle{Understanding and visualizing data iteration in machine learning}. In \bibinfo{booktitle}{\emph{Proceedings of the 2020 CHI conference on human factors in computing systems}}. \bibinfo{pages}{1--13}.
\newblock


\bibitem[Hosseini et~al\mbox{.}(2017)]%
        {hosseini2017google}
\bibfield{author}{\bibinfo{person}{Hossein Hosseini}, \bibinfo{person}{Baicen Xiao}, {and} \bibinfo{person}{Radha Poovendran}.} \bibinfo{year}{2017}\natexlab{}.
\newblock \showarticletitle{Google's cloud vision api is not robust to noise}. In \bibinfo{booktitle}{\emph{2017 16th IEEE international conference on machine learning and applications (ICMLA)}}. IEEE, \bibinfo{pages}{101--105}.
\newblock


\bibitem[Huang et~al\mbox{.}(2021)]%
        {huang2021modelci}
\bibfield{author}{\bibinfo{person}{Yizheng Huang}, \bibinfo{person}{Huaizheng Zhang}, \bibinfo{person}{Yonggang Wen}, \bibinfo{person}{Peng Sun}, {and} \bibinfo{person}{Nguyen Binh~Duong Ta}.} \bibinfo{year}{2021}\natexlab{}.
\newblock \showarticletitle{Modelci-e: Enabling continual learning in deep learning serving systems}.
\newblock \bibinfo{journal}{\emph{arXiv preprint arXiv:2106.03122}} (\bibinfo{year}{2021}).
\newblock


\bibitem[Humble and Farley(2010)]%
        {humble2010continuous}
\bibfield{author}{\bibinfo{person}{Jez Humble} {and} \bibinfo{person}{David Farley}.} \bibinfo{year}{2010}\natexlab{}.
\newblock \bibinfo{booktitle}{\emph{Continuous delivery: reliable software releases through build, test, and deployment automation}}.
\newblock \bibinfo{publisher}{Pearson Education}.
\newblock


\bibitem[Hummer et~al\mbox{.}(2019)]%
        {hummer2019modelops}
\bibfield{author}{\bibinfo{person}{Waldemar Hummer}, \bibinfo{person}{Vinod Muthusamy}, \bibinfo{person}{Thomas Rausch}, \bibinfo{person}{Parijat Dube}, \bibinfo{person}{Kaoutar El~Maghraoui}, \bibinfo{person}{Anupama Murthi}, {and} \bibinfo{person}{Punleuk Oum}.} \bibinfo{year}{2019}\natexlab{}.
\newblock \showarticletitle{Modelops: Cloud-based lifecycle management for reliable and trusted ai}. In \bibinfo{booktitle}{\emph{2019 IEEE International Conference on Cloud Engineering (IC2E)}}. IEEE, \bibinfo{pages}{113--120}.
\newblock


\bibitem[Ibidunmoye et~al\mbox{.}(2015)]%
        {ibidunmoye2015performance}
\bibfield{author}{\bibinfo{person}{Olumuyiwa Ibidunmoye}, \bibinfo{person}{Francisco Hern{\'a}ndez-Rodriguez}, {and} \bibinfo{person}{Erik Elmroth}.} \bibinfo{year}{2015}\natexlab{}.
\newblock \showarticletitle{Performance anomaly detection and bottleneck identification}.
\newblock \bibinfo{journal}{\emph{ACM Computing Surveys (CSUR)}} \bibinfo{volume}{48}, \bibinfo{number}{1} (\bibinfo{year}{2015}), \bibinfo{pages}{1--35}.
\newblock


\bibitem[Iglesia and Weyns(2015)]%
        {iglesia2015mape}
\bibfield{author}{\bibinfo{person}{Didac Gil De~La Iglesia} {and} \bibinfo{person}{Danny Weyns}.} \bibinfo{year}{2015}\natexlab{}.
\newblock \showarticletitle{MAPE-K formal templates to rigorously design behaviors for self-adaptive systems}.
\newblock \bibinfo{journal}{\emph{ACM Transactions on Autonomous and Adaptive Systems (TAAS)}} \bibinfo{volume}{10}, \bibinfo{number}{3} (\bibinfo{year}{2015}), \bibinfo{pages}{1--31}.
\newblock


\bibitem[Ilyas and Chu(2019)]%
        {ilyas2019data}
\bibfield{author}{\bibinfo{person}{Ihab~F Ilyas} {and} \bibinfo{person}{Xu Chu}.} \bibinfo{year}{2019}\natexlab{}.
\newblock \bibinfo{booktitle}{\emph{Data cleaning}}.
\newblock \bibinfo{publisher}{Morgan \& Claypool}.
\newblock


\bibitem[ISO/IEC(2022)]%
        {ISO2020AI}
\bibfield{author}{\bibinfo{person}{ISO/IEC}.} \bibinfo{year}{2022}\natexlab{}.
\newblock \bibinfo{booktitle}{\emph{Information technology — Artificial intelligence — Overview of trustworthiness in artificial intelligence}}.
\newblock \bibinfo{type}{{T}echnical {R}eport} 24028:2020. \bibinfo{address}{Geneva, Switzerland}.
\newblock


\bibitem[Jobin et~al\mbox{.}(2019)]%
        {jobin2019global}
\bibfield{author}{\bibinfo{person}{Anna Jobin}, \bibinfo{person}{Marcello Ienca}, {and} \bibinfo{person}{Effy Vayena}.} \bibinfo{year}{2019}\natexlab{}.
\newblock \showarticletitle{The global landscape of AI ethics guidelines}.
\newblock \bibinfo{journal}{\emph{Nature Machine Intelligence}} \bibinfo{volume}{1}, \bibinfo{number}{9} (\bibinfo{year}{2019}), \bibinfo{pages}{389--399}.
\newblock


\bibitem[John et~al\mbox{.}(2021)]%
        {john2021towards}
\bibfield{author}{\bibinfo{person}{Meenu~Mary John}, \bibinfo{person}{Helena~Holmstr{\"o}m Olsson}, {and} \bibinfo{person}{Jan Bosch}.} \bibinfo{year}{2021}\natexlab{}.
\newblock \showarticletitle{Towards mlops: A framework and maturity model}. In \bibinfo{booktitle}{\emph{2021 47th Euromicro Conference on Software Engineering and Advanced Applications (SEAA)}}. IEEE, \bibinfo{pages}{1--8}.
\newblock


\bibitem[Jordan and Mitchell(2015)]%
        {jordan2015machine}
\bibfield{author}{\bibinfo{person}{Michael~I Jordan} {and} \bibinfo{person}{Tom~M Mitchell}.} \bibinfo{year}{2015}\natexlab{}.
\newblock \showarticletitle{Machine learning: Trends, perspectives, and prospects}.
\newblock \bibinfo{journal}{\emph{Science}} \bibinfo{volume}{349}, \bibinfo{number}{6245} (\bibinfo{year}{2015}), \bibinfo{pages}{255--260}.
\newblock


\bibitem[Karamitsos et~al\mbox{.}(2020)]%
        {karamitsos2020applying}
\bibfield{author}{\bibinfo{person}{Ioannis Karamitsos}, \bibinfo{person}{Saeed Albarhami}, {and} \bibinfo{person}{Charalampos Apostolopoulos}.} \bibinfo{year}{2020}\natexlab{}.
\newblock \showarticletitle{Applying DevOps practices of continuous automation for machine learning}.
\newblock \bibinfo{journal}{\emph{Information}} \bibinfo{volume}{11}, \bibinfo{number}{7} (\bibinfo{year}{2020}), \bibinfo{pages}{363}.
\newblock


\bibitem[Karla\v{s} et~al\mbox{.}(2020)]%
        {karlavs2020nearest}
\bibfield{author}{\bibinfo{person}{Bojan Karla\v{s}}, \bibinfo{person}{Peng Li}, \bibinfo{person}{Renzhi Wu}, \bibinfo{person}{Nezihe~Merve G\"{u}rel}, \bibinfo{person}{Xu Chu}, \bibinfo{person}{Wentao Wu}, {and} \bibinfo{person}{Ce Zhang}.} \bibinfo{year}{2020}\natexlab{}.
\newblock \showarticletitle{Nearest Neighbor Classifiers over Incomplete Information: From Certain Answers to Certain Predictions}.
\newblock \bibinfo{journal}{\emph{Proc. VLDB Endow.}} \bibinfo{volume}{14}, \bibinfo{number}{3} (\bibinfo{date}{nov} \bibinfo{year}{2020}), \bibinfo{pages}{255–267}.
\newblock


\bibitem[Karmakar et~al\mbox{.}(2021)]%
        {karmakar2021assessing}
\bibfield{author}{\bibinfo{person}{Gour Karmakar}, \bibinfo{person}{Abdullahi Chowdhury}, \bibinfo{person}{Rajkumar Das}, \bibinfo{person}{Joarder Kamruzzaman}, {and} \bibinfo{person}{Syed Islam}.} \bibinfo{year}{2021}\natexlab{}.
\newblock \showarticletitle{Assessing trust level of a driverless car using deep learning}.
\newblock \bibinfo{journal}{\emph{IEEE Transactions on Intelligent Transportation Systems}} \bibinfo{volume}{22}, \bibinfo{number}{7} (\bibinfo{year}{2021}), \bibinfo{pages}{4457--4466}.
\newblock


\bibitem[Kaur et~al\mbox{.}(2020)]%
        {kaur2020requirements}
\bibfield{author}{\bibinfo{person}{Davinder Kaur}, \bibinfo{person}{Suleyman Uslu}, {and} \bibinfo{person}{Arjan Durresi}.} \bibinfo{year}{2020}\natexlab{}.
\newblock \showarticletitle{Requirements for trustworthy artificial intelligence--a review}. In \bibinfo{booktitle}{\emph{International Conference on Network-Based Information Systems}}. Springer, \bibinfo{pages}{105--115}.
\newblock


\bibitem[Kaur et~al\mbox{.}(2022)]%
        {kaur2022trustworthy}
\bibfield{author}{\bibinfo{person}{Davinder Kaur}, \bibinfo{person}{Suleyman Uslu}, \bibinfo{person}{Kaley~J Rittichier}, {and} \bibinfo{person}{Arjan Durresi}.} \bibinfo{year}{2022}\natexlab{}.
\newblock \showarticletitle{Trustworthy artificial intelligence: a review}.
\newblock \bibinfo{journal}{\emph{ACM Computing Surveys (CSUR)}} \bibinfo{volume}{55}, \bibinfo{number}{2} (\bibinfo{year}{2022}), \bibinfo{pages}{1--38}.
\newblock


\bibitem[Kawahara and Sugiyama(2012)]%
        {Kawahara201sequential}
\bibfield{author}{\bibinfo{person}{Yoshinobu Kawahara} {and} \bibinfo{person}{Masashi Sugiyama}.} \bibinfo{year}{2012}\natexlab{}.
\newblock \showarticletitle{Sequential change-point detection based on direct density-ratio estimation}.
\newblock \bibinfo{journal}{\emph{Statistical Analysis and Data Mining: The ASA Data Science Journal}} \bibinfo{volume}{5}, \bibinfo{number}{2} (\bibinfo{year}{2012}), \bibinfo{pages}{114--127}.
\newblock


\bibitem[Keskar and Socher(2017)]%
        {keskar2017improving}
\bibfield{author}{\bibinfo{person}{Nitish~Shirish Keskar} {and} \bibinfo{person}{Richard Socher}.} \bibinfo{year}{2017}\natexlab{}.
\newblock \showarticletitle{Improving generalization performance by switching from adam to sgd}.
\newblock \bibinfo{journal}{\emph{arXiv preprint arXiv:1712.07628}} (\bibinfo{year}{2017}).
\newblock


\bibitem[Kitano(2007)]%
        {kitano2007towards}
\bibfield{author}{\bibinfo{person}{Hiroaki Kitano}.} \bibinfo{year}{2007}\natexlab{}.
\newblock \bibinfo{title}{Towards a theory of biological robustness}.
\newblock , \bibinfo{numpages}{137}~pages.
\newblock


\bibitem[Klaise et~al\mbox{.}(2020)]%
        {klaise2020monitoring}
\bibfield{author}{\bibinfo{person}{Janis Klaise}, \bibinfo{person}{Arnaud Van~Looveren}, \bibinfo{person}{Clive Cox}, \bibinfo{person}{Giovanni Vacanti}, {and} \bibinfo{person}{Alexandru Coca}.} \bibinfo{year}{2020}\natexlab{}.
\newblock \showarticletitle{Monitoring and explainability of models in production}.
\newblock \bibinfo{journal}{\emph{arXiv preprint arXiv:2007.06299}} (\bibinfo{year}{2020}).
\newblock


\bibitem[Koziarski et~al\mbox{.}(2020)]%
        {koziarski2020combined}
\bibfield{author}{\bibinfo{person}{Micha{\l} Koziarski}, \bibinfo{person}{Micha{\l} Wo{\'z}niak}, {and} \bibinfo{person}{Bartosz Krawczyk}.} \bibinfo{year}{2020}\natexlab{}.
\newblock \showarticletitle{Combined cleaning and resampling algorithm for multi-class imbalanced data with label noise}.
\newblock \bibinfo{journal}{\emph{Knowledge-Based Systems}}  \bibinfo{volume}{204} (\bibinfo{year}{2020}), \bibinfo{pages}{106223}.
\newblock


\bibitem[Kreuzberger et~al\mbox{.}(2023)]%
        {kreuzberger2022machine}
\bibfield{author}{\bibinfo{person}{Dominik Kreuzberger}, \bibinfo{person}{Niklas K{\"u}hl}, {and} \bibinfo{person}{Sebastian Hirschl}.} \bibinfo{year}{2023}\natexlab{}.
\newblock \showarticletitle{Machine learning operations (mlops): Overview, definition, and architecture}.
\newblock \bibinfo{journal}{\emph{IEEE access}} (\bibinfo{year}{2023}).
\newblock


\bibitem[Kumar et~al\mbox{.}(2020)]%
        {kumar2020trustworthy}
\bibfield{author}{\bibinfo{person}{Abhishek Kumar}, \bibinfo{person}{Tristan Braud}, \bibinfo{person}{Sasu Tarkoma}, {and} \bibinfo{person}{Pan Hui}.} \bibinfo{year}{2020}\natexlab{}.
\newblock \showarticletitle{Trustworthy AI in the age of pervasive computing and big data}. In \bibinfo{booktitle}{\emph{2020 IEEE International Conference on Pervasive Computing and Communications Workshops (PerCom Workshops)}}. IEEE, \bibinfo{pages}{1--6}.
\newblock


\bibitem[Kurian et~al\mbox{.}(2021)]%
        {kurian2021boat}
\bibfield{author}{\bibinfo{person}{John~Joy Kurian}, \bibinfo{person}{Marcel Dix}, \bibinfo{person}{Ido Amihai}, \bibinfo{person}{Glenn Ceusters}, {and} \bibinfo{person}{Ajinkya Prabhune}.} \bibinfo{year}{2021}\natexlab{}.
\newblock \showarticletitle{BOAT: A bayesian optimization automl time-series framework for industrial applications}. In \bibinfo{booktitle}{\emph{2021 IEEE Seventh International Conference on Big Data Computing Service and Applications (BigDataService)}}. IEEE, \bibinfo{pages}{17--24}.
\newblock


\bibitem[Larracy et~al\mbox{.}(2021)]%
        {larracy2021machine}
\bibfield{author}{\bibinfo{person}{Robyn Larracy}, \bibinfo{person}{Angkoon Phinyomark}, {and} \bibinfo{person}{Erik Scheme}.} \bibinfo{year}{2021}\natexlab{}.
\newblock \showarticletitle{Machine learning model validation for early stage studies with small sample sizes}. In \bibinfo{booktitle}{\emph{2021 43rd Annual International Conference of the IEEE Engineering in Medicine \& Biology Society (EMBC)}}. IEEE, \bibinfo{pages}{2314--2319}.
\newblock


\bibitem[Li et~al\mbox{.}(2022)]%
        {li2021trustworthy}
\bibfield{author}{\bibinfo{person}{Bo Li}, \bibinfo{person}{Peng Qi}, \bibinfo{person}{Bo Liu}, \bibinfo{person}{Shuai Di}, \bibinfo{person}{Jingen Liu}, \bibinfo{person}{Jiquan Pei}, \bibinfo{person}{Jinfeng Yi}, {and} \bibinfo{person}{Bowen Zhou}.} \bibinfo{year}{2022}\natexlab{}.
\newblock \showarticletitle{Trustworthy AI: From Principles to Practices}.
\newblock \bibinfo{journal}{\emph{ACM Comput. Surv.}} (\bibinfo{date}{aug} \bibinfo{year}{2022}).
\newblock
\showISSN{0360-0300}
\urldef\tempurl%
\url{https://doi.org/10.1145/3555803}
\showDOI{\tempurl}


\bibitem[Li and Gui(2020)]%
        {li2020cms}
\bibfield{author}{\bibinfo{person}{KeDi Li} {and} \bibinfo{person}{Ning Gui}.} \bibinfo{year}{2020}\natexlab{}.
\newblock \showarticletitle{CMS: A continuous machine-learning and serving platform for industrial big data}.
\newblock \bibinfo{journal}{\emph{Future Internet}} \bibinfo{volume}{12}, \bibinfo{number}{6} (\bibinfo{year}{2020}), \bibinfo{pages}{102}.
\newblock


\bibitem[Liberty et~al\mbox{.}(2020)]%
        {liberty2020elastic}
\bibfield{author}{\bibinfo{person}{Edo Liberty}, \bibinfo{person}{Zohar Karnin}, \bibinfo{person}{Bing Xiang}, \bibinfo{person}{Laurence Rouesnel}, \bibinfo{person}{Baris Coskun}, \bibinfo{person}{Ramesh Nallapati}, \bibinfo{person}{Julio Delgado}, \bibinfo{person}{Amir Sadoughi}, \bibinfo{person}{Yury Astashonok}, \bibinfo{person}{Piali Das}, {et~al\mbox{.}}} \bibinfo{year}{2020}\natexlab{}.
\newblock \showarticletitle{Elastic machine learning algorithms in amazon sagemaker}. In \bibinfo{booktitle}{\emph{Proceedings of the 2020 ACM SIGMOD International Conference on Management of Data}}. \bibinfo{pages}{731--737}.
\newblock


\bibitem[Liu et~al\mbox{.}(2022)]%
        {liu2021trustworthy}
\bibfield{author}{\bibinfo{person}{Haochen Liu}, \bibinfo{person}{Yiqi Wang}, \bibinfo{person}{Wenqi Fan}, \bibinfo{person}{Xiaorui Liu}, \bibinfo{person}{Yaxin Li}, \bibinfo{person}{Shaili Jain}, \bibinfo{person}{Yunhao Liu}, \bibinfo{person}{Anil~K. Jain}, {and} \bibinfo{person}{Jiliang Tang}.} \bibinfo{year}{2022}\natexlab{}.
\newblock \showarticletitle{Trustworthy AI: A Computational Perspective}.
\newblock \bibinfo{journal}{\emph{ACM Trans. Intell. Syst. Technol.}} (\bibinfo{date}{jun} \bibinfo{year}{2022}).
\newblock
\showISSN{2157-6904}


\bibitem[Lu et~al\mbox{.}(2019)]%
        {Lu2016LearningUnder}
\bibfield{author}{\bibinfo{person}{Jie Lu}, \bibinfo{person}{Anjin Liu}, \bibinfo{person}{Fan Dong}, \bibinfo{person}{Feng Gu}, \bibinfo{person}{João Gama}, {and} \bibinfo{person}{Guangquan Zhang}.} \bibinfo{year}{2019}\natexlab{}.
\newblock \showarticletitle{Learning under Concept Drift: A Review}.
\newblock \bibinfo{journal}{\emph{IEEE Transactions on Knowledge and Data Engineering}} \bibinfo{volume}{31}, \bibinfo{number}{12} (\bibinfo{year}{2019}), \bibinfo{pages}{2346--2363}.
\newblock


\bibitem[Lu et~al\mbox{.}(2017)]%
        {lu2017dynamic}
\bibfield{author}{\bibinfo{person}{Yang Lu}, \bibinfo{person}{Yiu-ming Cheung}, {and} \bibinfo{person}{Yuan~Yan Tang}.} \bibinfo{year}{2017}\natexlab{}.
\newblock \showarticletitle{Dynamic Weighted Majority for Incremental Learning of Imbalanced Data Streams with Concept Drift.}. In \bibinfo{booktitle}{\emph{IJCAI}}. \bibinfo{pages}{2393--2399}.
\newblock


\bibitem[Luo(2016)]%
        {luo2016review}
\bibfield{author}{\bibinfo{person}{Gang Luo}.} \bibinfo{year}{2016}\natexlab{}.
\newblock \showarticletitle{A review of automatic selection methods for machine learning algorithms and hyper-parameter values}.
\newblock \bibinfo{journal}{\emph{Network Modeling Analysis in Health Informatics and Bioinformatics}} \bibinfo{volume}{5}, \bibinfo{number}{1} (\bibinfo{year}{2016}), \bibinfo{pages}{1--16}.
\newblock


\bibitem[Luo et~al\mbox{.}(2018)]%
        {luo20183d}
\bibfield{author}{\bibinfo{person}{Yuan Luo}, \bibinfo{person}{Peter Szolovits}, \bibinfo{person}{Anand~S Dighe}, {and} \bibinfo{person}{Jason~M Baron}.} \bibinfo{year}{2018}\natexlab{}.
\newblock \showarticletitle{3D-MICE: integration of cross-sectional and longitudinal imputation for multi-analyte longitudinal clinical data}.
\newblock \bibinfo{journal}{\emph{Journal of the American Medical Informatics Association}} \bibinfo{volume}{25}, \bibinfo{number}{6} (\bibinfo{year}{2018}), \bibinfo{pages}{645--653}.
\newblock


\bibitem[Lusby et~al\mbox{.}(2018)]%
        {lusby2018survey}
\bibfield{author}{\bibinfo{person}{Richard~M Lusby}, \bibinfo{person}{Jesper Larsen}, {and} \bibinfo{person}{Simon Bull}.} \bibinfo{year}{2018}\natexlab{}.
\newblock \showarticletitle{A survey on robustness in railway planning}.
\newblock \bibinfo{journal}{\emph{European Journal of Operational Research}} \bibinfo{volume}{266}, \bibinfo{number}{1} (\bibinfo{year}{2018}), \bibinfo{pages}{1--15}.
\newblock


\bibitem[Lyu et~al\mbox{.}(2021)]%
        {lyu2021synthesizing}
\bibfield{author}{\bibinfo{person}{Yuting Lyu}, \bibinfo{person}{Junghui Chen}, {and} \bibinfo{person}{Zhihuan Song}.} \bibinfo{year}{2021}\natexlab{}.
\newblock \showarticletitle{Synthesizing labeled data to enhance soft sensor performance in data-scarce regions}.
\newblock \bibinfo{journal}{\emph{Control Engineering Practice}}  \bibinfo{volume}{115} (\bibinfo{year}{2021}), \bibinfo{pages}{104903}.
\newblock


\bibitem[Ma et~al\mbox{.}(2022)]%
        {ma2022towards}
\bibfield{author}{\bibinfo{person}{J Ma}, \bibinfo{person}{L Schneider}, \bibinfo{person}{S Lapuschkin}, \bibinfo{person}{R Achtibat}, \bibinfo{person}{M Duchrau}, \bibinfo{person}{J Krois}, \bibinfo{person}{F Schwendicke}, {and} \bibinfo{person}{W Samek}.} \bibinfo{year}{2022}\natexlab{}.
\newblock \showarticletitle{Towards Trustworthy AI in Dentistry}.
\newblock \bibinfo{journal}{\emph{Journal of Dental Research}} (\bibinfo{year}{2022}), \bibinfo{pages}{00220345221106086}.
\newblock


\bibitem[Mahajan et~al\mbox{.}(2021)]%
        {mahajan2021connection}
\bibfield{author}{\bibinfo{person}{Divyat Mahajan}, \bibinfo{person}{Shruti Tople}, {and} \bibinfo{person}{Amit Sharma}.} \bibinfo{year}{2021}\natexlab{}.
\newblock \showarticletitle{The Connection between Out-of-Distribution Generalization and Privacy of ML Models}.
\newblock \bibinfo{journal}{\emph{arXiv preprint arXiv:2110.03369}} (\bibinfo{year}{2021}).
\newblock


\bibitem[Mahdavinejad et~al\mbox{.}(2018)]%
        {mahdavinejad2018machine}
\bibfield{author}{\bibinfo{person}{Mohammad~Saeid Mahdavinejad}, \bibinfo{person}{Mohammadreza Rezvan}, \bibinfo{person}{Mohammadamin Barekatain}, \bibinfo{person}{Peyman Adibi}, \bibinfo{person}{Payam Barnaghi}, {and} \bibinfo{person}{Amit~P Sheth}.} \bibinfo{year}{2018}\natexlab{}.
\newblock \showarticletitle{Machine learning for Internet of Things data analysis: A survey}.
\newblock \bibinfo{journal}{\emph{Digital Communications and Networks}} \bibinfo{volume}{4}, \bibinfo{number}{3} (\bibinfo{year}{2018}), \bibinfo{pages}{161--175}.
\newblock


\bibitem[M{\"a}kinen et~al\mbox{.}(2021)]%
        {makinen2021needs}
\bibfield{author}{\bibinfo{person}{Sasu M{\"a}kinen}, \bibinfo{person}{Henrik Skogstr{\"o}m}, \bibinfo{person}{Eero Laaksonen}, {and} \bibinfo{person}{Tommi Mikkonen}.} \bibinfo{year}{2021}\natexlab{}.
\newblock \showarticletitle{Who needs MLOps: What data scientists seek to accomplish and how can MLOps help?}. In \bibinfo{booktitle}{\emph{2021 IEEE/ACM 1st Workshop on AI Engineering-Software Engineering for AI (WAIN)}}. IEEE, \bibinfo{pages}{109--112}.
\newblock


\bibitem[Mallick et~al\mbox{.}(2022)]%
        {mallick2022matchmaker}
\bibfield{author}{\bibinfo{person}{Ankur Mallick}, \bibinfo{person}{Kevin Hsieh}, \bibinfo{person}{Behnaz Arzani}, {and} \bibinfo{person}{Gauri Joshi}.} \bibinfo{year}{2022}\natexlab{}.
\newblock \showarticletitle{Matchmaker: Data Drift Mitigation in Machine Learning for Large-Scale Systems}.
\newblock \bibinfo{journal}{\emph{Proceedings of Machine Learning and Systems}}  \bibinfo{volume}{4} (\bibinfo{year}{2022}), \bibinfo{pages}{77--94}.
\newblock


\bibitem[Masood and Hashmi(2019)]%
        {masood2019aiops}
\bibfield{author}{\bibinfo{person}{Adnan Masood} {and} \bibinfo{person}{Adnan Hashmi}.} \bibinfo{year}{2019}\natexlab{}.
\newblock \showarticletitle{AIOps: predictive analytics \& machine learning in operations}.
\newblock In \bibinfo{booktitle}{\emph{Cognitive Computing Recipes}}. \bibinfo{publisher}{Springer}, \bibinfo{pages}{359--382}.
\newblock


\bibitem[Mathew et~al\mbox{.}(2021)]%
        {mathew2021defraudnet}
\bibfield{author}{\bibinfo{person}{Jose Mathew}, \bibinfo{person}{Meghana Negi}, \bibinfo{person}{Rutvik Vijjali}, {and} \bibinfo{person}{Jairaj Sathyanarayana}.} \bibinfo{year}{2021}\natexlab{}.
\newblock \showarticletitle{DeFraudNet: An End-to-End Weak Supervision Framework to Detect Fraud in Online Food Delivery}. In \bibinfo{booktitle}{\emph{Joint European Conference on Machine Learning and Knowledge Discovery in Databases}}. Springer, \bibinfo{pages}{85--99}.
\newblock


\bibitem[Matsui and Goya(2022)]%
        {matsui2022mlops}
\bibfield{author}{\bibinfo{person}{Beatriz~MA Matsui} {and} \bibinfo{person}{Denise~H Goya}.} \bibinfo{year}{2022}\natexlab{}.
\newblock \showarticletitle{MLOps: A Guide to its Adoption in the Context of Responsible AI}. In \bibinfo{booktitle}{\emph{2022 IEEE/ACM 1st International Workshop on Software Engineering for Responsible Artificial Intelligence (SE4RAI)}}. IEEE, \bibinfo{pages}{45--49}.
\newblock


\bibitem[McGraw et~al\mbox{.}(2020)]%
        {mcgraw2020top}
\bibfield{author}{\bibinfo{person}{Gary McGraw}, \bibinfo{person}{Richie Bonett}, \bibinfo{person}{Victor Shepardson}, {and} \bibinfo{person}{Harold Figueroa}.} \bibinfo{year}{2020}\natexlab{}.
\newblock \showarticletitle{The Top 10 Risks of Machine Learning Security}.
\newblock \bibinfo{journal}{\emph{Computer}} \bibinfo{volume}{53}, \bibinfo{number}{6} (\bibinfo{year}{2020}), \bibinfo{pages}{57--61}.
\newblock


\bibitem[Medhat et~al\mbox{.}(2020)]%
        {medhat2020framework}
\bibfield{author}{\bibinfo{person}{Noha Medhat}, \bibinfo{person}{Sherin~M Moussa}, \bibinfo{person}{Nagwa~Lotfy Badr}, {and} \bibinfo{person}{Mohamed~F Tolba}.} \bibinfo{year}{2020}\natexlab{}.
\newblock \showarticletitle{A framework for continuous regression and integration testing in IoT systems based on deep learning and search-based techniques}.
\newblock \bibinfo{journal}{\emph{IEEE Access}}  \bibinfo{volume}{8} (\bibinfo{year}{2020}), \bibinfo{pages}{215716--215726}.
\newblock


\bibitem[Meister et~al\mbox{.}(2020)]%
        {meister2020maggy}
\bibfield{author}{\bibinfo{person}{Moritz Meister}, \bibinfo{person}{Sina Sheikholeslami}, \bibinfo{person}{Amir~H Payberah}, \bibinfo{person}{Vladimir Vlassov}, {and} \bibinfo{person}{Jim Dowling}.} \bibinfo{year}{2020}\natexlab{}.
\newblock \showarticletitle{Maggy: Scalable asynchronous parallel hyperparameter search}. In \bibinfo{booktitle}{\emph{Proceedings of the 1st Workshop on Distributed Machine Learning}}. \bibinfo{pages}{28--33}.
\newblock


\bibitem[Meng et~al\mbox{.}(2020)]%
        {meng2020survey}
\bibfield{author}{\bibinfo{person}{Tong Meng}, \bibinfo{person}{Xuyang Jing}, \bibinfo{person}{Zheng Yan}, {and} \bibinfo{person}{Witold Pedrycz}.} \bibinfo{year}{2020}\natexlab{}.
\newblock \showarticletitle{A survey on machine learning for data fusion}.
\newblock \bibinfo{journal}{\emph{Information Fusion}}  \bibinfo{volume}{57} (\bibinfo{year}{2020}), \bibinfo{pages}{115--129}.
\newblock


\bibitem[Min et~al\mbox{.}(2021)]%
        {min2021sensix++}
\bibfield{author}{\bibinfo{person}{Chulhong Min}, \bibinfo{person}{Akhil Mathur}, \bibinfo{person}{Utku~Gunay Acer}, \bibinfo{person}{Alessandro Montanari}, {and} \bibinfo{person}{Fahim Kawsar}.} \bibinfo{year}{2021}\natexlab{}.
\newblock \showarticletitle{SensiX++: Bringing mlops and multi-tenant model serving to sensory edge devices}.
\newblock \bibinfo{journal}{\emph{arXiv preprint arXiv:2109.03947}} (\bibinfo{year}{2021}).
\newblock


\bibitem[Mosqueira-Rey et~al\mbox{.}(2023)]%
        {mosqueira2023human}
\bibfield{author}{\bibinfo{person}{Eduardo Mosqueira-Rey}, \bibinfo{person}{Elena Hern{\'a}ndez-Pereira}, \bibinfo{person}{David Alonso-R{\'\i}os}, \bibinfo{person}{Jos{\'e} Bobes-Bascar{\'a}n}, {and} \bibinfo{person}{{\'A}ngel Fern{\'a}ndez-Leal}.} \bibinfo{year}{2023}\natexlab{}.
\newblock \showarticletitle{Human-in-the-loop machine learning: a state of the art}.
\newblock \bibinfo{journal}{\emph{Artificial Intelligence Review}} \bibinfo{volume}{56}, \bibinfo{number}{4} (\bibinfo{year}{2023}), \bibinfo{pages}{3005--3054}.
\newblock


\bibitem[Moulton et~al\mbox{.}(2018)]%
        {moulton2018clustering}
\bibfield{author}{\bibinfo{person}{Richard~Hugh Moulton}, \bibinfo{person}{Herna~L Viktor}, \bibinfo{person}{Nathalie Japkowicz}, {and} \bibinfo{person}{Jo{\~a}o Gama}.} \bibinfo{year}{2018}\natexlab{}.
\newblock \showarticletitle{Clustering in the presence of concept drift}. In \bibinfo{booktitle}{\emph{Joint European Conference on Machine Learning and Knowledge Discovery in Databases}}. Springer, \bibinfo{pages}{339--355}.
\newblock


\bibitem[Munappy et~al\mbox{.}(2020)]%
        {munappy2020ad}
\bibfield{author}{\bibinfo{person}{Aiswarya~Raj Munappy}, \bibinfo{person}{David~Issa Mattos}, \bibinfo{person}{Jan Bosch}, \bibinfo{person}{Helena~Holmstr{\"o}m Olsson}, {and} \bibinfo{person}{Anas Dakkak}.} \bibinfo{year}{2020}\natexlab{}.
\newblock \showarticletitle{From ad-hoc data analytics to dataops}. In \bibinfo{booktitle}{\emph{Proceedings of the International Conference on Software and System Processes}}. \bibinfo{pages}{165--174}.
\newblock


\bibitem[Olson(2003)]%
        {olson2003data}
\bibfield{author}{\bibinfo{person}{Jack~E Olson}.} \bibinfo{year}{2003}\natexlab{}.
\newblock \bibinfo{booktitle}{\emph{Data quality: the accuracy dimension}}.
\newblock \bibinfo{publisher}{Elsevier}.
\newblock


\bibitem[Oreizy et~al\mbox{.}(1999)]%
        {oreizy1999architecture}
\bibfield{author}{\bibinfo{person}{Peyman Oreizy}, \bibinfo{person}{Michael~M Gorlick}, \bibinfo{person}{Richard~N Taylor}, \bibinfo{person}{Dennis Heimhigner}, \bibinfo{person}{Gregory Johnson}, \bibinfo{person}{Nenad Medvidovic}, \bibinfo{person}{Alex Quilici}, \bibinfo{person}{David~S Rosenblum}, {and} \bibinfo{person}{Alexander~L Wolf}.} \bibinfo{year}{1999}\natexlab{}.
\newblock \showarticletitle{An architecture-based approach to self-adaptive software}.
\newblock \bibinfo{journal}{\emph{IEEE Intelligent Systems and Their Applications}} \bibinfo{volume}{14}, \bibinfo{number}{3} (\bibinfo{year}{1999}), \bibinfo{pages}{54--62}.
\newblock


\bibitem[Park et~al\mbox{.}(2021)]%
        {park2021facilitating}
\bibfield{author}{\bibinfo{person}{Soya Park}, \bibinfo{person}{April~Yi Wang}, \bibinfo{person}{Ban Kawas}, \bibinfo{person}{Q~Vera Liao}, \bibinfo{person}{David Piorkowski}, {and} \bibinfo{person}{Marina Danilevsky}.} \bibinfo{year}{2021}\natexlab{}.
\newblock \showarticletitle{Facilitating knowledge sharing from domain experts to data scientists for building nlp models}. In \bibinfo{booktitle}{\emph{26th International Conference on Intelligent User Interfaces}}. \bibinfo{pages}{585--596}.
\newblock


\bibitem[Qian and Wiederhold(1986)]%
        {qian1986knowledge}
\bibfield{author}{\bibinfo{person}{Xiaolei Qian} {and} \bibinfo{person}{Gio Wiederhold}.} \bibinfo{year}{1986}\natexlab{}.
\newblock \showarticletitle{Knowledge-based Integrity Constraint Validation.}. In \bibinfo{booktitle}{\emph{VLDB}}, Vol.~\bibinfo{volume}{86}. Citeseer, \bibinfo{pages}{25--28}.
\newblock


\bibitem[Quionero-Candela et~al\mbox{.}(2009)]%
        {Candela2009DatasetS}
\bibfield{author}{\bibinfo{person}{Joaquin Quionero-Candela}, \bibinfo{person}{Masashi Sugiyama}, \bibinfo{person}{Anton Schwaighofer}, {and} \bibinfo{person}{Neil~D. Lawrence}.} \bibinfo{year}{2009}\natexlab{}.
\newblock \bibinfo{booktitle}{\emph{Dataset Shift in Machine Learning}}.
\newblock \bibinfo{publisher}{The MIT Press}.
\newblock
\showISBNx{0262170051}


\bibitem[Racca and Magri(2021)]%
        {racca2021robust}
\bibfield{author}{\bibinfo{person}{Alberto Racca} {and} \bibinfo{person}{Luca Magri}.} \bibinfo{year}{2021}\natexlab{}.
\newblock \showarticletitle{Robust Optimization and Validation of Echo State Networks for learning chaotic dynamics}.
\newblock \bibinfo{journal}{\emph{Neural Networks}}  \bibinfo{volume}{142} (\bibinfo{year}{2021}), \bibinfo{pages}{252--268}.
\newblock


\bibitem[Raj(2021)]%
        {raj2021engineering}
\bibfield{author}{\bibinfo{person}{Emmanuel Raj}.} \bibinfo{year}{2021}\natexlab{}.
\newblock \bibinfo{booktitle}{\emph{Engineering MLOps: Rapidly build, test, and manage production-ready machine learning life cycles at scale}}.
\newblock \bibinfo{publisher}{Packt Publishing Ltd}.
\newblock


\bibitem[Raj et~al\mbox{.}(2021)]%
        {raj2021edge}
\bibfield{author}{\bibinfo{person}{Emmanuel Raj}, \bibinfo{person}{David Buffoni}, \bibinfo{person}{Magnus Westerlund}, {and} \bibinfo{person}{Kimmo Ahola}.} \bibinfo{year}{2021}\natexlab{}.
\newblock \showarticletitle{Edge MLOps: An automation framework for aiot applications}. In \bibinfo{booktitle}{\emph{2021 IEEE International Conference on Cloud Engineering (IC2E)}}. IEEE, \bibinfo{pages}{191--200}.
\newblock


\bibitem[Raschka and Mirjalili(2019)]%
        {raschka2019python}
\bibfield{author}{\bibinfo{person}{Sebastian Raschka} {and} \bibinfo{person}{Vahid Mirjalili}.} \bibinfo{year}{2019}\natexlab{}.
\newblock \bibinfo{booktitle}{\emph{Python machine learning: Machine learning and deep learning with Python, scikit-learn, and TensorFlow 2}}.
\newblock \bibinfo{publisher}{Packt Publishing Ltd}.
\newblock


\bibitem[Ratner et~al\mbox{.}(2017)]%
        {ratner2017snorkel}
\bibfield{author}{\bibinfo{person}{Alexander Ratner}, \bibinfo{person}{Stephen~H Bach}, \bibinfo{person}{Henry Ehrenberg}, \bibinfo{person}{Jason Fries}, \bibinfo{person}{Sen Wu}, {and} \bibinfo{person}{Christopher R{\'e}}.} \bibinfo{year}{2017}\natexlab{}.
\newblock \showarticletitle{Snorkel: Rapid training data creation with weak supervision}. In \bibinfo{booktitle}{\emph{Proceedings of the VLDB Endowment. International Conference on Very Large Data Bases}}, Vol.~\bibinfo{volume}{11}. NIH Public Access, \bibinfo{pages}{269}.
\newblock


\bibitem[R{\'e} et~al\mbox{.}(2019)]%
        {re2019overton}
\bibfield{author}{\bibinfo{person}{Christopher R{\'e}}, \bibinfo{person}{Feng Niu}, \bibinfo{person}{Pallavi Gudipati}, {and} \bibinfo{person}{Charles Srisuwananukorn}.} \bibinfo{year}{2019}\natexlab{}.
\newblock \showarticletitle{Overton: A data system for monitoring and improving machine-learned products}.
\newblock \bibinfo{journal}{\emph{arXiv preprint arXiv:1909.05372}} (\bibinfo{year}{2019}).
\newblock


\bibitem[Ren et~al\mbox{.}(2018)]%
        {ren2018learning}
\bibfield{author}{\bibinfo{person}{Mengye Ren}, \bibinfo{person}{Wenyuan Zeng}, \bibinfo{person}{Bin Yang}, {and} \bibinfo{person}{Raquel Urtasun}.} \bibinfo{year}{2018}\natexlab{}.
\newblock \showarticletitle{Learning to reweight examples for robust deep learning}. In \bibinfo{booktitle}{\emph{International conference on machine learning}}. PMLR, \bibinfo{pages}{4334--4343}.
\newblock


\bibitem[Renggli et~al\mbox{.}(2021)]%
        {renggli2021data}
\bibfield{author}{\bibinfo{person}{Cedric Renggli}, \bibinfo{person}{Luka Rimanic}, \bibinfo{person}{Nezihe~Merve G{\"u}rel}, \bibinfo{person}{Bojan Karla{\v{s}}}, \bibinfo{person}{Wentao Wu}, {and} \bibinfo{person}{Ce Zhang}.} \bibinfo{year}{2021}\natexlab{}.
\newblock \showarticletitle{A data quality-driven view of mlops}.
\newblock \bibinfo{journal}{\emph{arXiv preprint arXiv:2102.07750}} (\bibinfo{year}{2021}).
\newblock


\bibitem[Renggli et~al\mbox{.}(2020)]%
        {renggli2020ease}
\bibfield{author}{\bibinfo{person}{Cedric Renggli}, \bibinfo{person}{Luka Rimanic}, \bibinfo{person}{Luka Kolar}, \bibinfo{person}{Nora Hollenstein}, \bibinfo{person}{Wentao Wu}, {and} \bibinfo{person}{Ce Zhang}.} \bibinfo{year}{2020}\natexlab{}.
\newblock \showarticletitle{Ease. ml/snoopy: Towards Automatic Feasibility Study for Machine Learning Applications}.
\newblock \bibinfo{journal}{\emph{arXiv preprint arXiv:2010.08410}} (\bibinfo{year}{2020}).
\newblock


\bibitem[Rozsa et~al\mbox{.}(2016)]%
        {rozsa2016accuracy}
\bibfield{author}{\bibinfo{person}{Andras Rozsa}, \bibinfo{person}{Manuel G{\"u}nther}, {and} \bibinfo{person}{Terrance~E Boult}.} \bibinfo{year}{2016}\natexlab{}.
\newblock \showarticletitle{Are accuracy and robustness correlated}. In \bibinfo{booktitle}{\emph{2016 15th IEEE international conference on machine learning and applications (ICMLA)}}. IEEE, \bibinfo{pages}{227--232}.
\newblock


\bibitem[Rudin and Wagstaff(2014)]%
        {rudin2014machine}
\bibfield{author}{\bibinfo{person}{Cynthia Rudin} {and} \bibinfo{person}{Kiri~L Wagstaff}.} \bibinfo{year}{2014}\natexlab{}.
\newblock \bibinfo{title}{Machine learning for science and society}.
\newblock , \bibinfo{numpages}{9}~pages.
\newblock


\bibitem[Ruf et~al\mbox{.}(2021)]%
        {ruf2021demystifying}
\bibfield{author}{\bibinfo{person}{Philipp Ruf}, \bibinfo{person}{Manav Madan}, \bibinfo{person}{Christoph Reich}, {and} \bibinfo{person}{Djaffar Ould-Abdeslam}.} \bibinfo{year}{2021}\natexlab{}.
\newblock \showarticletitle{Demystifying mlops and presenting a recipe for the selection of open-source tools}.
\newblock \bibinfo{journal}{\emph{Applied Sciences}} \bibinfo{volume}{11}, \bibinfo{number}{19} (\bibinfo{year}{2021}), \bibinfo{pages}{8861}.
\newblock


\bibitem[Sarker(2021)]%
        {sarker2021machine}
\bibfield{author}{\bibinfo{person}{Iqbal~H Sarker}.} \bibinfo{year}{2021}\natexlab{}.
\newblock \showarticletitle{Machine learning: Algorithms, real-world applications and research directions}.
\newblock \bibinfo{journal}{\emph{SN Computer Science}} \bibinfo{volume}{2}, \bibinfo{number}{3} (\bibinfo{year}{2021}), \bibinfo{pages}{1--21}.
\newblock


\bibitem[Sarker(2022)]%
        {sarker2022ai}
\bibfield{author}{\bibinfo{person}{Iqbal~H Sarker}.} \bibinfo{year}{2022}\natexlab{}.
\newblock \showarticletitle{A{I}-based modeling: Techniques, applications and research issues towards automation, intelligent and smart systems}.
\newblock \bibinfo{journal}{\emph{SN Computer Science}} \bibinfo{volume}{3}, \bibinfo{number}{2} (\bibinfo{year}{2022}), \bibinfo{pages}{1--20}.
\newblock


\bibitem[Schelter et~al\mbox{.}(2018a)]%
        {schelter2018challenges}
\bibfield{author}{\bibinfo{person}{Sebastian Schelter}, \bibinfo{person}{Felix Biessmann}, \bibinfo{person}{Tim Januschowski}, \bibinfo{person}{David Salinas}, \bibinfo{person}{Stephan Seufert}, {and} \bibinfo{person}{Gyuri Szarvas}.} \bibinfo{year}{2018}\natexlab{a}.
\newblock \showarticletitle{On challenges in machine learning model management}.
\newblock  (\bibinfo{year}{2018}).
\newblock


\bibitem[Schelter et~al\mbox{.}(2018b)]%
        {schelter2018automating}
\bibfield{author}{\bibinfo{person}{Sebastian Schelter}, \bibinfo{person}{Dustin Lange}, \bibinfo{person}{Philipp Schmidt}, \bibinfo{person}{Meltem Celikel}, \bibinfo{person}{Felix Biessmann}, {and} \bibinfo{person}{Andreas Grafberger}.} \bibinfo{year}{2018}\natexlab{b}.
\newblock \showarticletitle{Automating large-scale data quality verification}.
\newblock \bibinfo{journal}{\emph{Proceedings of the VLDB Endowment}} \bibinfo{volume}{11}, \bibinfo{number}{12} (\bibinfo{year}{2018}), \bibinfo{pages}{1781--1794}.
\newblock


\bibitem[Seliya et~al\mbox{.}(2009)]%
        {seliya2009study}
\bibfield{author}{\bibinfo{person}{Naeem Seliya}, \bibinfo{person}{Taghi~M Khoshgoftaar}, {and} \bibinfo{person}{Jason Van~Hulse}.} \bibinfo{year}{2009}\natexlab{}.
\newblock \showarticletitle{A study on the relationships of classifier performance metrics}. In \bibinfo{booktitle}{\emph{2009 21st IEEE international conference on tools with artificial intelligence}}. IEEE, \bibinfo{pages}{59--66}.
\newblock


\bibitem[Semola et~al\mbox{.}(2022)]%
        {semola2022continual}
\bibfield{author}{\bibinfo{person}{Rudy Semola}, \bibinfo{person}{Vincenzo Lomonaco}, {and} \bibinfo{person}{Davide Bacciu}.} \bibinfo{year}{2022}\natexlab{}.
\newblock \showarticletitle{Continual-Learning-as-a-Service (CLaaS): On-Demand Efficient Adaptation of Predictive Models}.
\newblock \bibinfo{journal}{\emph{arXiv preprint arXiv:2206.06957}} (\bibinfo{year}{2022}).
\newblock


\bibitem[Shafahi et~al\mbox{.}(2019)]%
        {shafahi2019adversarially}
\bibfield{author}{\bibinfo{person}{Ali Shafahi}, \bibinfo{person}{Parsa Saadatpanah}, \bibinfo{person}{Chen Zhu}, \bibinfo{person}{Amin Ghiasi}, \bibinfo{person}{Christoph Studer}, \bibinfo{person}{David Jacobs}, {and} \bibinfo{person}{Tom Goldstein}.} \bibinfo{year}{2019}\natexlab{}.
\newblock \showarticletitle{Adversarially robust transfer learning}.
\newblock \bibinfo{journal}{\emph{arXiv preprint arXiv:1905.08232}} (\bibinfo{year}{2019}).
\newblock


\bibitem[Shahoud et~al\mbox{.}(2019)]%
        {shahoud2019facilitating}
\bibfield{author}{\bibinfo{person}{Shadi Shahoud}, \bibinfo{person}{Sonja Gunnarsdottir}, \bibinfo{person}{Hatem Khalloof}, \bibinfo{person}{Clemens Duepmeier}, {and} \bibinfo{person}{Veit Hagenmeyer}.} \bibinfo{year}{2019}\natexlab{}.
\newblock \showarticletitle{Facilitating and managing machine learning and data analysis tasks in Big Data environments using web and microservice technologies}. In \bibinfo{booktitle}{\emph{Proceedings of the 11th International Conference on Management of Digital EcoSystems}}. \bibinfo{pages}{80--87}.
\newblock


\bibitem[Shrivastava et~al\mbox{.}(2020)]%
        {shrivastava2020dqlearn}
\bibfield{author}{\bibinfo{person}{Shrey Shrivastava}, \bibinfo{person}{Dhaval Patel}, \bibinfo{person}{Nianjun Zhou}, \bibinfo{person}{Arun Iyengar}, {and} \bibinfo{person}{Anuradha Bhamidipaty}.} \bibinfo{year}{2020}\natexlab{}.
\newblock \showarticletitle{DQLearn: a toolkit for structured data quality learning}. In \bibinfo{booktitle}{\emph{2020 IEEE International Conference on Big Data (Big Data)}}. IEEE, \bibinfo{pages}{1644--1653}.
\newblock


\bibitem[Silva et~al\mbox{.}(2020)]%
        {silva2020benchmarking}
\bibfield{author}{\bibinfo{person}{Lucas~Cardoso Silva}, \bibinfo{person}{Fernando~Rezende Zagatti}, \bibinfo{person}{Bruno~Silva Sette}, \bibinfo{person}{Lucas~Nildaimon dos Santos~Silva}, \bibinfo{person}{Daniel Lucr{\'e}dio}, \bibinfo{person}{Diego~Furtado Silva}, {and} \bibinfo{person}{Helena de Medeiros~Caseli}.} \bibinfo{year}{2020}\natexlab{}.
\newblock \showarticletitle{Benchmarking machine learning solutions in production}. In \bibinfo{booktitle}{\emph{2020 19th IEEE International Conference on Machine Learning and Applications (ICMLA)}}. IEEE, \bibinfo{pages}{626--633}.
\newblock


\bibitem[Soh and Singh(2020)]%
        {soh2020machine}
\bibfield{author}{\bibinfo{person}{Julian Soh} {and} \bibinfo{person}{Priyanshi Singh}.} \bibinfo{year}{2020}\natexlab{}.
\newblock \showarticletitle{Machine learning operations}.
\newblock In \bibinfo{booktitle}{\emph{Data Science Solutions on Azure}}. \bibinfo{publisher}{Springer}, \bibinfo{pages}{259--279}.
\newblock


\bibitem[Soin et~al\mbox{.}(2022)]%
        {soin2022chexstray}
\bibfield{author}{\bibinfo{person}{Arjun Soin}, \bibinfo{person}{Jameson Merkow}, \bibinfo{person}{Jin Long}, \bibinfo{person}{Joesph~Paul Cohen}, \bibinfo{person}{Smitha Saligrama}, \bibinfo{person}{Stephen Kaiser}, \bibinfo{person}{Steven Borg}, \bibinfo{person}{Ivan Tarapov}, {and} \bibinfo{person}{Matthew~P Lungren}.} \bibinfo{year}{2022}\natexlab{}.
\newblock \showarticletitle{CheXstray: Real-time Multi-Modal Data Concordance for Drift Detection in Medical Imaging AI}.
\newblock \bibinfo{journal}{\emph{arXiv preprint arXiv:2202.02833}} (\bibinfo{year}{2022}).
\newblock


\bibitem[Song et~al\mbox{.}(2019)]%
        {song2019privacy}
\bibfield{author}{\bibinfo{person}{Liwei Song}, \bibinfo{person}{Reza Shokri}, {and} \bibinfo{person}{Prateek Mittal}.} \bibinfo{year}{2019}\natexlab{}.
\newblock \showarticletitle{Privacy risks of securing machine learning models against adversarial examples}. In \bibinfo{booktitle}{\emph{Proceedings of the 2019 ACM SIGSAC Conference on Computer and Communications Security}}. \bibinfo{pages}{241--257}.
\newblock


\bibitem[Song et~al\mbox{.}(2021)]%
        {Song2021SEGA}
\bibfield{author}{\bibinfo{person}{Yiliao Song}, \bibinfo{person}{Jie Lu}, \bibinfo{person}{Anjin Liu}, \bibinfo{person}{Haiyan Lu}, {and} \bibinfo{person}{Guangquan Zhang}.} \bibinfo{year}{2021}\natexlab{}.
\newblock \showarticletitle{A Segment-Based Drift Adaptation Method for Data Streams}.
\newblock \bibinfo{journal}{\emph{IEEE Transactions on Neural Networks and Learning Systems}} (\bibinfo{year}{2021}).
\newblock


\bibitem[Spjuth et~al\mbox{.}(2021)]%
        {spjuth2021machine}
\bibfield{author}{\bibinfo{person}{Ola Spjuth}, \bibinfo{person}{Jens Frid}, {and} \bibinfo{person}{Andreas Hellander}.} \bibinfo{year}{2021}\natexlab{}.
\newblock \showarticletitle{The machine learning life cycle and the cloud: implications for drug discovery}.
\newblock \bibinfo{journal}{\emph{Expert opinion on drug discovery}} \bibinfo{volume}{16}, \bibinfo{number}{9} (\bibinfo{year}{2021}), \bibinfo{pages}{1071--1079}.
\newblock


\bibitem[Spray et~al\mbox{.}(2021)]%
        {spray2021building}
\bibfield{author}{\bibinfo{person}{John Spray}, \bibinfo{person}{Roopak Sinha}, \bibinfo{person}{Arnab Sen}, {and} \bibinfo{person}{Xingbin Cheng}.} \bibinfo{year}{2021}\natexlab{}.
\newblock \showarticletitle{Building Maintainable Software Using Abstraction Layering}.
\newblock \bibinfo{journal}{\emph{IEEE Transactions on Software Engineering}} \bibinfo{volume}{48}, \bibinfo{number}{11} (\bibinfo{year}{2021}), \bibinfo{pages}{4397--4410}.
\newblock


\bibitem[Steidl et~al\mbox{.}(2023)]%
        {steidl2023pipeline}
\bibfield{author}{\bibinfo{person}{Monika Steidl}, \bibinfo{person}{Michael Felderer}, {and} \bibinfo{person}{Rudolf Ramler}.} \bibinfo{year}{2023}\natexlab{}.
\newblock \showarticletitle{The pipeline for the continuous development of artificial intelligence models—Current state of research and practice}.
\newblock \bibinfo{journal}{\emph{Journal of Systems and Software}}  \bibinfo{volume}{199} (\bibinfo{year}{2023}), \bibinfo{pages}{111615}.
\newblock


\bibitem[Sugimoto et~al\mbox{.}(2013)]%
        {sugimoto2013comparison}
\bibfield{author}{\bibinfo{person}{Masahiro Sugimoto}, \bibinfo{person}{Masahiro Takada}, {and} \bibinfo{person}{Masakazu Toi}.} \bibinfo{year}{2013}\natexlab{}.
\newblock \showarticletitle{Comparison of robustness against missing values of alternative decision tree and multiple logistic regression for predicting clinical data in primary breast cancer}. In \bibinfo{booktitle}{\emph{2013 35th Annual International Conference of the IEEE Engineering in Medicine and Biology Society (EMBC)}}. IEEE, \bibinfo{pages}{3054--3057}.
\newblock


\bibitem[Sugiyama and Kawanabe(2012)]%
        {Sugiyama2012IntroductiontoCS}
\bibfield{author}{\bibinfo{person}{Masashi Sugiyama} {and} \bibinfo{person}{Motoaki Kawanabe}.} \bibinfo{year}{2012}\natexlab{}.
\newblock \bibinfo{booktitle}{\emph{Machine Learning in Non-Stationary Environments: Introduction to Covariate Shift Adaptation}}.
\newblock \bibinfo{publisher}{The MIT Press}.
\newblock
\showISBNx{0262017091}


\bibitem[Sugiyama et~al\mbox{.}(2007)]%
        {Sugiyama2007KLIEP}
\bibfield{author}{\bibinfo{person}{Masashi Sugiyama}, \bibinfo{person}{Shinichi Nakajima}, \bibinfo{person}{Hisashi Kashima}, \bibinfo{person}{Paul Buenau}, {and} \bibinfo{person}{Motoaki Kawanabe}.} \bibinfo{year}{2007}\natexlab{}.
\newblock \showarticletitle{Direct importance estimation with model selection and its application to covariate shift adaptation}.
\newblock \bibinfo{journal}{\emph{Advances in neural information processing systems}}  \bibinfo{volume}{20} (\bibinfo{year}{2007}).
\newblock


\bibitem[Symeonidis et~al\mbox{.}(2022)]%
        {symeonidis2022mlops}
\bibfield{author}{\bibinfo{person}{Georgios Symeonidis}, \bibinfo{person}{Evangelos Nerantzis}, \bibinfo{person}{Apostolos Kazakis}, {and} \bibinfo{person}{George~A Papakostas}.} \bibinfo{year}{2022}\natexlab{}.
\newblock \showarticletitle{MLOps-Definitions, Tools and Challenges}. In \bibinfo{booktitle}{\emph{2022 IEEE 12th Annual Computing and Communication Workshop and Conference (CCWC)}}. IEEE, \bibinfo{pages}{0453--0460}.
\newblock


\bibitem[Tamburri(2020)]%
        {tamburri2020sustainable}
\bibfield{author}{\bibinfo{person}{Damian~A Tamburri}.} \bibinfo{year}{2020}\natexlab{}.
\newblock \showarticletitle{Sustainable mlops: Trends and challenges}. In \bibinfo{booktitle}{\emph{2020 22nd international symposium on symbolic and numeric algorithms for scientific computing (SYNASC)}}. IEEE, \bibinfo{pages}{17--23}.
\newblock


\bibitem[Tan et~al\mbox{.}(2022)]%
        {tan2022risks}
\bibfield{author}{\bibinfo{person}{Samson Tan}, \bibinfo{person}{Araz Taeihagh}, {and} \bibinfo{person}{Kathy Baxter}.} \bibinfo{year}{2022}\natexlab{}.
\newblock \showarticletitle{The Risks of Machine Learning Systems}.
\newblock \bibinfo{journal}{\emph{arXiv preprint arXiv:2204.09852}} (\bibinfo{year}{2022}).
\newblock


\bibitem[Testi et~al\mbox{.}(2022)]%
        {testi2022mlops}
\bibfield{author}{\bibinfo{person}{Matteo Testi}, \bibinfo{person}{Matteo Ballabio}, \bibinfo{person}{Emanuele Frontoni}, \bibinfo{person}{Giulio Iannello}, \bibinfo{person}{Sara Moccia}, \bibinfo{person}{Paolo Soda}, {and} \bibinfo{person}{Gennaro Vessio}.} \bibinfo{year}{2022}\natexlab{}.
\newblock \showarticletitle{MLOps: A Taxonomy and a Methodology}.
\newblock \bibinfo{journal}{\emph{IEEE Access}} (\bibinfo{year}{2022}).
\newblock


\bibitem[Thornton et~al\mbox{.}(2013)]%
        {thornton2013auto}
\bibfield{author}{\bibinfo{person}{Chris Thornton}, \bibinfo{person}{Frank Hutter}, \bibinfo{person}{Holger~H Hoos}, {and} \bibinfo{person}{Kevin Leyton-Brown}.} \bibinfo{year}{2013}\natexlab{}.
\newblock \showarticletitle{Auto-WEKA: Combined selection and hyperparameter optimization of classification algorithms}. In \bibinfo{booktitle}{\emph{Proceedings of the 19th ACM SIGKDD international conference on Knowledge discovery and data mining}}. \bibinfo{pages}{847--855}.
\newblock


\bibitem[Thulasidasan et~al\mbox{.}(2021)]%
        {thulasidasan2021effective}
\bibfield{author}{\bibinfo{person}{Sunil Thulasidasan}, \bibinfo{person}{Sushil Thapa}, \bibinfo{person}{Sayera Dhaubhadel}, \bibinfo{person}{Gopinath Chennupati}, \bibinfo{person}{Tanmoy Bhattacharya}, {and} \bibinfo{person}{Jeff Bilmes}.} \bibinfo{year}{2021}\natexlab{}.
\newblock \showarticletitle{An effective baseline for robustness to distributional shift}. In \bibinfo{booktitle}{\emph{2021 20th IEEE International Conference on Machine Learning and Applications (ICMLA)}}. IEEE, \bibinfo{pages}{278--285}.
\newblock


\bibitem[Treveil et~al\mbox{.}(2020)]%
        {treveil2020introducing}
\bibfield{author}{\bibinfo{person}{Mark Treveil}, \bibinfo{person}{Nicolas Omont}, \bibinfo{person}{Cl{\'e}ment Stenac}, \bibinfo{person}{Kenji Lefevre}, \bibinfo{person}{Du Phan}, \bibinfo{person}{Joachim Zentici}, \bibinfo{person}{Adrien Lavoillotte}, \bibinfo{person}{Makoto Miyazaki}, {and} \bibinfo{person}{Lynn Heidmann}.} \bibinfo{year}{2020}\natexlab{}.
\newblock \bibinfo{booktitle}{\emph{Introducing MLOps}}.
\newblock \bibinfo{publisher}{O'Reilly Media}.
\newblock


\bibitem[Tsipras et~al\mbox{.}(2018)]%
        {tsipras2018robustness}
\bibfield{author}{\bibinfo{person}{Dimitris Tsipras}, \bibinfo{person}{Shibani Santurkar}, \bibinfo{person}{Logan Engstrom}, \bibinfo{person}{Alexander Turner}, {and} \bibinfo{person}{Aleksander Madry}.} \bibinfo{year}{2018}\natexlab{}.
\newblock \showarticletitle{Robustness may be at odds with accuracy}.
\newblock \bibinfo{journal}{\emph{arXiv preprint arXiv:1805.12152}} (\bibinfo{year}{2018}).
\newblock


\bibitem[Tsymbal et~al\mbox{.}(2008)]%
        {Tsymbal2008Local}
\bibfield{author}{\bibinfo{person}{Alexey Tsymbal}, \bibinfo{person}{Mykola Pechenizkiy}, \bibinfo{person}{Pádraig Cunningham}, {and} \bibinfo{person}{Seppo Puuronen}.} \bibinfo{year}{2008}\natexlab{}.
\newblock \showarticletitle{Dynamic integration of classifiers for handling concept drift}.
\newblock \bibinfo{journal}{\emph{Information Fusion}} \bibinfo{volume}{9}, \bibinfo{number}{1} (\bibinfo{year}{2008}), \bibinfo{pages}{56--68}.
\newblock
\showISSN{1566-2535}
\newblock
\shownote{Special Issue on Applications of Ensemble Methods}.


\bibitem[Usuga~Cadavid et~al\mbox{.}(2020)]%
        {usuga2020machine}
\bibfield{author}{\bibinfo{person}{Juan~Pablo Usuga~Cadavid}, \bibinfo{person}{Samir Lamouri}, \bibinfo{person}{Bernard Grabot}, \bibinfo{person}{Robert Pellerin}, {and} \bibinfo{person}{Arnaud Fortin}.} \bibinfo{year}{2020}\natexlab{}.
\newblock \showarticletitle{Machine learning applied in production planning and control: a state-of-the-art in the era of industry 4.0}.
\newblock \bibinfo{journal}{\emph{Journal of Intelligent Manufacturing}} \bibinfo{volume}{31}, \bibinfo{number}{6} (\bibinfo{year}{2020}), \bibinfo{pages}{1531--1558}.
\newblock


\bibitem[Vabalas et~al\mbox{.}(2019)]%
        {vabalas2019machine}
\bibfield{author}{\bibinfo{person}{Andrius Vabalas}, \bibinfo{person}{Emma Gowen}, \bibinfo{person}{Ellen Poliakoff}, {and} \bibinfo{person}{Alexander~J Casson}.} \bibinfo{year}{2019}\natexlab{}.
\newblock \showarticletitle{Machine learning algorithm validation with a limited sample size}.
\newblock \bibinfo{journal}{\emph{PloS one}} \bibinfo{volume}{14}, \bibinfo{number}{11} (\bibinfo{year}{2019}), \bibinfo{pages}{e0224365}.
\newblock


\bibitem[Van Der~Weide et~al\mbox{.}(2017)]%
        {van2017versioning}
\bibfield{author}{\bibinfo{person}{Tom Van Der~Weide}, \bibinfo{person}{Dimitris Papadopoulos}, \bibinfo{person}{Oleg Smirnov}, \bibinfo{person}{Michal Zielinski}, {and} \bibinfo{person}{Tim Van~Kasteren}.} \bibinfo{year}{2017}\natexlab{}.
\newblock \showarticletitle{Versioning for end-to-end machine learning pipelines}. In \bibinfo{booktitle}{\emph{Proceedings of the 1st Workshop on Data Management for End-to-End Machine Learning}}. \bibinfo{pages}{1--9}.
\newblock


\bibitem[Vartak et~al\mbox{.}(2015)]%
        {vartak2015supporting}
\bibfield{author}{\bibinfo{person}{Manasi Vartak}, \bibinfo{person}{Pablo Ortiz}, \bibinfo{person}{Kathryn Siegel}, \bibinfo{person}{Harihar Subramanyam}, \bibinfo{person}{Samuel Madden}, {and} \bibinfo{person}{Matei Zaharia}.} \bibinfo{year}{2015}\natexlab{}.
\newblock \showarticletitle{Supporting fast iteration in model building}. In \bibinfo{booktitle}{\emph{NIPS Workshop LearningSys}}. \bibinfo{pages}{1--6}.
\newblock


\bibitem[Vieira et~al\mbox{.}(2021)]%
        {vieira2021driftage}
\bibfield{author}{\bibinfo{person}{Diogo~Munaro Vieira}, \bibinfo{person}{Chrystinne Fernandes}, \bibinfo{person}{Carlos Lucena}, {and} \bibinfo{person}{S{\'e}rgio Lifschitz}.} \bibinfo{year}{2021}\natexlab{}.
\newblock \showarticletitle{Driftage: a multi-agent system framework for concept drift detection}.
\newblock \bibinfo{journal}{\emph{GigaScience}} \bibinfo{volume}{10}, \bibinfo{number}{6} (\bibinfo{year}{2021}), \bibinfo{pages}{giab030}.
\newblock


\bibitem[Vincent-Lancrin and van~der Vlies(2020)]%
        {vincent2020trustworthy}
\bibfield{author}{\bibinfo{person}{St{\'e}phan Vincent-Lancrin} {and} \bibinfo{person}{Reyer van~der Vlies}.} \bibinfo{year}{2020}\natexlab{}.
\newblock \showarticletitle{Trustworthy artificial intelligence (AI) in education: Promises and challenges}.
\newblock  (\bibinfo{year}{2020}).
\newblock


\bibitem[Virmani(2015)]%
        {virmani2015understanding}
\bibfield{author}{\bibinfo{person}{Manish Virmani}.} \bibinfo{year}{2015}\natexlab{}.
\newblock \showarticletitle{Understanding DevOps \& bridging the gap from continuous integration to continuous delivery}. In \bibinfo{booktitle}{\emph{Fifth international conference on the innovative computing technology (intech 2015)}}. IEEE, \bibinfo{pages}{78--82}.
\newblock


\bibitem[Wagstaff(2012)]%
        {wagstaff2012machine}
\bibfield{author}{\bibinfo{person}{Kiri Wagstaff}.} \bibinfo{year}{2012}\natexlab{}.
\newblock \showarticletitle{Machine learning that matters}.
\newblock \bibinfo{journal}{\emph{arXiv preprint arXiv:1206.4656}} (\bibinfo{year}{2012}).
\newblock


\bibitem[Wang et~al\mbox{.}(2022)]%
        {wang2022generalizing}
\bibfield{author}{\bibinfo{person}{Jindong Wang}, \bibinfo{person}{Cuiling Lan}, \bibinfo{person}{Chang Liu}, \bibinfo{person}{Yidong Ouyang}, \bibinfo{person}{Tao Qin}, \bibinfo{person}{Wang Lu}, \bibinfo{person}{Yiqiang Chen}, \bibinfo{person}{Wenjun Zeng}, {and} \bibinfo{person}{Philip Yu}.} \bibinfo{year}{2022}\natexlab{}.
\newblock \showarticletitle{Generalizing to unseen domains: A survey on domain generalization}.
\newblock \bibinfo{journal}{\emph{IEEE Transactions on Knowledge and Data Engineering}} (\bibinfo{year}{2022}).
\newblock


\bibitem[Wang and Wang(2019)]%
        {wang2019time}
\bibfield{author}{\bibinfo{person}{Xi Wang} {and} \bibinfo{person}{Chen Wang}.} \bibinfo{year}{2019}\natexlab{}.
\newblock \showarticletitle{Time series data cleaning: A survey}.
\newblock \bibinfo{journal}{\emph{Ieee Access}}  \bibinfo{volume}{8} (\bibinfo{year}{2019}), \bibinfo{pages}{1866--1881}.
\newblock


\bibitem[Wing(2021)]%
        {wing2021trustworthy}
\bibfield{author}{\bibinfo{person}{Jeannette~M Wing}.} \bibinfo{year}{2021}\natexlab{}.
\newblock \showarticletitle{Trustworthy {AI}}.
\newblock \bibinfo{journal}{\emph{Commun. ACM}} \bibinfo{volume}{64}, \bibinfo{number}{10} (\bibinfo{year}{2021}), \bibinfo{pages}{64--71}.
\newblock


\bibitem[Witten et~al\mbox{.}(2017)]%
        {WITTEN2017Probabilistic}
\bibfield{author}{\bibinfo{person}{Ian~H. Witten}, \bibinfo{person}{Eibe Frank}, \bibinfo{person}{Mark~A. Hall}, {and} \bibinfo{person}{Christopher~J. Pal}.} \bibinfo{year}{2017}\natexlab{}.
\newblock \showarticletitle{Chapter 9 - Probabilistic methods}.
\newblock In \bibinfo{booktitle}{\emph{Data Mining: Practical Machine Learning Tools and Techniques} (\bibinfo{edition}{4th} ed.)}, \bibfield{editor}{\bibinfo{person}{Ian~H. Witten}, \bibinfo{person}{Eibe Frank}, \bibinfo{person}{Mark~A. Hall}, {and} \bibinfo{person}{Christopher~J. Pal}} (Eds.). \bibinfo{publisher}{Morgan Kaufmann}, \bibinfo{pages}{335--416}.
\newblock
\showISBNx{978-0-12-804291-5}


\bibitem[Woitsch et~al\mbox{.}(2021)]%
        {woitsch2021collaborative}
\bibfield{author}{\bibinfo{person}{Robert Woitsch}, \bibinfo{person}{Wilfrid Utz}, \bibinfo{person}{Anna Sumereder}, \bibinfo{person}{Bernhard Dieber}, \bibinfo{person}{Benjamin Breiling}, \bibinfo{person}{Laura Crompton}, \bibinfo{person}{Michael Funk}, \bibinfo{person}{Karin Bruckm{\"u}ller}, {and} \bibinfo{person}{Stefan Schumann}.} \bibinfo{year}{2021}\natexlab{}.
\newblock \showarticletitle{Collaborative model-based process assessment for trustworthy AI in robotic platforms}. In \bibinfo{booktitle}{\emph{International Conference on Society 5.0}}. Springer, \bibinfo{pages}{163--174}.
\newblock


\bibitem[Wu et~al\mbox{.}(2020c)]%
        {wu2020collaborate}
\bibfield{author}{\bibinfo{person}{Huaming Wu}, \bibinfo{person}{Ziru Zhang}, \bibinfo{person}{Chang Guan}, \bibinfo{person}{Katinka Wolter}, {and} \bibinfo{person}{Minxian Xu}.} \bibinfo{year}{2020}\natexlab{c}.
\newblock \showarticletitle{Collaborate edge and cloud computing with distributed deep learning for smart city internet of things}.
\newblock \bibinfo{journal}{\emph{IEEE Internet of Things Journal}} \bibinfo{volume}{7}, \bibinfo{number}{9} (\bibinfo{year}{2020}), \bibinfo{pages}{8099--8110}.
\newblock


\bibitem[Wu et~al\mbox{.}(2020a)]%
        {wu2020deltagrad}
\bibfield{author}{\bibinfo{person}{Yinjun Wu}, \bibinfo{person}{Edgar Dobriban}, {and} \bibinfo{person}{Susan Davidson}.} \bibinfo{year}{2020}\natexlab{a}.
\newblock \showarticletitle{Deltagrad: Rapid retraining of machine learning models}. In \bibinfo{booktitle}{\emph{International Conference on Machine Learning}}. PMLR, \bibinfo{pages}{10355--10366}.
\newblock


\bibitem[Wu et~al\mbox{.}(2020b)]%
        {wu2020priu}
\bibfield{author}{\bibinfo{person}{Yinjun Wu}, \bibinfo{person}{Val Tannen}, {and} \bibinfo{person}{Susan~B Davidson}.} \bibinfo{year}{2020}\natexlab{b}.
\newblock \showarticletitle{Priu: A provenance-based approach for incrementally updating regression models}. In \bibinfo{booktitle}{\emph{Proceedings of the 2020 ACM SIGMOD International Conference on Management of Data}}. \bibinfo{pages}{447--462}.
\newblock


\bibitem[Xiao et~al\mbox{.}(2015)]%
        {xiao2015learning}
\bibfield{author}{\bibinfo{person}{Tong Xiao}, \bibinfo{person}{Tian Xia}, \bibinfo{person}{Yi Yang}, \bibinfo{person}{Chang Huang}, {and} \bibinfo{person}{Xiaogang Wang}.} \bibinfo{year}{2015}\natexlab{}.
\newblock \showarticletitle{Learning from massive noisy labeled data for image classification}. In \bibinfo{booktitle}{\emph{Proceedings of the IEEE conference on computer vision and pattern recognition}}. \bibinfo{pages}{2691--2699}.
\newblock


\bibitem[Xin et~al\mbox{.}(2018)]%
        {xin2018accelerating}
\bibfield{author}{\bibinfo{person}{Doris Xin}, \bibinfo{person}{Litian Ma}, \bibinfo{person}{Jialin Liu}, \bibinfo{person}{Stephen Macke}, \bibinfo{person}{Shuchen Song}, {and} \bibinfo{person}{Aditya Parameswaran}.} \bibinfo{year}{2018}\natexlab{}.
\newblock \showarticletitle{Accelerating human-in-the-loop machine learning: Challenges and opportunities}. In \bibinfo{booktitle}{\emph{Proceedings of the second workshop on data management for end-to-end machine learning}}. \bibinfo{pages}{1--4}.
\newblock


\bibitem[Xu et~al\mbox{.}(2024)]%
        {xu2024scarcity}
\bibfield{author}{\bibinfo{person}{Chaobin Xu}, \bibinfo{person}{Wei Li}, \bibinfo{person}{Xiaohui Cui}, \bibinfo{person}{Zhenyu Wang}, \bibinfo{person}{Fengling Zheng}, \bibinfo{person}{Xiaowu Zhang}, {and} \bibinfo{person}{Bin Chen}.} \bibinfo{year}{2024}\natexlab{}.
\newblock \showarticletitle{Scarcity-GAN: Scarce data augmentation for defect detection via generative adversarial nets}.
\newblock \bibinfo{journal}{\emph{Neurocomputing}}  \bibinfo{volume}{566} (\bibinfo{year}{2024}), \bibinfo{pages}{127061}.
\newblock


\bibitem[Yang et~al\mbox{.}(2021)]%
        {yang2021cade}
\bibfield{author}{\bibinfo{person}{Limin Yang}, \bibinfo{person}{Wenbo Guo}, \bibinfo{person}{Qingying Hao}, \bibinfo{person}{Arridhana Ciptadi}, \bibinfo{person}{Ali Ahmadzadeh}, \bibinfo{person}{Xinyu Xing}, {and} \bibinfo{person}{Gang Wang}.} \bibinfo{year}{2021}\natexlab{}.
\newblock \showarticletitle{{CADE}: Detecting and explaining concept drift samples for security applications}. In \bibinfo{booktitle}{\emph{30th USENIX Security Symposium (USENIX Security 21)}}. \bibinfo{pages}{2327--2344}.
\newblock


\bibitem[Yang et~al\mbox{.}(2020)]%
        {yang2020closer}
\bibfield{author}{\bibinfo{person}{Yao-Yuan Yang}, \bibinfo{person}{Cyrus Rashtchian}, \bibinfo{person}{Hongyang Zhang}, \bibinfo{person}{Russ~R Salakhutdinov}, {and} \bibinfo{person}{Kamalika Chaudhuri}.} \bibinfo{year}{2020}\natexlab{}.
\newblock \showarticletitle{A closer look at accuracy vs. robustness}.
\newblock \bibinfo{journal}{\emph{Advances in neural information processing systems}}  \bibinfo{volume}{33} (\bibinfo{year}{2020}), \bibinfo{pages}{8588--8601}.
\newblock


\bibitem[Ying(2019)]%
        {ying2019overview}
\bibfield{author}{\bibinfo{person}{Xue Ying}.} \bibinfo{year}{2019}\natexlab{}.
\newblock \showarticletitle{An overview of overfitting and its solutions}. In \bibinfo{booktitle}{\emph{Journal of physics: Conference series}}, Vol.~\bibinfo{volume}{1168}. IOP Publishing, \bibinfo{pages}{022022}.
\newblock


\bibitem[Young and Steele(2022)]%
        {young2022empirical}
\bibfield{author}{\bibinfo{person}{Zachary Young} {and} \bibinfo{person}{Robert Steele}.} \bibinfo{year}{2022}\natexlab{}.
\newblock \showarticletitle{Empirical evaluation of performance degradation of machine learning-based predictive models--A case study in healthcare information systems}.
\newblock \bibinfo{journal}{\emph{International Journal of Information Management Data Insights}} \bibinfo{volume}{2}, \bibinfo{number}{1} (\bibinfo{year}{2022}), \bibinfo{pages}{100070}.
\newblock


\bibitem[Yu et~al\mbox{.}(2022)]%
        {yu2022dataops}
\bibfield{author}{\bibinfo{person}{Shaochen Yu}, \bibinfo{person}{Tianwa Chen}, \bibinfo{person}{Lei Han}, \bibinfo{person}{Gianluca Demartini}, {and} \bibinfo{person}{Shazia Sadiq}.} \bibinfo{year}{2022}\natexlab{}.
\newblock \showarticletitle{DataOps-4G: On Supporting Generalists in Data Quality Discovery}.
\newblock \bibinfo{journal}{\emph{IEEE Transactions on Knowledge and Data Engineering}} (\bibinfo{year}{2022}).
\newblock


\bibitem[Zaharia et~al\mbox{.}(2018)]%
        {zaharia2018accelerating}
\bibfield{author}{\bibinfo{person}{Matei Zaharia}, \bibinfo{person}{Andrew Chen}, \bibinfo{person}{Aaron Davidson}, \bibinfo{person}{Ali Ghodsi}, \bibinfo{person}{Sue~Ann Hong}, \bibinfo{person}{Andy Konwinski}, \bibinfo{person}{Siddharth Murching}, \bibinfo{person}{Tomas Nykodym}, \bibinfo{person}{Paul Ogilvie}, \bibinfo{person}{Mani Parkhe}, {et~al\mbox{.}}} \bibinfo{year}{2018}\natexlab{}.
\newblock \showarticletitle{Accelerating the machine learning lifecycle with MLflow.}
\newblock \bibinfo{journal}{\emph{IEEE Data Eng. Bull.}} \bibinfo{volume}{41}, \bibinfo{number}{4} (\bibinfo{year}{2018}), \bibinfo{pages}{39--45}.
\newblock


\bibitem[Z{\'a}rate et~al\mbox{.}(2022)]%
        {zarate2022k2e}
\bibfield{author}{\bibinfo{person}{Gorka Z{\'a}rate}, \bibinfo{person}{Ra{\'u}l Mi{\~n}{\'o}n}, \bibinfo{person}{Josu D{\'\i}az-de Arcaya}, {and} \bibinfo{person}{Ana~I Torre-Bastida}.} \bibinfo{year}{2022}\natexlab{}.
\newblock \showarticletitle{K2E: Building MLOps Environments for Governing Data and Models Catalogues while Tracking Versions}. In \bibinfo{booktitle}{\emph{2022 IEEE 19th International Conference on Software Architecture Companion (ICSA-C)}}. IEEE, \bibinfo{pages}{206--209}.
\newblock


\bibitem[Zeller et~al\mbox{.}(2024)]%
        {zeller2024toward}
\bibfield{author}{\bibinfo{person}{Marc Zeller}, \bibinfo{person}{Thomas Waschulzik}, \bibinfo{person}{Reiner Schmid}, {and} \bibinfo{person}{Claus Bahlmann}.} \bibinfo{year}{2024}\natexlab{}.
\newblock \showarticletitle{Toward a safe MLOps process for the continuous development and safety assurance of ML-based systems in the railway domain}.
\newblock \bibinfo{journal}{\emph{AI and Ethics}} (\bibinfo{year}{2024}), \bibinfo{pages}{1--8}.
\newblock


\bibitem[Zenke et~al\mbox{.}(2017)]%
        {zenke2017continual}
\bibfield{author}{\bibinfo{person}{Friedemann Zenke}, \bibinfo{person}{Ben Poole}, {and} \bibinfo{person}{Surya Ganguli}.} \bibinfo{year}{2017}\natexlab{}.
\newblock \showarticletitle{Continual learning through synaptic intelligence}. In \bibinfo{booktitle}{\emph{International Conference on Machine Learning}}. PMLR, \bibinfo{pages}{3987--3995}.
\newblock


\bibitem[Zhang et~al\mbox{.}(2021)]%
        {zhang2021survey}
\bibfield{author}{\bibinfo{person}{Chen Zhang}, \bibinfo{person}{Yu Xie}, \bibinfo{person}{Hang Bai}, \bibinfo{person}{Bin Yu}, \bibinfo{person}{Weihong Li}, {and} \bibinfo{person}{Yuan Gao}.} \bibinfo{year}{2021}\natexlab{}.
\newblock \showarticletitle{A survey on federated learning}.
\newblock \bibinfo{journal}{\emph{Knowledge-Based Systems}}  \bibinfo{volume}{216} (\bibinfo{year}{2021}), \bibinfo{pages}{106775}.
\newblock


\bibitem[Zhang et~al\mbox{.}(2020)]%
        {zhang2020generalizing}
\bibfield{author}{\bibinfo{person}{Ling Zhang}, \bibinfo{person}{Xiaosong Wang}, \bibinfo{person}{Dong Yang}, \bibinfo{person}{Thomas Sanford}, \bibinfo{person}{Stephanie Harmon}, \bibinfo{person}{Baris Turkbey}, \bibinfo{person}{Bradford~J Wood}, \bibinfo{person}{Holger Roth}, \bibinfo{person}{Andriy Myronenko}, \bibinfo{person}{Daguang Xu}, {et~al\mbox{.}}} \bibinfo{year}{2020}\natexlab{}.
\newblock \showarticletitle{Generalizing deep learning for medical image segmentation to unseen domains via deep stacked transformation}.
\newblock \bibinfo{journal}{\emph{IEEE transactions on medical imaging}} \bibinfo{volume}{39}, \bibinfo{number}{7} (\bibinfo{year}{2020}), \bibinfo{pages}{2531--2540}.
\newblock


\bibitem[Zhao(2021)]%
        {zhao2021machine}
\bibfield{author}{\bibinfo{person}{Yizhen Zhao}.} \bibinfo{year}{2021}\natexlab{}.
\newblock \bibinfo{title}{Machine Learning in Production: A Literature Review}.
\newblock
\newblock


\bibitem[Zhao et~al\mbox{.}(2021)]%
        {zhao2021impact}
\bibfield{author}{\bibinfo{person}{Yanjie Zhao}, \bibinfo{person}{Li Li}, \bibinfo{person}{Haoyu Wang}, \bibinfo{person}{Haipeng Cai}, \bibinfo{person}{Tegawend{\'e}~F Bissyand{\'e}}, \bibinfo{person}{Jacques Klein}, {and} \bibinfo{person}{John Grundy}.} \bibinfo{year}{2021}\natexlab{}.
\newblock \showarticletitle{On the impact of sample duplication in machine-learning-based android malware detection}.
\newblock \bibinfo{journal}{\emph{ACM Transactions on Software Engineering and Methodology (TOSEM)}} \bibinfo{volume}{30}, \bibinfo{number}{3} (\bibinfo{year}{2021}), \bibinfo{pages}{1--38}.
\newblock


\bibitem[Zhou and Levine(2021)]%
        {zhou2021bayesian}
\bibfield{author}{\bibinfo{person}{Aurick Zhou} {and} \bibinfo{person}{Sergey Levine}.} \bibinfo{year}{2021}\natexlab{}.
\newblock \showarticletitle{Bayesian Adaptation for Covariate Shift}.
\newblock \bibinfo{journal}{\emph{Advances in Neural Information Processing Systems}}  \bibinfo{volume}{34} (\bibinfo{year}{2021}), \bibinfo{pages}{914--927}.
\newblock


\bibitem[Zhou et~al\mbox{.}(2021)]%
        {zhou2021domain}
\bibfield{author}{\bibinfo{person}{Kaiyang Zhou}, \bibinfo{person}{Ziwei Liu}, \bibinfo{person}{Yu Qiao}, \bibinfo{person}{Tao Xiang}, {and} \bibinfo{person}{Chen~Change Loy}.} \bibinfo{year}{2021}\natexlab{}.
\newblock \showarticletitle{Domain generalization in vision: A survey}.
\newblock \bibinfo{journal}{\emph{arXiv preprint arXiv:2103.02503}} (\bibinfo{year}{2021}).
\newblock


\bibitem[Zhou et~al\mbox{.}(2017)]%
        {zhou2017machine}
\bibfield{author}{\bibinfo{person}{Lina Zhou}, \bibinfo{person}{Shimei Pan}, \bibinfo{person}{Jianwu Wang}, {and} \bibinfo{person}{Athanasios~V Vasilakos}.} \bibinfo{year}{2017}\natexlab{}.
\newblock \showarticletitle{Machine learning on big data: Opportunities and challenges}.
\newblock \bibinfo{journal}{\emph{Neurocomputing}}  \bibinfo{volume}{237} (\bibinfo{year}{2017}), \bibinfo{pages}{350--361}.
\newblock


\bibitem[Zhou et~al\mbox{.}(2022)]%
        {zhou2022time}
\bibfield{author}{\bibinfo{person}{Xiaofeng Zhou}, \bibinfo{person}{Naiju Zhai}, \bibinfo{person}{Shuai Li}, {and} \bibinfo{person}{Haibo Shi}.} \bibinfo{year}{2022}\natexlab{}.
\newblock \showarticletitle{Time Series Prediction Method of Industrial Process with Limited Data Based on Transfer Learning}.
\newblock \bibinfo{journal}{\emph{IEEE Transactions on Industrial Informatics}} (\bibinfo{year}{2022}).
\newblock


\bibitem[Zhou(2022)]%
        {zhou2022open}
\bibfield{author}{\bibinfo{person}{Zhi-Hua Zhou}.} \bibinfo{year}{2022}\natexlab{}.
\newblock \showarticletitle{Open-environment machine learning}.
\newblock \bibinfo{journal}{\emph{National Science Review}} \bibinfo{volume}{9}, \bibinfo{number}{8} (\bibinfo{year}{2022}), \bibinfo{pages}{nwac123}.
\newblock


\bibitem[{\v{Z}}liobaite(2010)]%
        {vzliobaite2010change}
\bibfield{author}{\bibinfo{person}{Indre {\v{Z}}liobaite}.} \bibinfo{year}{2010}\natexlab{}.
\newblock \showarticletitle{Change with delayed labeling: When is it detectable?}. In \bibinfo{booktitle}{\emph{2010 IEEE International Conference on Data Mining Workshops}}. IEEE, \bibinfo{pages}{843--850}.
\newblock


\end{thebibliography}
